\documentclass[letterpaper,11pt]{article}
\usepackage{parskip}
\usepackage[english]{babel}
\usepackage[nottoc]{tocbibind}

\usepackage[square]{natbib}

\usepackage{diagbox}
\usepackage{makecell}
\usepackage{booktabs}
\usepackage{tikz}
\usepackage{amsmath}
\usepackage{amsfonts}
\usepackage{amssymb}
\usepackage{amsthm}
\usepackage{url,ifthen}
\usepackage{enumitem}
\usepackage{srcltx}
\usepackage{multirow}
\usepackage{boxedminipage}
\usepackage[margin=1.1in]{geometry}
\usepackage{nicefrac}
\usepackage{xspace}
\usepackage{graphicx}
\usepackage{color}
\usepackage{colortbl}
\usepackage{setspace}
\usepackage{soul}

\usepackage{hyperref}
\usepackage{cleveref}
\usepackage{mathptmx}

\definecolor{DarkGreen}{rgb}{0.1,0.5,0.1}
\definecolor{DarkRed}{rgb}{0.5,0.1,0.1}
\definecolor{DarkBlue}{rgb}{0.1,0.1,0.5}
\definecolor{Gray}{rgb}{0.2,0.2,0.2}

\usepackage{listings}
\lstdefinestyle{mystyle}{
    commentstyle=\color{DarkBlue},
    keywordstyle=\color{DarkRed},
    numberstyle=\tiny\color{Gray},
    stringstyle=\color{DarkGreen},
    basicstyle=\footnotesize,
    breakatwhitespace=false,         
    breaklines=true,                 
    captionpos=b,                    
    keepspaces=true,                 
    numbers=left,                    
    numbersep=5pt,                  
    showspaces=false,                
    showstringspaces=false,
    showtabs=false,                  
    tabsize=2
}
\usepackage[small]{caption}
\usepackage{hyperref}
\hypersetup{
    unicode=false,          
    pdftoolbar=true,        
    pdfmenubar=true,        
    pdffitwindow=false,      
    pdfnewwindow=true,      
    colorlinks=true,       
    linkcolor=DarkGreen,          
    citecolor=DarkGreen,        
    filecolor=DarkRed,      
    urlcolor=DarkBlue,          
    pdftitle={},
    pdfauthor={},
}

\def\draft{1}

\def\submit{0}

\ifnum\draft=1 
    \def\ShowAuthNotes{1}
\else
    \def\ShowAuthNotes{0}
\fi

\ifnum\submit=1
\newcommand{\forsubmit}[1]{#1}
\newcommand{\forreals}[1]{}
\else
\newcommand{\forreals}[1]{#1}
\newcommand{\forsubmit}[1]{}
\fi

\ifnum\ShowAuthNotes=1
\newcommand{\authnote}[2]{{ \footnotesize \bf{\color{DarkRed}[#1's Note:
{\color{DarkBlue}#2}]}}}
\else
\newcommand{\authnote}[2]{}
\fi

\newtheorem{theorem}{Theorem}[section]

\newtheorem{lemma}[theorem]{Lemma}
\newtheorem{corollary}[theorem]{Corollary}
\newtheorem{proposition}[theorem]{Proposition}

\theoremstyle{definition}
\newtheorem{definition}[theorem]{Definition}
\newtheorem{assumption}[theorem]{Assumption}

\newtheoremstyle{example_contd}
{\topsep} {\topsep}%
{}
{}
{\bfseries}
{.}
{1em}
{\thmname{#1} \thmnumber{ #2}\thmnote{#3} (continued)}

\theoremstyle{example_contd}

\newcommand{\chapterref}[1]{\hyperref[ch:#1]{Chapter~\ref{ch:#1}}}

\newcommand{\claimref}[1]{\hyperref[claim:#1]{Claim~\ref{claim:#1}}}

\newcommand{\corollaryref}[1]{\hyperref[cor:#1]{Corollary~\ref{cor:#1}}}

\newcommand{\definitionref}[1]{\hyperref[def:#1]{Definition~\ref{def:#1}}}

\newcommand{\equationref}[1]{\hyperref[eq:#1]{Equation~\ref{eq:#1}}}

\newcommand{\factref}[1]{\hyperref[fact:#1]{Fact~\ref{fact:#1}}}

\newcommand{\figureref}[1]{\hyperref[fig:#1]{Figure~\ref{fig:#1}}}

\newcommand{\tableref}[1]{\hyperref[tab:#1]{Table~\ref{tab:#1}}}

\newcommand{\itemref}[1]{\hyperref[item:#1]{Item~(\ref{item:#1})}}

\newcommand{\lemmaref}[1]{\hyperref[lem:#1]{Lemma~\ref{lem:#1}}}

\newcommand{\propref}[1]{\hyperref[prop:#1]{Proposition~\ref{prop:#1}}}

\newcommand{\propositionref}[1]{\hyperref[prop:#1]{Proposition~\ref{prop:#1}}}

\newcommand{\remarkref}[1]{\hyperref[rem:#1]{Remark~\ref{rem:#1}}}

\newcommand{\sectionref}[1]{\hyperref[sec:#1]{Section~\ref{sec:#1}}}

\newcommand{\theoremref}[1]{\hyperref[thm:#1]{Theorem~\ref{thm:#1}}}

\usepackage[utf8]{inputenc}
\usepackage[T1]{fontenc}
\usepackage{kpfonts}
\usepackage{microtype}

\renewcommand{\hat}{\widehat}

\renewcommand{\leq}{\leqslant}
\renewcommand{\le}{\leqslant}
\renewcommand{\geq}{\geqslant}
\renewcommand{\ge}{\geqslant}

\usepackage{bm}

\newcommand{\ignore}[1]{}

\renewcommand{\epsilon}{\varepsilon}

\newcommand{\remove}[1]{}

\usepackage{color-edits}
\addauthor{hui}{cyan}

\usepackage{bm}
\usepackage{amsfonts}
\usepackage[utf8]{inputenc} 
\usepackage[T1]{fontenc}    
\usepackage{hyperref}       
\usepackage{url}            
\usepackage{booktabs}       
\usepackage{amsfonts}       
\usepackage{nicefrac}       
\usepackage{microtype}      
\usepackage{xcolor}         

\usepackage{algorithm}
\usepackage{algorithmic}
\usepackage{diagbox}
\usepackage{makecell}
\usepackage{booktabs}
\usepackage{tikz}
\usepackage{amsmath}
\allowdisplaybreaks
\usepackage{amssymb}
\usepackage{amsthm}
\usepackage{url,ifthen}
\usepackage{enumitem}
\usepackage{srcltx}
\usepackage{multirow}
\usepackage{boxedminipage}
\usepackage{nicefrac}
\usepackage{xspace}
\usepackage{graphicx}
\usepackage{color}
\usepackage{colortbl}
\usepackage{setspace}
\usepackage{soul}
\usepackage{graphicx}
\usepackage{wrapfig}
\usepackage{subcaption}
\usepackage{enumitem}
\usepackage{cleveref}
\definecolor{DarkGreen}{rgb}{0.1,0.5,0.1}
\definecolor{DarkRed}{rgb}{0.5,0.1,0.1}
\definecolor{DarkBlue}{rgb}{0.1,0.1,0.5}
\definecolor{Gray}{rgb}{0.2,0.2,0.2}

\title{Provable Pluralistic Alignment: Multi-Party RLHF under Offline Human Feedback}

\usepackage{authblk}

\author[1]{Huiying Zhong}
\author[2]{Tianwei Gao}
\author[3]{Zhiwei Steven Wu}
\author[4]{Linjun Zhang}
\author[5]{Weijie J.~Su}
\author[2]{Zhun Deng}
\affil[1]{Massachusetts Institute of Technology}
\affil[2]{University of North Carolina at Chapel Hill}
\affil[3]{Carnegie Mellon University}
\affil[4]{Rutgers University}
\affil[5]{University of Pennsylvania}

\date{}
\begin{document}

\maketitle

\begin{abstract}

Pluralistic alignment requires learning from feedback that reflects persistent and potentially conflicting stakeholder preferences while ultimately selecting a single collective policy. We study this problem in offline reinforcement learning from human feedback (RLHF), where the party associated with each comparison is observed. Under a shared low-rank linear reward model, we jointly estimate party-specific rewards and perform pessimistic policy optimization under Nash, Utilitarian, and Egalitarian social-welfare objectives. We establish nonasymptotic bounds for party-specific reward estimation and the resulting policy suboptimality under offline coverage conditions. We further consider general pairwise preferences that need not admit a scalar reward representation and may exhibit cycles. In this setting, we construct a pessimistic von Neumann winner policy and derive corresponding performance guarantees. Under these models, our results provide a unified finite-sample solution to a central challenge in pluralistic alignment: learning from limited, heterogeneous, and potentially cyclic feedback, and producing a single policy with explicit collective-welfare guarantees. Our framework thereby makes preference aggregation an explicit and statistically analyzable design choice rather than an implicit consequence of pooling human feedback.
\end{abstract}

\section{Introduction}

Pluralistic alignment requires learning from human feedback that may reflect persistent and potentially conflicting preferences while ultimately selecting a collective policy. Reinforcement Learning with Human Feedback (RLHF) has become a widely used approach for this broader alignment problem, particularly in language-model post-training and sequential decision making \citep{NIPS2017_d5e2c0ad,ouyang2022training,ziegler2019fine}. Standard RLHF methods typically learn a single reward model from preference data, often collected as pairwise comparisons, and then optimize a policy against that reward. This pipeline implicitly assumes that feedback from different users can be represented by a common reward function.

This \emph{single-party} assumption is poorly suited to settings in which users have heterogeneous and potentially conflicting viewpoints. In open-ended social decision-making, variation in feedback may reflect genuine disagreement rather than noise around one latent objective \citep{casper2023open,santurkar2023whose,siththaranjan2024distributional}. Pooling the resulting comparisons therefore discards information about which parties hold which preferences. It also makes a consequential normative decision implicitly: the statistical procedure used to fit the pooled reward determines how conflicting preferences are combined.

A preliminary version of this manuscript was publicly circulated in March 2024 and was among the first theoretical studies to formulate heterogeneous human feedback as an explicit multi-party RLHF problem. Our work inspired subsequent research that has generalized, operationalized, or directly analyzed components of this formulation. Specifically, rather than treating disagreement as statistical noise, resolving it solely through personalization, or assuming that the relevant rewards are already known, we study how to learn unknown party-specific preferences from offline comparisons and use them to select one collective policy. This formulation makes both the statistical cost of preserving heterogeneity and the normative choice of aggregation rule explicit.

The problem lies at the intersection of three areas that are commonly treated separately: offline preference learning, shared-representation learning across related parties, and social-choice aggregation. Existing offline RLHF theory largely analyzes the estimation and optimization of a single reward function \citep{zhan2024provable,zhu2023principled}. Personalized approaches learn different rewards or policies for different users \citep{park2024rlhf,liu2025shared,lee2024low}, whereas social-choice and policy-aggregation approaches often begin with preferences, rewards, or policies that are already specified \citep{NEURIPS2024_9328208f,NEURIPS2024_7e670825}. Our setting is distinct: party identities are observed, party-specific preferences are unknown and must be learned from limited offline comparisons, and the final output must be one collective policy rather than a collection of personalized policies.

To see why this distinction matters, consider the following example.

\paragraph{Example (2 parties and 3 alternatives).}
Consider three alternatives $A,B,C$. Half of the population prefers $A \succ B \succ C$, with rewards
\[
R_1(A)=1,\qquad R_1(B)=1/2,\qquad R_1(C)=0,
\]
while the other half prefers $C \succ B \succ A$, with rewards
\[
R_2(C)=1,\qquad R_2(B)=1/2,\qquad R_2(A)=0.
\]
When the party labels are discarded and the two sub-populations are pooled, the population-level pairwise comparisons cannot distinguish among $A,B,C$. A single reward model fitted to the pooled comparisons therefore cannot identify $B$ as the compromise. Finite-sample variation or arbitrary tie-breaking may instead select $A$ or $C$, leaving one party with reward $0$. By retaining the party labels, learning the two rewards separately, and then applying Nash or Egalitarian aggregation, the collective policy selects $B$ and gives both parties reward $1/2$. Thus, the difficulty is not only estimating preferences accurately; it is preserving the structure of disagreement until an explicit aggregation rule is chosen.

This example motivates the central question of the paper:
\begin{center}
    \textit{How can one learn a single collective policy from offline, party-indexed comparisons without first collapsing heterogeneous feedback into one implicit reward model?}
\end{center}

\paragraph{Scope of pluralistic alignment.} We use pluralistic alignment in a collective-decision sense. Our objective is not to require every model response to display multiple viewpoints, nor to produce a separate personalized policy for each user. Instead, we retain observed party-specific preferences during learning, quantify their statistical uncertainty separately, and select one collective policy through an explicitly specified social-welfare or pairwise social-choice rule. Our formulation therefore studies the statistical foundations of collective preference aggregation, complementing Overton, steerable, and distributional approaches that focus on the diversity, controllability, or population calibration of model outputs. We take the set of parties, their normative weights, and the admissibility of their preferences as externally specified.

\paragraph{Our approach and technical novelty.}
As one of the earliest theoretical frameworks for multi-party RLHF, our work connects offline preference learning, shared-representation learning, and social-welfare optimization within a single finite-sample framework. We explicitly separate two coupled problems. The first is statistical: learning party-specific preferences from limited and unevenly covered offline feedback while exploiting structure shared across parties. The second is normative: selecting a social-welfare rule that converts the learned preferences into a single collective policy. This separation makes preference aggregation an explicit design choice rather than an incidental consequence of pooling the data.

More specifically, to our knowledge, this work provides the first offline finite-sample analysis that jointly learns observed party-specific reward models through an unknown shared low-rank representation and performs pessimistic optimization of nonlinear social-welfare objectives. The novelty is therefore not merely the application of a welfare function after reward learning. The party rewards are themselves unknown and statistically coupled, and uncertainty from their estimation must be propagated through the nonlinear aggregation rule to the final policy. This creates a different statistical problem from both single-reward offline RLHF and social-welfare optimization with known rewards.

We first study a reward-representable regime. Each party follows a linear Bradley--Terry--Luce preference model, and the party-specific reward parameters share an unknown low-rank representation. This shared structure allows information from different parties to be used jointly without pooling away their distinct reward functions. We estimate the party rewards simultaneously, construct party-specific confidence sets, and optimize the resulting pessimistic social welfare. The same statistical framework supports Utilitarian, Nash, and Egalitarian objectives, thereby separating the choice of collective-welfare criterion from the statistical problem of reward estimation.

We then consider a substantially more general regime in which pairwise preferences need not admit any scalar reward representation and may be stochastic, nontransitive, or cyclic. In this setting, reward maximization is no longer an appropriate solution concept. We instead estimate party-indexed pairwise-preference matrices and construct a pessimistic von Neumann winner directly from the offline comparisons. The resulting framework therefore covers both utility-representable preferences and general pairwise preferences under a common principle: preserve heterogeneous feedback, quantify its statistical uncertainty, and apply an explicit collective-decision rule.

We formulate the main problem as a contextual bandit and extend the framework to trajectory-level preference feedback in Markov decision processes. Our results characterize how the difficulty of collective alignment depends jointly on the shared reward structure, the number of parties, and the coverage of the offline comparison data.

Our main contributions are summarized as follows:
\begin{itemize}
    \item \textbf{An end-to-end framework for pluralistic offline RLHF.}
    We formulate collective policy learning from party-indexed offline comparisons and explicitly separate preference learning from preference aggregation. Unlike personalization, our objective is a single collective policy; unlike conventional social-choice aggregation, the underlying party preferences are unknown and must be estimated from data.

    \item \textbf{Finite-sample theory for learning and aggregating heterogeneous rewards.}
    Under a shared low-rank linear reward model, we jointly estimate party-specific rewards, construct confidence sets, and optimize pessimistic Utilitarian, Nash, and Egalitarian welfare objectives. We establish nonasymptotic bounds for party-specific reward estimation and for the suboptimality of the resulting collective policy under realizability, diversity, and offline-coverage conditions. The analysis quantifies how preference-estimation uncertainty propagates through nonlinear social-welfare optimization.

    \item \textbf{Collective alignment beyond scalar reward models.}
    For general and potentially cyclic pairwise preferences, we construct a pessimistic von Neumann winner \citep{dudik2015contextual,fishburn1984probabilistic} directly from offline party-indexed comparisons. We establish finite-sample performance guarantees and, under additional structural conditions, analyze ex-post efficiency. This extension addresses settings in which heterogeneous feedback cannot be reduced to the maximization of party-specific scalar rewards.
\end{itemize}

\paragraph{Historical positioning and subsequent developments.}
A preliminary version of this manuscript was publicly circulated in March 2024 and was among the first theoretical treatments of multi-party RLHF. At that time, only a small number of contemporaneous studies had begun to investigate heterogeneous preferences in RLHF \citep{chakraborty2024maxmin,park2024rlhf}. Our broad historical claim is that this work was among the earliest contributions to the emerging theory of multi-party and pluralistic RLHF. Our more specific priority claim is that, to our knowledge, it was the first to provide finite-sample guarantees for the combination of shared low-rank party-specific reward learning, pessimistic nonlinear social-welfare optimization, and offline policy selection. It also introduced a multi-party offline von Neumann formulation for preferences that need not admit scalar reward representations.

Subsequent research has generalized, operationalized, or directly analyzed components of this formulation. \citet{lee2024low} extend the predefined-party model to contextual low-rank user heterogeneity; \citet{xiong2025projection} develop an efficient projection framework addressing the computational cost of nonlinear multi-objective and multi-group aggregation; and \citet{buening2025strategyproof} directly analyze the incentive properties of pluralistic RLHF procedures, including the pessimistic social-welfare rule studied here. Other work has developed axiomatic reward aggregation and policy-level social choice \citep{NEURIPS2024_9328208f,NEURIPS2024_7e670825}. These developments demonstrate that our early formulation anticipated several statistical, computational, and strategic questions that have since become central to pluralistic alignment, while also clarifying the scope of our guarantees. In particular, our analysis assumes observed, persistent party identities and non-strategic feedback under the stated preference-model and offline-coverage conditions.

\begin{figure}[ht]
    \centering
    \begin{minipage}[t]{0.66\textwidth}
        \centering
        \includegraphics[width=0.99\textwidth]{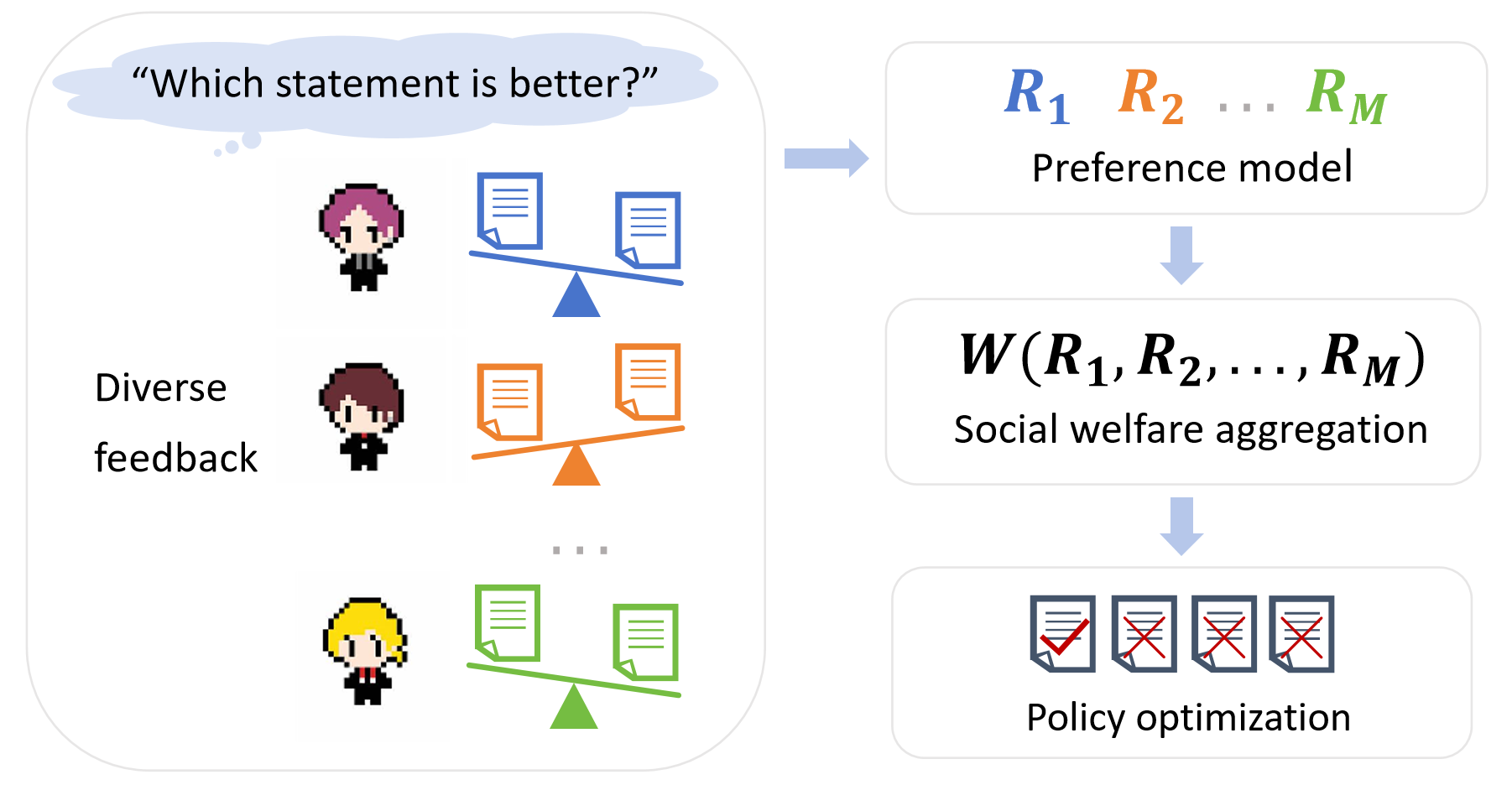}
        \subcaption{}
        \label{fig:framework}
    \end{minipage}
    \begin{minipage}[t]{0.33\textwidth}
        \centering
        \includegraphics[width=0.99\textwidth]{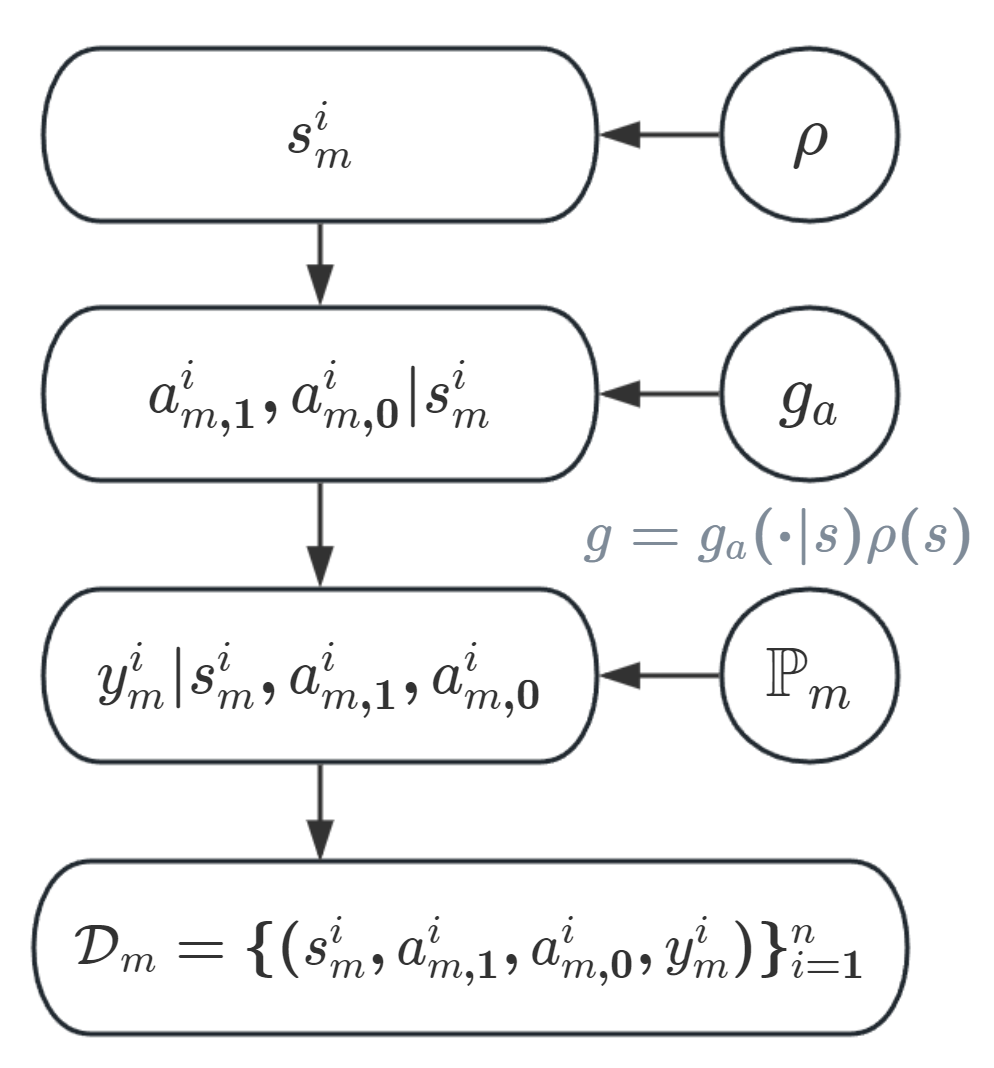}
        \subcaption{}
        \label{fig:data}
    \end{minipage}
    \caption{Multi-party RLHF: (a) party-specific preferences are learned and combined through an explicit social-welfare rule; (b) the offline feedback-collection process retains the party associated with each comparison.}
\end{figure}

\subsection{Related Work}

\paragraph{Offline RLHF theory.}
Preference comparisons are widely used in reinforcement learning because they are often easier for humans to provide than numerical rewards \citep{pmlr-v162-chen22ag,ouyang2022training,ziegler2019fine}. The theoretical foundations most closely related to our offline setting include \citet{zhu2023principled}, who study pessimistic offline RLHF under a linear Bradley--Terry--Luce (BTL) reward model, and \citet{zhan2024provable}, who extend the analysis to general function approximation. \citet{wang2023rlhf} analyze reward-based and general pairwise-preference models in online single-party RLHF. These works primarily study feedback generated by a single latent preference model. In contrast, our analysis retains observed party identities, jointly estimates multiple statistically related reward models, and propagates their uncertainty through a collective social-welfare objective. The resulting problem combines preference estimation and preference aggregation rather than treating them as separate stages with a single reward.

\paragraph{Heterogeneous and pluralistic RLHF.}
A growing literature studies how RLHF systems should respond to heterogeneous human feedback. Empirical evidence demonstrates that human preferences can vary systematically across demographic, cultural, and contextual dimensions \citep{santurkar2023whose,siththaranjan2024distributional}. \citet{bakker2022fine} use isoelastic social-welfare functions to fine-tune language models toward consensus among heterogeneous preferences. \citet{chakraborty2024maxmin} learn a mixture of reward models and apply an Egalitarian objective to align a language model with diverse latent preferences. \citet{park2024rlhf} study both personalization and collective aggregation, proposing representation-learning and clustering methods for personalized rewards as well as reward- and preference-level aggregation mechanisms.

Our work is distinguished by the specific combination of observed persistent parties, shared low-rank reward estimation, offline statistical uncertainty, nonlinear social-welfare optimization, and a single collective policy. Unlike personalization, we do not produce a different policy for each user. Unlike latent-mixture approaches, party membership is observed in the feedback data and carries normative meaning in the subsequent aggregation step. More specifically, our contribution is an end-to-end finite-sample analysis of how uncertainty in jointly estimated party rewards propagates through pessimistic Utilitarian, Nash, and Egalitarian policy optimization.

\paragraph{Shared representation and personalized preference learning.}
Our reward-estimation method is related to multi-task and representation learning, where information from multiple related tasks is used to estimate a shared low-dimensional structure \citep{baxter2000model,maurer2016benefit}. Finite-sample guarantees for shared-representation linear models have been developed by \citet{du2021few,Tripuraneni2020ProvableMO,li2023multi}, with related extensions to offline multi-task reinforcement learning \citep{bose2024offline}. In heterogeneous-feedback learning, \citet{liu2025shared} use shared low-rank adaptation to learn personalized reward models, while \citet{lee2024low} model low-rank interactions between user contexts and query--response features and study distribution shift.

These approaches primarily target personalization or continuously varying user contexts. Our predefined parties can be viewed as a discrete and observed form of heterogeneity, but our decision problem is different: the objective is to learn one collective policy under a specified social-welfare rule. Accordingly, our analysis must control not only party-specific estimation error but also its effect on the nonlinear aggregation of party values.

\paragraph{Social choice, welfare aggregation, and multi-objective RL.}
Social choice theory studies how individual preferences or utilities should be combined into collective decisions. Relevant solution concepts include social-welfare functions \citep{moulin2004fair}, Pareto efficiency \citep{aumann2010pareto,barman2018finding}, von Neumann winners or maximal lotteries \citep{brandl2016consistent,dudik2015contextual,rivest2010optimal}, and ex-post efficiency \citep{fishburn1984probabilistic,zeng2020fairness}. \citet{fish2023generative} connect generative AI with approval-based social choice. \citet{NEURIPS2024_9328208f} study axioms for aggregating pairwise human feedback into a collective linear reward, showing that standard statistical reward-learning procedures can violate natural social-choice properties. \citet{NEURIPS2024_7e670825} instead aggregate already-defined policies through the geometry of state--action occupancy measures.

Related work in multi-agent and multi-objective RL studies collective policies when the relevant reward functions or objectives are already specified. \citet{ogryczak2013compromise,pmlr-v119-siddique20a} optimize generalized-Gini objectives, while \citet{mandal2022socially,ju2024achieving} study broader families of social-welfare functions, including Utilitarian, Nash, and Egalitarian objectives, with online or offline guarantees. Multi-objective alignment methods likewise balance several specified objectives through reward, parameter, or policy combinations \citep{siddique2023fairness,wang2024arithmetic,zhou2023beyond,rame2023rewarded}.

Our setting differs from these formulations because the party-specific rewards are not given: they must first be inferred from offline comparisons. Consequently, the confidence sets in our analysis capture uncertainty arising from preference learning, and the final policy must remain reliable after this uncertainty is propagated through the chosen social-welfare function. The novelty lies in coupling statistical reward learning with social-welfare optimization, rather than introducing either component in isolation.

\paragraph{Preferences beyond scalar reward models.}
Several recent approaches avoid assuming that pairwise preferences can be represented by a scalar reward. \citet{pmlr-v235-munos24a} formulate Nash Learning from Human Feedback as a two-player preference game, while \citet{pmlr-v235-swamy24a} develop a self-play method based on a minimax winner that accommodates stochastic, nontransitive, and non-Markovian preferences. These methods demonstrate the importance of solution concepts defined directly from pairwise comparisons.

Our general pairwise-preference setting shares this motivation but addresses a different statistical problem. We observe offline, party-indexed comparisons and seek a single collective decision under uncertainty. We aggregate the party-specific pairwise preference matrices and construct a pessimistic von Neumann winner, obtaining finite-sample guarantees without requiring each party's preferences to admit a scalar reward representation.

\paragraph{Subsequent extensions and limitations.}
Subsequent work has generalized or operationalized specific components of our formulation. \citet{lee2024low} extend predefined parties to contextual low-rank user heterogeneity. \citet{xiong2025projection} develop a computationally efficient projection framework for nonlinear multi-objective and multi-group aggregation, addressing the cost of repeatedly optimizing different nonlinear welfare functions. \citet{pmlr-v267-wang25af} propose an efficient post-processing approach that combines policies already trained for different preferences. These methods complement our focus on reward estimation, offline uncertainty, and finite-sample welfare guarantees.

Other work clarifies limitations of the underlying modeling assumptions. \citet{buening2025strategyproof} show that pluralistic RLHF procedures, including pessimistic social-welfare optimization, need not be strategyproof and analyze the trade-off between truthful elicitation and collective welfare. \citet{xiao2024algorithmic} show that downstream KL-regularized policy optimization can produce preference collapse even when reward information is available. Active data-collection approaches select informative prompts or feedback providers rather than treating the offline dataset as fixed \citep{liu2024dual}. These developments clarify the scope of our results: we study the offline statistical problem with observed party identities and non-strategic feedback, while incentive compatibility, active elicitation, contextual personalization, and scalable policy optimization constitute complementary directions.

\section{Preliminaries}\label{sec:preliminaries}
\paragraph{Notation.} Let $[n]=\{1,2\cdots,n\}$. Let $a\lor b=\max\{a,b\}$. We use $\|\cdot\|_2$ or $\|\cdot\|$ to denote the $\ell_2$ norm of a vector or the spectral norm of a matrix, and $\|x\|_\Sigma=\sqrt{x^{\top}\Sigma x}$ to denote the matrix-induced norm for a positive-semidefinite matrix $\Sigma$. We use $\mathcal{O}_{d_1\times d_2}$ where $d_1>d_2$ to denote the set of orthogonal matrices (i.e., the columns are orthonormal). Let $\langle\cdot,\cdot\rangle$ be the Euclidean inner product between vectors or matrices. 
For a matrix $A\in\mathbb R^{m\times n}$, let $\sigma_i(A)$ be its $i$-th largest singular value.
Our use of $O(\cdot),\Omega(\cdot),\Theta(\cdot)$ follows the standard notation.


\subsection{Offline Data Collection and Comparison Model}\label{sec:predata}  

We ground our theoretical investigation of RLHF in the contextual bandit (CB) framework \citep{xiong2023iterative,zhu2023principled}. This formulation naturally models the reinforcement learning dynamics of LLMs, where the generation of a completion $a \in \mathcal{A}$ given a prompt $s \in \mathcal{S}$ is evaluated by a reward model $R_\theta(s,a)$ \citep{chakraborty2024maxmin,ouyang2022training,xiong2023iterative,zhu2023principled,ziegler2019fine}. By operating within the CB setting, which is a foundational case of MDPs with a horizon $H=1$ and no transitions, we can isolate and rigorously analyze the fundamental complexities of multi-party preference aggregation. We subsequently extend our theoretical framework to general MDPs, demonstrating that these insights naturally generalize to sequential settings. Formally, we define a multi-party CB environment by the tuple $E=(\mathcal{S},\mathcal{A},\{R_m\}_{m=1}^M,\rho)$, where $\mathcal{S}$ is the state space (e.g., prompts), $\mathcal{A}$ is the action space (e.g., responses), $M$ is the total number of parties, $R_m:\mathcal{S}\times\mathcal{A}\rightarrow \mathbb R$ denotes the reward function specific to the $m$-th party, and $\rho\in \Delta(\mathcal{S})$ specifies the initial state distribution. A policy maps a state to an action, taking the form of either a deterministic mapping $\pi:\mathcal{S}\rightarrow\mathcal{A}$ or a randomized distribution $\pi:\mathcal{S}\rightarrow \Delta(\mathcal{A})$. Although the reward functions are indexed by party, the learner's output is a single collective policy \(\pi\) aligned with respect to all parties' preference, rather than a collection of personalized policies \(\{\pi_m\}_{m=1}^M\).

We focus on the offline learning paradigm, which is an approach extensively explored in previous research on offline RLHF \citep{zhu2023principled, zhan2024provable}, driven by the practical constraints of LLM alignment:  Collecting high-quality human feedback is resource-intensive and time-consuming, which typically precludes online, interactive human queries during the training phase. Consequently, models have to be aligned using pre-collected datasets. We assume access to an offline dataset $\mathcal{D}=\bigcup_{m=1}^M\mathcal{D}_m$, where $\mathcal{D}_m=\{(s_{m}^i, a_{m,1}^i, a_{m,0}^i,y_{m}^i)\}_{i=1}^{n}$ represents the pairwise comparison data specific to the $m$-th party.
Specifically, the state $s_m^i$ represents a prompt and is drawn from the initial distribution $\rho$, the action pair $(a_{m,1}^i,a_{m,0}^i)$ represents two candidate responses, and the outcome $y_m^i \in \{0,1\}$ is the party's binary preference label provided by an annotator. As illustrated in Figure \ref{fig:data}, the data generation process follows a structured sequence. In the language-model setting, the index \(m\) identifies a \emph{party}: 
a persistent preference source whose preferences are observed across prompts. 
A party may be an individual or a stakeholder group represented by multiple 
feedback providers. For each party \(m\), the dataset \(\mathcal D_m\) contains 
comparison observations indexed by \(i\). Specifically, each time a prompt 
\(s_m^i\) is drawn from the prompt distribution \(\rho\), and two candidate 
completions, \(a_{m,1}^i\) and \(a_{m,0}^i\), are generated for that prompt. 
An 
\emph{annotator}, or \emph{feedback provider}, supplies comparison label \(y_m^i\) recording which completion is preferred by this party. When a party represents a stakeholder group, multiple feedback 
providers within this group may contribute observations to the same party-specific dataset 
\(\mathcal D_m\).
 

\subsection{Social Welfare Function}\label{sec:preswf}

Having specified how party-indexed preference comparison data are collected, we next describe how the learned party preferences are used to select a collective policy. In the reward-representable setting, each party is associated with a reward function, and a policy may provide different values to different parties. Social welfare functions provide a systematic way to aggregate these party-specific values into a single objective for collective policy selection. Specifically, for a policy $\pi$,  let $     V_m(\pi)    =\mathbb{E}_{{s\sim\rho,a\sim\pi(\cdot\mid s)}}
    \left[R_m(s,a)\right] $
denote its expected value to party $m$. Each policy
therefore induces a vector of party values
\[
    V(\pi)
    =
    \bigl(V_1(\pi),\ldots,V_M(\pi)\bigr).
\]

A social welfare function maps this vector to a scalar measure of collective welfare. Different welfare functions express different approaches to balancing the interests of the parties and may therefore lead to different collective policies. In our framework, the offline comparison dataset is used to learn the party-specific reward and thus estimate the expected values for each party, and the selected welfare function determines how those values are aggregated. This structure separates the statistical
task of preference learning from the normative choice of collective aggregation.

We consider the isoelastic family of social welfare functions \citep{moulin2004fair,sen2018collective}. For a positive value vector $v=(v_1,\ldots,v_M)\in\mathbb{R}_{+}^M$, this family is defined as
\begin{equation}\label{eq:swfWa}
    W_\alpha(v)
    =
    \begin{cases}
        \displaystyle
        \left(
            \frac{1}{M}\sum_{m=1}^{M}v_m^\alpha
        \right)^{1/\alpha},
        & \alpha\leq 1,\quad \alpha\neq 0, \\[8pt]
        \displaystyle
        \left(
            \prod_{m=1}^{M}v_m
        \right)^{1/M},
        & \alpha=0.
    \end{cases}
\end{equation}
Its axiomatic characterization and additional properties are reviewed in
Appendix~\ref{ap:swfdef}. In this article, we primarily focus on three representative members of this
family: Utilitarian, Egalitarian, and Nash welfare.

\paragraph*{Utilitarian welfare.}
The Utilitarian objective corresponds to $\alpha=1$ and maximizes the average
value across parties:
\[
    J_{\mathrm{U}}(\pi)
    =
    \frac{1}{M}\sum_{m=1}^{M}V_m(\pi).
\]
Under this objective, an improvement to any party contributes equally to collective welfare. The Utilitarian objective therefore emphasizes aggregate party value, although a high average value may coexist with a low value for one or more parties.

\paragraph*{Egalitarian welfare.}
The Egalitarian, or max-min, objective is obtained in the limit as
$\alpha\to-\infty$:
\[
    J_{\mathrm{E}}(\pi)
    =
    \min_{m\in[M]}V_m(\pi).
\]
This objective evaluates a policy according to the value it provides to the
worst-off party. Maximizing it therefore prioritizes improving the lowest
party value. 

\paragraph*{Nash welfare.}
The Nash objective corresponds to $\alpha=0$ and maximizes the geometric mean
of the party values:
\[
    J_{\mathrm{N}}(\pi)
    =
    \left(
        \prod_{m=1}^{M}V_m(\pi)
    \right)^{1/M}.
\]
The Nash objective rewards increases in aggregate welfare while remaining sensitive to parties with relatively low values. It therefore balances aggregate welfare and inequality across parties rather than focusing exclusively on either the average value or the worst-off party
\citep{Kelly1998,zhang2022proportional}.

These objectives assign equal normative weight to the $M$ parties. This equal weighting is a modeling and normative choice.  Other externally specified party weights are
possible, but we do not analyze them here.

In Section~\ref{sec:rwd}, we use Nash welfare as the primary example and subsequently apply the same reward-estimation and pessimistic-optimization framework to the Utilitarian and Egalitarian objectives in Section~\ref{ap:moreswf}. In the general pairwise-preference setting considered in Section~\ref{sec:ext}, scalar party values may not exist, so we instead define the collective solution directly from the party-specific
pairwise-preference matrices.

\section{Reward-Representable Regime}\label{sec:rwd}

Having introduced the collective welfare objectives, we now study how the party-specific rewards entering these objectives can be learned from offline preference comparisons. We instantiate the social welfare perspective from Section~\ref{sec:preliminaries} in a reward-representable regime, where each party’s preferences are represented by a scalar reward function. This reward-based formulation is the standard modeling approach in RLHF for LLMs~\citep{ouyang2022training,ziegler2019fine,NIPS2017_d5e2c0ad}. In this regime, collective alignment requires learning the party-specific rewards and using them to select a single collective policy according to an explicit social-welfare objective.

\paragraph*{Reward-based Model}
We begin by specifying the reward-based model and how each party's latent reward function determines the pairwise comparison data observed by the learner. Suppose that the preference distribution of party $m$ is induced by a reward function:
\begin{equation}\label{eq:rwdpro}
   \mathbb P_m(y=1|a_{1},a_{0},s)=\Phi(R_{m}(s,a_1)-R_{m}(s,a_0)), \quad \forall m\in [M],
\end{equation}
where $y=1$ indicates that action $a_1$ is preferred to action $a_0$ at state $s$, and $\Phi$ is the sigmoid function $\Phi(x)=1/(1+\exp(-x))$. This gives the well-known Bradley--Terry--Luce (BTL) pairwise comparison model~\citep{bradley1952rank}, which is widely used in RLHF frameworks~\citep{NIPS2017_d5e2c0ad,ouyang2022training}. 

Fix a constant $B>0$ and define $\mathbf{\Theta}_B:=\{\theta\in \mathbb{R}^{d}:\|\theta\|_{2}\leq B\}$. Following \cite{zhu2023principled}, we assume that the underlying reward $R_m$ for each party belongs to the family of linear functions $R_{\theta}(s,a)=\theta^{\top} \phi(s,a) + C_{0}$, where $\phi(s,a):\mathcal{S}\times\mathcal{A}\rightarrow \mathbb{R}^d$ is a known feature mapping with $\max_{s,a}\|\phi(s,a)\|_2\leq L$ and $C_{0}>BL$ is a fixed positive constant. Specifically, we assume the true reward of party $m$ is $R_{m}=R_{\theta_{m}^*}$, where  $\theta_m^*\in\mathbf{\Theta}_B$ denotes the ground truth parameter for the $m$-th party reward function. The constant $C_{0}$ cancels from every BTL comparison and therefore does
not affect the induced preference probabilities. Its role is to specify the fixed reward normalization used by the welfare objective. In particular,  
for every $\theta\in \mathbf{\Theta}_{B}$ and every policy $\pi$,
\begin{equation*}
    \mathbb{E}_{s\sim\rho,a\sim \pi(s)}
    [R_{\theta}(s,a)]
    \ge
    C_0-BL>0.
\end{equation*}
Thus, all policy value functions appearing in welfare objective are uniformly positive. We remark that different choices of $C_0$ may induce different welfare-maximizing policies, and all guarantees below are conditional on this fixed normalization. In LLM applications, \(\phi(s,a)\) may be instantiated as the representation of a prompt-response pair produced by a pretrained model before its final scalar output layer. The parameter \(\theta_m^*\) then serves as the $m$-th party linear reward head applied to this shared representation.

We further assume that the party-specific reward parameters share a common low-rank structure:  $$\theta_m^*=U^*\alpha_m^*,$$ where \(U^*\in\mathcal O_{d\times r}\), \(d\gg r\), has orthonormal columns spanning a shared \(r\)-dimensional subspace of the original parameter space, and \(\alpha_m^*\in\mathbb R^r\) is the party-specific coefficient vector satisfying
$ \|\alpha_m^*\|_2=\Theta(1).$ Equivalently,
\[
R_m(s,a)
=
{\alpha_m^*}^{\top}U^{*\top}\phi(s,a)+C_0.
\]Thus, \(U^{*\top}\phi(s,a)\) provides a shared low-dimensional representation of the state-action features, while \(\alpha_m^*\) captures how party \(m\) weights the resulting reward-relevant directions. This shared structure allows information from different parties to be used jointly without pooling their distinct reward functions into a single reward model. 
This low-rank structure is a standard assumption in shared-representation learning and underlies the estimation error reduction established later in this section
\citep{du2021few,Tripuraneni2020ProvableMO}.

Under this model, our framework proceeds in two stages. First, we jointly estimate the shared representation and the party-specific reward parameters from the offline comparisons and construct party-specific confidence sets that quantify their statistical uncertainty. Second, we select a single collective policy by optimizing a pessimistic Nash welfare objective over these confidence sets. The remainder of this section analyzes the resulting estimation error, policy suboptimality, and welfare guarantees before extending the framework to other social-welfare objectives and trajectory-level preference feedback in MDPs. Throughout this section, we restrict attention to the deterministic policy class. Accordingly, all optimality, suboptimality, and efficiency guarantees in this section are relative to the best deterministic policy.


\subsection{Pessimistic Nash Welfare Algorithm}\label{sec:swfpes}

Building on the reward-based model introduced above, we now present a two-step procedure for learning a deterministic collective policy from offline party-indexed comparisons. In the first step, we jointly estimate the shared representation and the party-specific reward parameters and construct confidence sets that quantify their statistical uncertainty. In the second step, we optimize a pessimistic Nash welfare objective over these confidence sets to obtain the final deterministic policy. Specifically, the pessimistic objective evaluates each candidate policy under the least favorable combination of party-specific reward parameters that remains plausible given the observed comparisons. This construction guards policy selection against estimation error arising from limited offline coverage. The complete procedure is summarized in Algorithm~\ref{alg1}. 

\begin{algorithm}[ht]
    \caption{Pessimistic Nash Welfare}\label{alg1}
	\textbf{Input:} The datasets $\mathcal{D}=\bigcup_{m=1}^M\mathcal{D}_m$, a failure probability $\delta$, and the initial distribution $\rho$.

    \textbf{Output:} The deterministic collective policy \(\widehat\pi\).
    
	\begin{algorithmic}[1]
        \STATE Jointly compute the MLE \((\widehat U,\widehat\alpha)\) in Equation 3 and set \(\widehat\theta_m=\widehat U\widehat\alpha_m\) for every \(m\in[M]\). 
		\STATE Construct the party-specific confidence sets $\mathbf{\Theta}_B(\hat{\theta}_m)=\{\theta\in\mathbf{\Theta}_B:\|\theta-\hat{\theta}_m\|_{\Sigma_m}\leq \Gamma(\delta,n,M,d,r)\}$ according to the bound $\Gamma$ on the estimation error established in Theorem \ref{th:metatheta}.
		\STATE Establish the pessimistic Nash welfare function in \eqref{eq:pesnash}.
        \STATE Calculate $\hat{\pi}=\arg\max_{\pi}\hat{J}(\pi)$.
	\end{algorithmic}  
\end{algorithm}

\textbf{Step 1: MLE and confidence set construction.}  
The objective of this first step is to estimate the party-specific reward functions from the observed comparisons while quantifying their statistical uncertainty. We jointly estimate the shared representation and the party-specific coefficients using all party-indexed datasets. 
To simplify the notation, denote $\alpha:=(\alpha_1,\cdots,\alpha_M)\in\mathbb R^{r\times M}$ as the low-dimensional parameter matrix, and $X_{m,i}=\phi(s_m^i,a_{m,1}^i)-\phi(s_m^i,a_{m,0}^i)$ as the feature difference vectors for $\forall i\in[n], m\in [M]$. Using the comparison dataset $\mathcal{D}=\bigcup_{m=1}^M\mathcal{D}_m$, we minimize the negative log-likelihood based on \eqref{eq:rwdpro}, which is given by
\begin{align}\label{eq:mle}
   \hat{U},\hat{\alpha} \in \arg\min_{U\in\mathcal{O}_{d\times r}, \|\alpha_m\|_2\leq B}l_{\mathcal{D}}(U,\alpha),
\end{align}
   $$l_{\mathcal{D}}(U,\alpha)=-\dfrac{1}{Mn}\sum_{m=1}^M\sum_{i=1}^{n}\Big\{\mathbf{1}(y_m^i=1)\log\big(\Phi(\langle U\alpha_m,X_{m,i}\rangle)\big)
   +\mathbf{1}(y_m^i=0)\log\big(\Phi(\langle U\alpha_m,-X_{m,i}\rangle)\big)\Big\}.$$
When the minimizer is not unique, we take any of the solutions $(\hat U,\hat\alpha)$ that achieves the minimum. The resulting estimator of party \(m\)’s reward parameter is
\(
\widehat\theta_m=\widehat U\widehat\alpha_m.
\)
Joint estimation allows comparisons from all parties to inform the shared representation \(\widehat U\), while retaining a distinct coefficient vector \(\widehat\alpha_m\) for each party.

We next construct a confidence set around each estimated reward parameter \(\widehat\theta_m\) to quantify uncertainty about the corresponding true parameter \(\theta_m^*\). Specifically, we define: 
\begin{equation*}
   \mathbf{\Theta}_B(\hat{\theta}_m)=\{\theta\in\mathbf{\Theta}_B:\|\theta-\hat{\theta}_m\|_{\Sigma_m}\leq \Gamma(\delta,n,M,d,r)\},
\end{equation*}
where $\Gamma$ is a confidence radius defined in the next subsection, $\delta$ is a prescribed failure probability, and $\Sigma_m\in \mathbb R^{d\times d}$ is the empirical covariance matrix for the $m$-th party, defined as $\Sigma_m:=\sum_{i=1}^nX_{m,i} X_{m,i}^\top/n$. For simplicity, we define $\Theta=(\theta_1,\cdots,\theta_M),\hat{\Theta}=(\hat{\theta}_1,\cdots,\hat{\theta}_M)$, and use $\Theta\in\mathbf{\Theta}_B(\hat{\Theta})$ to denote the event of $\theta_m\in\mathbf{\Theta}_B(\hat{\theta}_m)$ for any $m\in[M]$.  

\textbf{Step 2: Policy optimization under Nash welfare.}  Building on the confidence sets constructed in Step 1, we now select a deterministic collective policy by optimizing the Nash welfare objective while accounting for uncertainty in the learned rewards. For a candidate parameter \(\theta_m\), party \(m\)’s value under policy \(\pi\) is
$
\mathbb E_{s\sim\rho}
\left[R_{\theta_m}(s,\pi(s))\right].$
We aggregate these \(M\) party-specific values using the Nash welfare objective introduced in Section~\ref{sec:preswf}.

Specifically, we account for reward-estimation uncertainty by defining the following pessimistic Nash welfare objective:
\begin{equation}\label{eq:pesnash}
   \hat{J}(\pi)=\min_{\Theta\in\mathbf{\Theta}_B(\hat{\Theta})}\prod_{m=1}^M[\mathbb E_{s\sim\rho} R_{{\theta}_m}(s,\pi(s))]^{1/M}.
\end{equation}
For each candidate policy, the inner minimization evaluates its Nash welfare under the least favorable combination of party-specific reward parameters contained in the confidence sets. It therefore prevents the policy optimizer from relying on optimistic reward estimates that are weakly supported by the offline comparisons. In particular, policies whose values depend strongly on poorly covered feature directions can receive a lower pessimistic welfare value.
We obtain the final deterministic collective policy by maximizing this pessimistic objective: $$\hat{\pi}=\arg\max_{\pi}\hat{J}(\pi).$$ Here, pessimism specifically addresses statistical uncertainty arising from finite and imperfectly covered offline data \citep{li2022pessimism,xie2021policy,zhan2024provable,zhu2023principled}.

\subsection{Sub-optimality}\label{sec:swfcomp}
We now establish finite-sample guarantees for the deterministic collective policy returned by Algorithm 1. Our analysis proceeds in three parts. First, we bound the error of the joint shared-representation estimator for each party’s reward parameter. Second, we propagate this estimation error through the pessimistic Nash welfare objective to obtain a policy-suboptimality bound. Finally, we provide a minimax lower bound that identifies an unavoidable dependence on the representation dimension and clarifies the remaining gap in our upper bound. Together, these results characterize how shared reward structure and offline data coverage affect the quality of the learned collective policy.
Formally, we define the true Nash welfare function $J$ and the sub-optimality of a deterministic policy ${\pi}$ as follows:
\begin{align*}
    &J({\pi}):=\prod_{m=1}^M [\mathbb E_{s\sim\rho} R_{m,\theta_m^*}(s,{\pi}(s))]^{1/M},\quad \texttt{SubOpt}(\pi):=J(\pi^*)-J(\pi),
\end{align*}
where $\pi^*=\arg\max_{\pi}J(\pi)$ is the optimal deterministic policy.  
Here, \(J(\pi)\) is the true Nash welfare evaluated using the ground-truth party rewards, and \(\pi^*\) is an optimal deterministic policy under this objective. The suboptimality measures the welfare loss of a candidate policy relative to this benchmark. Our goal is to show that the pessimistic policy \(\widehat\pi\) has small suboptimality when the shared reward structure is identifiable and the offline comparisons provide sufficient coverage. 
We introduce several assumptions first.
\begin{assumption}[Diversity design]\label{as:well}
    Let $\Theta^*=(\theta_1^*,\cdots,\theta_M^*)\in\mathbb R^{d\times M}$ be the original parameter matrix, $\nu=\sigma_r(\frac{\Theta^{*\top} \Theta^*}{M})=\sigma_r(\frac{\alpha^{*\top} \alpha^*}{M})$ be the smallest singular value, and $\kappa=\sigma_1(\frac{\Theta^{*\top} \Theta^*}{M})/\nu=\sigma_1(\frac{\alpha^{*\top} \alpha^*}{M})/\nu$ be the condition number. We assume that $\nu>0$ and $\kappa\leq O(1)$.
\end{assumption}
\begin{assumption}[Feature design]\label{as:feature}
    Let $\Sigma_*=\mathbb E_g(\phi(s,a_{1})-\phi(s,a_{0}))(\phi(s,a_{1})-\phi(s,a_{0}))^{\top}$ be the covariance matrix, where the randomness is induced by the generation distribution $g$ for the triplet $(s,a_{0},a_{1})$. We assume that $\Sigma_*$ has bounded positive eigenvalues, i.e., $C_{\text{max}}\geq\lambda_1\geq\cdots\geq \lambda_d\geq C_{\text{min}}>0$. 
\end{assumption}

{Assumption~\ref{as:well} is a standard diversity condition on the underlying reward parameters $\{\theta_m^*\}_{m=1}^M$ in shared-representation learning and multi-task learning; see Assumption~2 in \cite{Tripuraneni2020ProvableMO} and Assumption~4.3 in \cite{du2021few}. In our setting, it requires the coefficient vectors \(\{\alpha_m^*\}_{m=1}^M\) to collectively span the \(r\)-dimensional shared subspace and to do so in a well-conditioned manner. This does not require every pair of parties to have substantially different preferences. Rather, it requires sufficient variation across the collection of parties to identify all directions of the shared representation. Assumption~\ref{as:feature} is a well-conditioned design condition on the feature-difference covariance matrix $\Sigma_*$. Since the BTL model depends on the reward parameters through the feature differences $\phi(s,a_1)-\phi(s,a_0)$, this assumption ensures that these feature differences contain enough information to distinguish different reward parameters. Our results can also be modified to accommodate the case where $\Sigma_*$ is not invertible by replacing it with $\Sigma_*+\lambda I$ for any regularization parameter $\lambda\in\mathbb R^+$; see Remark~3.4 of \cite{zhu2023principled}.}

Before moving on to sub-optimality, we first quantify the estimation error of the jointly learned party-specific reward parameters.
\begin{theorem}\label{th:metatheta}
    Suppose Assumptions \ref{as:well},\ref{as:feature} hold. If the number of samples satisfies $n\gg d+\log(M/\delta)$ for $\delta\in (0,1)$, then with probability at least $1-\delta$, for any $m\in [M]$,
    \begin{equation*}
        \|\hat{\theta}_m-\theta_m^*\|_{\Sigma_m}
        {\lesssim \sqrt{\frac{1}{n}\big({r^2}+\dfrac{dr^2\log{d}}{M}+\log(M/\delta)\big)}}.
    \end{equation*}
\end{theorem}
\textit{Proof Sketch.} We provide a proof outline here, and defer the details to Appendix \ref{pf:meta}. First, we establish an overall bound on $\sum_{m=1}^M\|X_m\hat{\theta}_m-X_m\theta_m^*\|^2$ by leveraging the strong convexity of $l_{\mathcal{D}}(U,\alpha)$, based on second-order Taylor expansion and a careful use of concentration inequalities. 
Next, we utilize Assumption~\ref{as:well} to derive a bound on $\|\hat{\theta}_m-\theta_m^*\|_{\Sigma_m}$ in terms of $U$ and $\alpha_m$ using the Davis-Kahan $\sin \theta$ theorem and Bernstein-type inequality. \qed


{Theorem \ref{th:metatheta} implies that with probability at least $1-\delta$, $\theta_m^*\in\mathbf{\Theta}_B(\hat{\theta}_m)$ for $\forall m\in [M]$. Then we specify $\Gamma(\delta,n,M,d,r)=K\sqrt{\frac{1}{n}\big({r^2}+\dfrac{dr^2\log{d}}{M}+\log(M/\delta)\big)}$ in Algorithm \ref{alg1} with $K$ a constant, thus completing the pessimistic component. It is a generalization of the upper bound of Lemma 3.1 in \cite{zhu2023principled}, which improves the previous bound of $O(\frac{d}{n})$ to $O(\frac{dr^2}{Mn})$ up to logarithmic factors when $r^2\ll \min\{M,d\}$. We will see that the advancement also improves subsequent sub-optimality bounds. 
}


Theorem~\ref{th:metatheta} controls reward-estimation error in the covariance norm induced by the offline comparisons. To derive a policy-suboptimality guarantee, we need to convert this parameter estimation error into an error in the expected reward of the optimal policy. We capture this relationship through the following concentratability coefficient: 
\begin{equation}\label{eq:concen-nash}
    C^*=\max_m\|\Sigma_m^{-1/2}\mathbb E_{s\sim\rho} \phi(s,\pi^*(s))\|_2.
\end{equation}
We have the following sub-optimality guarantee for the pessimistic policy. 
\begin{theorem}\label{th:pes+}
    Under the same condition as Theorem \ref{th:metatheta}, with probability at least $1-\delta$,
    \begin{equation*}
        \texttt{SubOpt}(\hat{\pi})
        {\lesssim  C^*\sqrt{\frac{1}{n}\big({r^2}+\dfrac{dr^2\log{d}}{M}+\log(M/\delta)\big)}}.
    \end{equation*}
\end{theorem}
\textit{Proof Sketch.} We provide a proof outline here, and defer the details to Appendix \ref{pf:pes+}. The sub-optimality can be decomposed as, $J(\pi^*)-J(\hat{\pi})=(J(\pi^*)-\hat{J}(\pi^*))+(\hat{J}(\pi^*)-\hat{J}(\hat{\pi}))+(\hat{J}(\hat{\pi})-J(\hat{\pi}))$ and what we truly need to focus on is the first term $J(\pi^*)-\hat{J}(\pi^*)$ since the latter two terms are smaller than zero with high probability. 
We then address the geometric form of the Nash welfare function by decomposing it into the aggregation of individual errors. \qed

As claimed before, the sample complexity reduction described in Theorem \ref{th:metatheta} extends to Theorem \ref{th:pes+}, highlighting the advantage of low rank shared-representation learning. In our multi-party setting, sub-optimality depends on a concentratability coefficient $C^*$ across the aggregation of $M$ parties, setting it apart from single-party RLHF. Notably, $C^*$ is assumed to be bounded \citep{zhan2024provable,zhu2023principled}, which provides a coverage guarantee for the feature vectors $\phi(s,\pi^*(s))$ from the dataset $\mathcal{D}$, thus yielding an accurate estimator.

To complement the upper bound, we proceed to investigate the lower bound in the following. We show that, for the problems with bounded concentratability coefficients $C^*$ as defined in \eqref{eq:concen-nash}, we obtain the following result, whose proof is deferred to Appendix \ref{pf:nash-lb}.

\begin{theorem}[Lower Bound]\label{th:nash-lb}
    Consider the family of instances $\text{CB}(M,\mathcal{C})=\{\rho,\{s_m^i,a_{m,1}^i,a_{m,0}^i\}_{i\in[n],m\in[M]},\\\Theta^*=U^*\alpha^*|C^*\leq\mathcal{C}\}$, where $C^*$ is defined in \eqref{eq:concen-nash}. For any bandit instance $\mathcal{Q}(M)\in \text{CB}(M,\mathcal{C})$, let $\texttt{SubOpt}_{\mathcal{Q}(M)}$ be the sub-optimality under instance $\mathcal{Q}(M)$. Suppose $r>6,n\gtrsim  r\mathcal{C}^2, \mathcal{C}\geq 2$, then there exists a feature mapping $\phi$ such that the following lower bound holds.
    \begin{equation*}
        \inf_{\hat{\pi}}\sup_{\mathcal{Q}(M)\in \text{CB}(M,\mathcal{C})} \texttt{SubOpt}_{\mathcal{Q}(M)}(\hat{\pi})\gtrsim\mathcal{C}\sqrt{\dfrac{r}{n}}.
    \end{equation*}
\end{theorem}

The lower bound, together with the previous upper bound, highlights the complexity challenges in multi-party RLHF. In this setting, we encounter separate samples for each party and a potentially larger concentratability coefficient $C^*$ defined in \eqref{eq:concen-nash}, in contrast to the concentratability coefficient presented in \cite{zhu2023principled}. Consequently, our shared-representation learning technique plays an important role in achieving an improved sub-optimality bound.
Obtaining tight dependence of our sub-optimality bound on feature dimensions is challenging. The $\sqrt r$ gap between Theorem \ref{th:pes+} and Theorem \ref{th:nash-lb} relates to an open problem in classical estimations, see a more detailed discussion on page 8 in \cite{Tripuraneni2020ProvableMO}. 

\subsection{Efficiency and Welfare Guarantees}\label{sec:social}
Beyond the optimality result, in this section, we demonstrate the efficiency guarantees of our method, as our proposed approach addresses social choice considerations. Section~\ref{sec:swfcomp} establishes that the deterministic policy \(\widehat\pi\) achieves Nash welfare close to that of the optimal policy \(\pi^*\). Because the Nash welfare objective is strictly increasing in every positive party value, an exact Nash-welfare maximizer is Pareto efficient. This observation motivates us to investigate the concept of $\tau$-\textit{approximate Pareto efficiency}. We now examine the approximate efficiency guarantee inherited by the learned policy when its Nash-welfare suboptimality is small. 
\begin{definition}[$\tau$-approximate Pareto Efficiency]
    A deterministic policy $\pi:\mathcal{S}\to \mathcal{A}$ is said to be $\tau$-approximate Pareto efficient, if there does not exist another deterministic policy $\pi'$ such that for all party $m \in [M]$, $\mathbb{E}_{s \sim \rho} R_{\theta_m^*}(s, \pi'(s)) \geq \mathbb{E}_{s \sim \rho} R_{ \theta_m^*}(s, \pi(s))$,
    and for at least one party $m \in [M]$,
    $\mathbb{E}_{s \sim \rho} R_{ \theta_m^*}(s, \pi'(s)) \geq (1 + \tau) \mathbb{E}_{s \sim \rho} R_{ \theta_m^*}(s, \pi(s))$.
\end{definition}

This notion of efficiency ensures an agreement and balance among multiple parties, contributing to overall societal wellness by optimizing the collective outcomes. It's an adaptation of the definitions by \cite{aumann2010pareto,barman2018finding} to RLHF settings. While previous work considered an efficiency notion that allows a $\tau$ improvement for all parties, we admit a $\tau$ improvement for only one party when maintaining others unchanged. 
We now show that the Nash-welfare suboptimality bound from Theorem~\ref{th:pes+} implies approximate Pareto efficiency of the learned deterministic policy.
\begin{theorem}\label{th:pareto}
    Under the same condition as Theorem \ref{th:metatheta}, with probability at least $1-\delta$, $\hat{\pi}$ is $\tau$-approximate Pareto efficient, where 
    $\tau
    {\lesssim  C^*M\sqrt{\frac{1}{n}\big({r^2}+\dfrac{dr^2\log{d}}{M}+\log(M/\delta)\big)}}$.
\end{theorem}
This result confirms that we cannot improve one party significantly without making another party worse, thus leading to an approximately efficient policy.
Besides, the pessimistic policy $\hat{\pi}$ follows the approximate Pigou-Dalton principle, which emphasizes the transfer of welfare from better-off individuals to weaker ones, ensuring a more equitable distribution of welfare outcomes \citep{moulin2004fair}. The definition proof of the efficiency and welfare guarantees are given in Appendix \ref{pf:pareto}

\subsection{Comparison with Various Social Welfare Functions}\label{ap:moreswf}

{The preceding subsections focus on Nash welfare as the primary aggregation rule, since it balances overall welfare with protection against severe imbalance across parties. As outlined in Section \ref{sec:preswf}, we provide a brief introduction to other social welfare functions and compare their results. Once the confidence sets for party-specific rewards have been constructed, we can replace the Nash aggregation in Line~3 of Algorithm~\ref{alg1} with another social welfare objective and optimize the corresponding pessimistic empirical value. To illustrate this flexibility, we consider two canonical alternatives introduced in Section~\ref{sec:preliminaries}: the Utilitarian objective, which optimizes average welfare, and the Egalitarian objective, which optimizes the welfare of the worst-off party. As shown below, both variants inherit the same statistical rate, so the choice of welfare function mainly determines the social criterion being optimized rather than the reward-learning difficulty.}

\paragraph*{Utilitarian ($\alpha=1$)} {Under the Utilitarian paradigm, the goal is to maximize the arithmetic mean of the expected returns across all parties.} Similarly, the (pessimistic) Utilitarian welfare function is defined as follows,
\begin{align*}
    &J({\pi}):=\dfrac{1}{M} \sum_{m=1}^M \mathbb E_{s\sim\rho} R_{m,\theta_m^*}(s,{\pi}(s)),\\
    &\hat{J}(\pi)=\min_{\Theta\in\mathbf{\Theta}_B(\hat{\Theta})}\dfrac{1}{M} \sum_{m=1}^M\mathbb E_{s\sim\rho}  R_{m,\theta_m}(s,{\pi}(s)).
\end{align*}

\paragraph*{Egalitarian ($\alpha\rightarrow -\infty$)} 
The Egalitarian approach prioritizes the worst-off party by maximizing the minimum expected return among all parties, denoted as $\min_m \mathbb E_{s\sim\rho}  R_{m,\theta_m}(s,{\pi}(s))$. Similarly, the (pessimistic) Egalitarian welfare function is defined as follows,
\begin{align*}
    &J({\pi}):=\min_m \mathbb E_{s\sim\rho}R_{m,\theta_m^*}(s,{\pi}(s)),\\
    &\hat{J}(\pi)=\min_{\Theta\in\mathbf{\Theta}_B(\hat{\Theta})} \min_m \mathbb E_{s\sim\rho}R_{m,\theta_m}(s,{\pi}(s)).
\end{align*}

For both cases, define $\hat{\pi}=\arg\max_{\pi}\hat{J}(\pi)$, $\pi^*=\arg\max_{\pi}J(\pi)$. 
We have the following sub-optimality bound.
\begin{theorem}\label{th:leximin}
    Under the same condition as Theorem \ref{th:metatheta}, with probability at least $1-\delta$,
    $$\texttt{SubOpt}(\hat{\pi})
    {\lesssim  C^*\sqrt{\frac{1}{n}\big({r^2}+\dfrac{dr^2\log{d}}{M}+\log(M/\delta)\big)}}.$$
    Here $C^*=\max_m\|(\Sigma_m^{-1/2}\mathbb E_{s\sim\rho}\phi(s,\pi^*(s)))\|_2$. 
\end{theorem}
Theorem \ref{th:leximin} implies that $\hat{\pi}$ is approximately optimal for the average/worst-off party, thus characterizing a relatively fair outcome. Each welfare function measures different aspects of social welfare, making them suitable for different practical settings.

\paragraph{Generalization to MDP}
As claimed earlier, the contextual bandit (CB) can be viewed as a special case of a Markov decision process (MDP) with finite horizons and transitions over states. For multi-party RLHF under MDP settings, preferences are based on the whole trajectories rather than single actions \citep{zhan2024provable}. The generalized algorithm for MDPs is similar to Algorithm \ref{alg1}, and sub-optimality upper bound is also guaranteed. However, in this context, we adopt a trajectory-based preference model and optimize a trajectory-wise value function. The key difference between MDPs and CBs is the inclusion of a trajectory-based occupancy measure $d^{\pi}(s)=\sum_{h=1}^H\mathbb P_h(s_h=s|\pi)$, which reduces to $\rho$ for CBs. To proceed with the result, we adopt the observation from \citep{zhu2023principled} that: 
\begin{equation*}
    \mathbb E_{s\sim\rho}[V_m^\pi(s)]=\mathbb E_{s\sim d^\pi} R_{m,\theta_m^*}(s,{\pi}(s)),
\end{equation*}
where $V_m^\pi(s)=\mathbb E[\sum\limits_{h=1}^H R_m(s_h,a_h)|s_1=s,\pi]$, and this expectation $\mathbb E$ is taken over the trajectory generated according to the transition kernel. The details are provided in Appendix \ref{sec:mdp}.

\section{Extension to General Pairwise-Preference Model}\label{sec:ext}

The reward-representable regime in Section~\ref{sec:rwd} represents each party's preferences through a scalar reward function, which is natural for RLHF but also imposes structural restrictions on the preference relation. In particular, scalar reward models are most suitable when preferences can be summarized by an underlying ranking of actions or trajectories. Human preferences, however, may exhibit more complex patterns, including intransitivities observed in psychological experiments~\citep{tversky1969intransitivity}. To capture such cases, we now extend our framework to a more general regime, where preferences are represented directly through pairwise comparison preference probabilities rather than through latent reward functions. We retain the central structure of the preceding sections, where we learn a single collective policy from offline dataset collected from all parties. Because scalar expected preference value for each party is no longer available for social-welfare aggregation, we instead average the party-specific pairwise preference margins for each state-action pair tuple $(s,a_0,a_1)$ and select a pessimistic von Neumann winner, which maximizes the worst-case aggregate preference margin against competing strategies.
Throughout this section, we focus on the tabular setting with finite state and action spaces, where $|\mathcal{S}|=S$ and $|\mathcal{A}|=A$. This allows us to avoid additional function-approximation assumptions and to work directly with state-wise preference matrices. It also leads naturally to finite zero-sum matrix games, which provide the algorithmic basis for computing von Neumann winners.

\paragraph*{General Pairwise-Preference Model} 
{In the general pairwise-preference setting, we no longer assume that pairwise preferences are generated by reward differences. Instead, for each party $m$ and state $s$, we represent preferences directly through the skew-symmetric matrix. Specifically, } To ensure symmetry, we define the following preference matrix 
\citep{dudik2015contextual}:
\begin{equation*}
    N_{s,m}^*(a_1,a_0)=2\mathbb P_m(y=1|a_1,a_0,s)-1, \quad \forall m\in [M], s\in \mathcal{S}.
\end{equation*}
Thus, $N_{s,m}^*$ is a skew-symmetric matrix.
In this general model, we refrain from directly specifying reward functions to model human preferences. Consequently, it's unfeasible to obtain an optimal policy based on an aggregated reward function. We consider an alternative solution concept, the \textit{von Neumann winner} \citep{dudik2015contextual}, which corresponds to a probabilistic strategy that outperforms every other strategy with at least a 50\% probability of being preferred. 
Formally, a von Neumann winner of an $A\times A$ skew-symmetric preference matrix $N$ is the max-min probabilistic strategy $p^*$, defined as follows, with matrix $N$ describing a matrix game: 
$$p^*,q^*=\arg\max_p\min_q p^{\top}Nq=\arg\max_p\min_q\sum_{i,j} p_iq_jN_{ij}.$$ 
The value of the matrix game is defined as $p^{*\top}Nq^*$, which equals $0$ when $N$ is skew-symmetric \citep{owen2013game}. 
To connect with the previous model, if \eqref{eq:rwdpro} holds, the von Neumann winner described in the following section corresponds to the Utilitarian welfare solution when $M=1$ or $M=2$. See Appendix \ref{ap:connect} for details.

\subsection{Pessimistic Von Neumann Winner Algorithm}\label{sec:pes-von}

We now construct a pessimistic von Neumann winner from the offline party-indexed comparisons dataset. The procedure has two stages. First, we estimate the party-specific preference matrices, aggregate them using equal party weights, and construct entrywise confidence bounds. Second, for each state, we solve a zero-sum matrix game using the pessimistic payoff matrix and use its max-min strategy to define a randomized collective policy. Algorithm 2 summarizes the procedure.
\begin{algorithm}[ht]
    \caption{Pessimistic Von Neumann Winner}\label{alg2}
	\textbf{Input:} The datasets $\mathcal{D}=\bigcup_{m=1}^M\mathcal{D}_m$, a failure probability $\delta$, and the initial distribution $\rho$.

    \textbf{Output:} The pessimistic policy $\hat{\pi}$.
	
	\begin{algorithmic}[1]
        \STATE Calculate the preference matrix $N_s$ in \eqref{eq:matrix} for $\forall s\in\mathcal{S}$.
		\STATE Construct the pessimistic lower confidence bound $N_s-B_s$ through \eqref{eq:matrix} for $\forall s\in\mathcal{S}$.
        \STATE Calculate the state-wise strategy $\hat{p}_s$ by solving the matrix game of $N_s-B_s$ for $\forall s\in\mathcal{S}$.
        \STATE Obtain the final randomized policy $\hat{\pi}$ (induced by $\hat{p}_s$) through \eqref{eq:pospace}.
	\end{algorithmic}  
\end{algorithm}

We construct a pessimistic von Neumann winner in a Utilitarian manner to learn a randomized policy. The algorithm is described in Algorithm \ref{alg2}. It mainly consists of two steps: Step 1 is to calculate the aggregated preference matrices and construct the confidence bounds; Step 2 is to solve a zero-sum matrix game of each state and finally construct a corresponding randomized policy in the policy space.


\textbf{Step 1: Preference matrix with confidence bonus.} We construct preference matrices by aggregating each party's preferences in a Utilitarian manner. Given $\mathcal{D}_m=\{(s_{m}^i, a_{m,1}^i, a_{m,0}^i,y_{m}^i)\}_{i=1}^{n}$, we define the \textit{visiting number} of an action pair at one state as $n_0(m,a,a',s):=\#_m(a\succ a';s)+\#_m(a'\succ a;s)$, where $\#_m(a\succ a';s)$ denotes the number of observations in $\mathcal{D}_m$ of preferring $a$ over $a'$ at state $s$. 
We then Define the preference matrix and its confidence bonus as follows:
\begin{equation}\label{eq:matrix}
    N_s(a,a')=\dfrac{1}{M}\sum_{m=1}^MN_{s,m}(a,a'), \quad B_s(a,a')=\sqrt{\frac{2\log (4MSA^2/\delta) }{\min_m n_0(m,a,a',s)\lor 1}}, \quad \forall a\neq a'\in \mathcal{A},
\end{equation}
where $\delta$ is a given failure probability, and $N_{s,m}(a,a')=\frac{\#_m(a\succ a';s)-\#_m(a'\succ a;s)}{\#_m(a\succ a';s)+\#_m(a'\succ a;s)}$ if the denominator is non-zero; otherwise, it is defined as $0$. The diagonal entries of these matrices are set to $0$. Here, the matrix $N_s$ is an aggregated preference matrix, while $B_s$ measures the uncertainty of $N_s$.

\textbf{Step 2: Max-min optimization.} After constructing the confidence bounds, we seek a max-min probabilistic strategy under a pessimistic model. This is achieved by solving the following zero-sum matrix game using the lower confidence bound $N_s-B_s$ of the preference matrix, and then obtaining the final probabilistic strategy:
\begin{equation}\label{eq:pospace}
    \hat{p}_s,\hat{q}_s=\arg\max_{p}\min_{q}p^{\top}(N_s-B_s)q, \forall s \in \mathcal{S}, \quad \hat{p}(\pi)=\prod_{s\in \mathcal{S}}\hat{p}_s(\pi(s)).
\end{equation}
To be clear, $\hat{p}_s$ is a probabilistic strategy for each state $s\in \mathcal{S}$, and $\hat{p}(\pi)$ is a probabilistic strategy over the policy space $\Pi$, where $\Pi$ encompasses all deterministic policy permutations (any $\pi: \mathcal{S}\rightarrow \mathcal{A}$). Thus, $\hat{p}$ induces a randomized policy $\hat{\pi}:\mathcal{S}\rightarrow \Delta(\mathcal{A})$. 



\subsection{Efficiency Guarantees}
In this section, we demonstrate that we obtain an approximate von Neumann winner of the ground truth preference matrix.
To see this, we define the population-wise preference matrices at state $s$ as $N_s^*(a,a')=\dfrac{1}{M}\sum_{m=1}^MN_{s,m}^*(a,a')$, and the one in the policy space $\Pi$ as $T^*(\pi,\pi')=\mathbb E_{s\sim\rho}N_s^*(\pi(s),\pi'(s))$, respectively. Here $N_s^*(a,a')$ denotes the aggregated preference between action pairs at state $s$, and $T^*(\pi,\pi')$ denotes the aggregated preference between deterministic policy pairs.

Recall that $\mathcal{D}_m$ is generated by distribution $g$ as described in Section \ref{sec:predata}. We define the occupancy measure $d_s^g(a,a')$ as the probability of pair $(a,a')$ appearing in the data generated by $g$ for a given state $s$: $d_s^g(a,a')=g_a(a,a'|s)+g_a(a',a|s)$. Similarly, define $d_s^{\mu,\nu}(a,a')$ as the probability of $(a,a')$ appearing in the data generated by policies $(\mu,\nu)$ respectively for a given state $s$. These definitions are adopted from those in \citep{cui2022offline}. 
Define $p_s^*,q_s^*=\arg\max_{p}\min_{q}p^{\top}N_s^*q$ as the ground truth max-min and min-max probabilistic strategy at state $s$. 

We further assume that $\rho(s)>0$ for each $s\in \mathcal{S}$. Since $|\mathcal{S}|<\infty$, this implies a uniform lower bound of $\rho(s)$ for $s\in \mathcal{S}$. Now we define the concentratability coefficient  as follows 
$$C^*=\max_{\nu,s,a\neq a'}\dfrac{d_s^{p_s^*,\nu}(a,a')}{\rho(s)d_s^g(a,a')},$$ 
where $C^*$ is assumed to be bounded \citep{cui2022offline}. This definition is similar to \eqref{eq:concen-nash}. It provides a coverage guarantee of the dataset: the ratio of the state-action occupancy induced by the optimal policies, to the data distribution, is bounded.

Till now, we can establish our main result, whose proof is deferred to Appendix \ref{pf:von-pes}.
\begin{theorem}\label{th:von-pes}
    With probability at least $1-\delta$, $\hat{p}_s$ and $\hat{p}$ are $\epsilon$-approximate von Neumann winner, i.e.,
    \begin{equation*}
        \min_q\hat{p}_s^{\top}N_s^*q\geq -\epsilon,  \forall s\in \mathcal{S} \quad \text{and}
        \quad \min_q\hat{p}^{\top} T^*q\geq -\epsilon,
    \end{equation*}
    where $\epsilon=8A\log (4MSA^2/\delta)\sqrt{\frac{C^*}{n}}$.
\end{theorem}
Since the value of the skew-symmetric matrix game is zero \citep{owen2013game}, Theorem \ref{th:von-pes} demonstrates  that the learned strategy achieves a worst-case population preference margin of at least \(-\epsilon\), both statewise and over the induced policy space. Thus, no competing strategy can defeat the learned strategy by more than \(\epsilon\) in aggregate preference margin. The error boound $\varepsilon$ decreases at rate \(n^{-1/2}\), grows with the action-space size, and depends only logarithmically on the numbers of parties and states apart from their influence through the concentratability coefficient $C^*$. 

As we focus on social choice aspects, we again present a (weak) Pareto efficiency result to ensure consensus among multiple individuals, adapted from the concept of \textit{ex-post efficiency} \citep{fishburn1984probabilistic}. 

\begin{definition}[$\tau$-Approximate Ex-post Efficiency]
    A probabilistic strategy $p_s$ is $\tau$-approximate ex-post efficient at state $s$ if any Pareto-dominated alternative $b$ receives probability $p_s(b)\leq \tau$. Here an alternative $b$ is Pareto-dominated if there exists an alternative $a$ s.t. $N_{s,m}^*(a,b)\geq 0$ for $\forall m \in [M]$. 
\end{definition}
Our definition differs from \cite{zeng2020fairness}, where they considered a strategy that always outputs a $\tau$-approximate Pareto efficient alternative. We consider an ignorable probability ($\leq \tau$) of outputting a Pareto-dominated (universally disliked) alternative instead. Thus, a smaller \(\tau\) means that the learned strategy is less likely to select an action that admits a common replacement weakly preferred by every party. We now present the following efficiency result. 

\begin{theorem}\label{th:expost}
    If the preference of each party is transitive, i.e., $N_{s,m}^*(x,y)\geq 0$ implies $N_{s,m}^*(x,z)\geq N_{s,m}^*(y,z)$ for $\forall z\in \mathcal{A}$, then with probability at least $1-\delta$, $\hat{p}_s$ is $\tau$-approximate ex-post efficient at state $s$, where $\tau\lesssim\left(A\log(2MSA^2/\delta)\sqrt{\frac{C^*}{n}}\right)^{1/2}$.
\end{theorem}
Theorem \ref{th:expost} guarantees that Pareto-dominated alternatives are unlikely to be selected.  Here the transitivity is necessary, implying the challenge of the general model. The proof and further details are provided in Appendix \ref{pf:expost}.
This result, in line with Theorem \ref{th:pareto}, underscores the efficiency and welfare guarantees of multi-party RLHF and offers a solid foundation from a social choice perspective.

\section{Numerical Illustrations}\label{sec:exp}
We provide two stylized contextual-bandit simulations to illustrate the behavior of the proposed methods under heterogeneous party preferences and unequal party sample sizes. The first considers the reward-representable model and compares pessimistic Nash-welfare optimization with a pooled reward-model baseline. The second applies the general pairwise-preference method without using the underlying reward representation. These simulations isolate the effect of retaining party identities and applying an explicit equal-weight aggregation rule.

\subsection{Reward-Representable Model}\label{sec:exp1}

We consider a reward-representable instance with an unknown low-rank representation, where $r\ll d$. Specifically, we construct $|\mathcal{S}|=3$ states, $|\mathcal{A}|=9$ actions, and $M=24$ parties, and set $d=24$ and $r=3$. The state distribution is uniform over $\mathcal{S}$.  We defer the complete feature construction, data-generating process, and assumption verification to the Appendix~\ref{app:exp1}. For every party, the offline dataset compares each of the eight non-reference actions $a_1,\ldots, a_8$ with the reference action $a_9$ in every state. The minimum per-party sample size ranges over \(n=150,300,\ldots,3000,\) and every sample size is repeated \(100\) times. In the experiment, the parties are divided into three equally sized clusters of eight; four parties in each cluster provide \(n\) comparisons and the other four provide \(2n\) comparisons (Note that our results remain valid for varying sample sizes $n$ across different parties by replacing $n$ with $\min_m n_m$). 

We compare our joint low-rank Pessimistic Nash Welfare algorithm with three plug-in baselines. 
1) The independent baseline estimates a separate unrestricted \(d\)-dimensional reward parameter for every party and therefore does not exploit the shared representation. 
2) The pooling-obs baseline discards party identities and fits one reward model by weighting every comparison equally, so parties with \(2n\) observations contribute twice as much as parties with \(n\) observations. 
3) The pooling-party baseline also fits a single reward model but normalizes each party’s contribution, giving every party equal total weight. 
The two pooling baselines therefore distinguish the effect of unequal observation counts from the more fundamental loss of party-specific preference information. All methods select from the complete class of \(9^3=729\) deterministic policies, and their performance is evaluated using the true party-specific rewards.


Figure~\ref{fig:nash_simulation} shows that our method achieves lower Nash welfare suboptimality than the independent and pooling baselines throughout the reported sample-size range. The independent estimator improves as the number of comparisons increases but remains less accurate than the joint method. Together with the estimation-error results in the appendix, this pattern supports the benefit of learning the shared low-dimensional reward representation. By contrast, both pooling baselines maintain a substantial suboptimality gap because a single pooled reward model cannot represent the three clusters’ heterogeneous preferences in this construction.

We further evaluate the same Nash-targeted policies under the Utilitarian and Egalitarian objectives and report their suboptimality relative to pooling-obs in the right panel. We also report a party-level dispersion measure that examines the difference between parties’ average regrets (see Appendix~\ref{app:exp}). All ratios remain below one, indicating that our Nash-targeted policy also has lower Utilitarian and Egalitarian suboptimality, and a more balanced party-level regret profile than the pooled policy. These results indicate that the empirical advantage of our method persists when the selected policy is evaluated under welfare criteria other than the one it was optimized for. This broader improvement across welfare criteria is consistent with the empirical findings of \citep{bakker2022fine}.
Additional details on evaluation metrics and results on confidence-set coverage and estimation accuracy are reported in Appendix~\ref{app:exp}.

\subsection{General Pairwise Preference model}\label{sec:exp2}

\begin{figure}[H]
    \centering
    \begin{minipage}[t]{0.6\textwidth}
		\centering
		\includegraphics[width=0.99\textwidth]{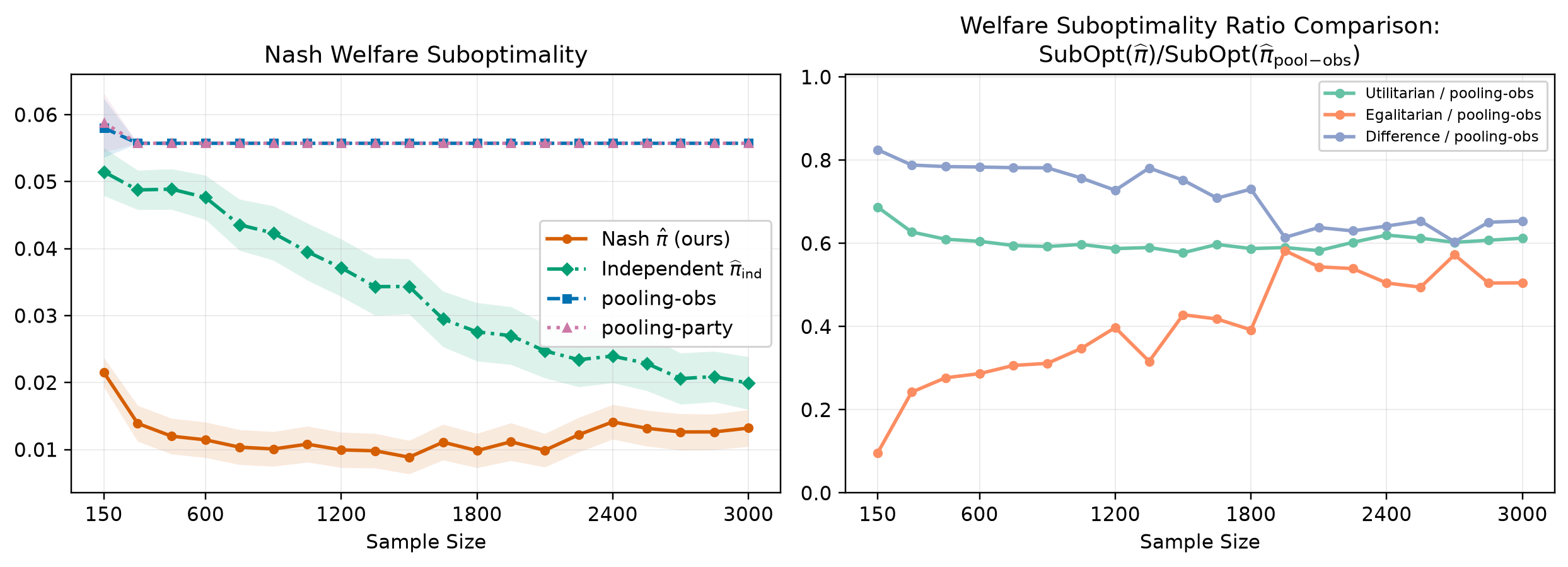}
        \subcaption{Convergence of Nash welfare suboptimality and robustness under alternative suboptimality measures defined in Section \ref{sec:rwd}.}
        \label{fig:nash_simulation}
	\end{minipage}
    \begin{minipage}[t]{0.35\textwidth}
		\centering
		\includegraphics[width=0.9\textwidth]{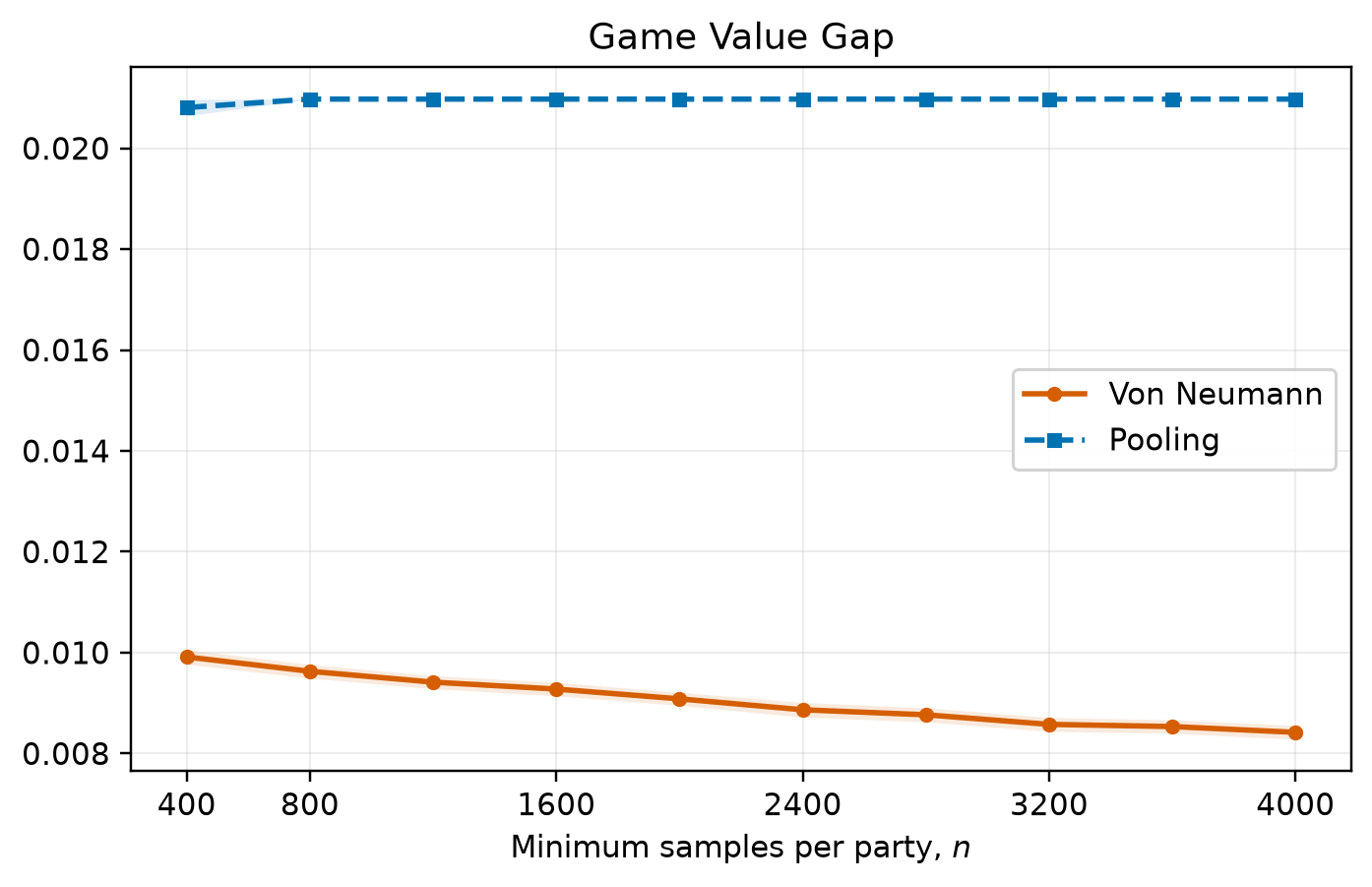}
        \subcaption{Convergence of the game value.}
        \label{fig:von_simulation}
	\end{minipage}
    \caption{Simulation results.}
\end{figure}

Additionally, we present the results for the general pairwise-preference model. We consider a separate one-state instance with three actions and two parties. Although we use a BTL model to generate a controlled population preference matrix, the proposed method and the pooling baseline observe only pairwise comparison outcomes; neither method receives the underlying rewards, features, or reward parameters. We query action pairs $(a_1,a_2)$; $(a_1,a_3)$ and $(a_2,a_3)$ an equal number of times within each party, while the sample sizes of Party 2 is twice that of Party 1. Let \(n\) denote the minimum per-party sample size. Party 1 provides \(n\) comparisons, whereas Party 2 provides \(2n\) comparisons, we range the the $n$ from $400$ to $4000$.  Each sample size is repeated $500$ times. The results are shown in Figure \ref{fig:von_simulation}. Again, as the sample size increases, the game-value suboptimality of the strategy learned by our method decreases toward zero as the sample size increases, whereas the observation-pooling baseline retains a non-vanishing suboptimality gap. This demonstrates the robustness of our approach in both structured and unstructured preference settings. We defer further experimental setup and analysis details in Appendix~\ref{app:general-preference-experiment}.

\section{Conclusion}
Pluralistic alignment is not only a problem of learning from more people; it is also a problem of preserving meaningful disagreement and deciding explicitly how that disagreement should shape a single collective policy. This paper develops an offline finite-sample framework for doing so from party-indexed human comparisons. Rather than pooling heterogeneous feedback into one implicit reward model, our approach retains party-specific preference information throughout learning and separates the statistical task of estimating preferences from the normative choice of how those preferences should be aggregated.
In the reward-representable regime, we exploit a shared low-rank structure to jointly estimate party-specific reward functions while preserving their differences. We then propagate the resulting statistical uncertainty through pessimistic Nash, Utilitarian, and Egalitarian welfare optimization. Under the stated realizability, diversity, normalization, and offline-coverage conditions, our guarantees characterize how shared structure, per-party estimation error, and data coverage jointly determine the quality and approximate efficiency of the learned collective policy.
We further move beyond scalar reward representations to settings in which preferences may be stochastic, nontransitive, or cyclic. By estimating party-specific pairwise preferences and constructing a pessimistic von Neumann winner, we obtain a randomized collective policy with an approximate max-min preference guarantee. Under additional structural conditions, the policy also assigns limited probability to Pareto-dominated alternatives.
Together, these results make preference aggregation an explicit and statistically analyzable component of RLHF, rather than an unintended consequence of data pooling. More broadly, they provide a principled foundation for collective learning systems in which both the statistical assumptions used to learn from heterogeneous feedback and the normative choices used to resolve disagreement are made transparent.

\section{Discussion}\label{sec:discuss}
\paragraph{KL penalty.} In practice, people maximize a KL-regularized value function, which includes an additional term $\gamma D_{\text{KL}}(\hat{\pi}(\cdot|s)\|\pi_{\text{ref}}(\cdot|s))$, where $\gamma$ is a penalty parameter and $\pi_{\text{ref}}$ is a reference policy. While our primary focus is the basic version, our results can be extended to the KL version, where the sub-optimality gap becomes accompanied by an additional $\gamma \mathbb E_{s\sim\rho}D_{\text{KL}}(\hat{\pi}(\cdot|s)\|\pi^*(\cdot|s))$ term \citep{xiong2023iterative}.

\paragraph{Computation.} Algorithm \ref{alg1} is an extension of algorithms in \cite{zhan2024provable,zhu2023principled},
all of which are typically computationally heavy. Here, we illustrate how to practically implement the algorithm.
Step 2 of Algorithm \ref{alg1} involves solving a constrained max-min optimization problem. By introducing the Lagrangian multiplier, which is akin to the approach employed in Appendix A of \cite{zhan2024provable}, it is feasible to apply gradient ascent-descent using policy gradient algorithms. 
In the general pairwise-preference model, finding a von Neumann winner, i.e., Step 2 of Algorithm \ref{alg2}, involves solving the matrix game. This can be achieved by linear programming \citep{barron2013game}, making it computationally efficient. Nevertheless, this efficiency for the general pairwise-preference model is achieved in tabular cases.

\paragraph{Future work.} Several directions remain open. First, it would be useful to extend shared multi-party reward learning beyond linear reward models to general function classes while retaining finite-sample confidence guarantees and propagating the resulting uncertainty through social-welfare optimization. Second, an important open problem is to develop finite-sample methods that jointly quantify uncertainty in latent or overlapping party membership and in potentially evolving party-specific preferences, and propagate these sources of uncertainty through social-welfare optimization of a single collective policy. Third, bridging the gap between the present stylized experiments and practical LLM alignment requires scalable methods for party-specific reward estimation, uncertainty-aware welfare optimization, and collective policy fine-tuning over large response and policy spaces.

\bibliographystyle{plainnat}
\bibliography{ref}

@inproceedings{zhu2023principled,
  title={Principled reinforcement learning with human feedback from pairwise or k-wise comparisons},
  author={Zhu, Banghua and Jordan, Michael and Jiao, Jiantao},
  booktitle={International Conference on Machine Learning},
  pages={43037--43067},
  year={2023},
  organization={PMLR}
}

@article{li2023multi,
  title={Multi-dimensional domain generalization with low-rank structures},
  author={Li, Sai and Zhang, Linjun},
  journal={arXiv preprint arXiv:2309.09555},
  year={2023}
}

@inproceedings{
zhan2024provable,
title={Provable Offline Preference-Based Reinforcement Learning},
author={Wenhao Zhan and Masatoshi Uehara and Nathan Kallus and Jason D. Lee and Wen Sun},
booktitle={The Twelfth International Conference on Learning Representations},
year={2024},
}

@inproceedings{dudik2015contextual,
  title={Contextual dueling bandits},
  author={Dud{\'\i}k, Miroslav and Hofmann, Katja and Schapire, Robert E and Slivkins, Aleksandrs and Zoghi, Masrour},
  booktitle={Conference on Learning Theory},
  pages={563--587},
  year={2015},
  organization={PMLR}
}

@article{fishburn1977condorcet,
  title={Condorcet social choice functions},
  author={Fishburn, Peter C},
  journal={SIAM Journal on applied Mathematics},
  volume={33},
  number={3},
  pages={469--489},
  year={1977},
  publisher={SIAM}
}

@article{xie2021policy,
  title={Policy finetuning: Bridging sample-efficient offline and online reinforcement learning},
  author={Xie, Tengyang and Jiang, Nan and Wang, Huan and Xiong, Caiming and Bai, Yu},
  journal={Advances in Neural Information Processing Systems},
  volume={34},
  pages={27395--27407},
  year={2021}
}

@article{cui2022offline,
  title={When are offline two-player zero-sum {M}arkov games solvable?},
  author={Cui, Qiwen and Du, Simon S},
  journal={Advances in Neural Information Processing Systems},
  volume={35},
  pages={25779--25791},
  year={2022}
}

@inproceedings{wang2023rlhf,
title={Is {RLHF} More Difficult than Standard {RL}? A Theoretical Perspective},
author={Yuanhao Wang and Qinghua Liu and Chi Jin},
booktitle={Thirty-seventh Conference on Neural Information Processing Systems},
year={2023}
}

@article{li2022pessimism,
  title={Pessimism for Offline Linear Contextual Bandits using $\ell_p$ Confidence Sets},
  author={Li, Gene and Ma, Cong and Srebro, Nati},
  journal={Advances in Neural Information Processing Systems},
  volume={35},
  pages={20974--20987},
  year={2022}
}

@InProceedings{pmlr-v162-chen22ag,
  title = 	 {Human-in-the-loop: Provably Efficient Preference-based Reinforcement Learning with General Function Approximation},
  author =       {Chen, Xiaoyu and Zhong, Han and Yang, Zhuoran and Wang, Zhaoran and Wang, Liwei},
  booktitle = 	 {International Conference on Machine Learning},
  pages = 	 {3773--3793},
  year = 	 {2022},
  volume = 	 {162},
  publisher =    {PMLR},
}

@inproceedings{NIPS2017_d5e2c0ad,
 author = {Christiano, Paul F and Leike, Jan and Brown, Tom and Martic, Miljan and Legg, Shane and Amodei, Dario},
 booktitle = {Advances in Neural Information Processing Systems},
 editor = {I. Guyon and U. Von Luxburg and S. Bengio and H. Wallach and R. Fergus and S. Vishwanathan and R. Garnett},
 pages = {4302–4310},
 location = {Long Beach, California, USA},
 title = {Deep Reinforcement Learning from Human Preferences},
 volume = {30},
 year = {2017}
}

@article{bradley1952rank,
  title={Rank analysis of incomplete block designs: I. The method of paired comparisons},
  author={Bradley, Ralph Allan and Terry, Milton E},
  journal={Biometrika},
  volume={39},
  number={3/4},
  pages={324--345},
  year={1952},
  publisher={JSTOR}
}

@inproceedings{rivest2010optimal,
  title={An optimal single-winner preferential voting system based on game theory},
  author={Rivest, Ronald L and Shen, Emily},
  booktitle={International Workshop on Computational Social Choice},
  pages={399--410},
  year={2010},
  organization={Citeseer}
}

@book{moulin2004fair,
  title={Fair division and collective welfare},
  author={Moulin, Herv{\'e}},
  year={2004},
  publisher={MIT press}
}

@article{fishburn1984probabilistic,
  title={Probabilistic social choice based on simple voting comparisons},
  author={Fishburn, Peter C},
  journal={The Review of Economic Studies},
  volume={51},
  number={4},
  pages={683--692},
  year={1984},
  publisher={Wiley-Blackwell}
}

@article{brandl2016consistent,
  title={Consistent probabilistic social choice},
  author={Brandl, Florian and Brandt, Felix and Seedig, Hans Georg},
  journal={Econometrica},
  volume={84},
  number={5},
  pages={1839--1880},
  year={2016},
  publisher={Wiley Online Library}
}

@article{Kelly1998,
  title={Rate control for communication networks: shadow prices, proportional fairness and stability},
  author={Kelly, Frank P and Maulloo, Aman K and Tan, David Kim Hong},
  journal={Journal of the Operational Research society},
  volume={49},
  pages={237--252},
  year={1998},
  publisher={Springer}
}

@article{zhang2022proportional,
  title={Proportional Fairness in Federated Learning},
  author={Zhang, Guojun and Malekmohammadi, Saber and Chen, Xi and Yu, Yaoliang},
  journal={arXiv preprint arXiv:2202.01666},
  year={2022}
}

@article{tversky1969intransitivity,
  title={Intransitivity of preferences.},
  author={Tversky, Amos},
  journal={Psychological review},
  volume={76},
  number={1},
  pages={31},
  year={1969},
  publisher={American Psychological Association}
}

@book{owen2013game,
  title={Game theory},
  author={Owen, Guillermo},
  year={2013},
  publisher={Emerald Group Publishing}
}

@book{barron2013game,
  title={Game theory: an introduction},
  author={Barron, Emmanual N},
  year={2013},
  publisher={John Wiley \& Sons}
}

@article{bakker2022fine,
  title={Fine-tuning language models to find agreement among humans with diverse preferences},
  author={Bakker, Michiel and Chadwick, Martin and Sheahan, Hannah and Tessler, Michael and Campbell-Gillingham, Lucy and Balaguer, Jan and McAleese, Nat and Glaese, Amelia and Aslanides, John and Botvinick, Matt and others},
  journal={Advances in Neural Information Processing Systems},
  volume={35},
  pages={38176--38189},
  year={2022}
}

@article{ziegler2019fine,
  title={Fine-tuning language models from human preferences},
  author={Ziegler, Daniel M and Stiennon, Nisan and Wu, Jeffrey and Brown, Tom B and Radford, Alec and Amodei, Dario and Christiano, Paul and Irving, Geoffrey},
  journal={arXiv preprint arXiv:1909.08593},
  year={2019}
}

@article{ouyang2022training,
  title={Training language models to follow instructions with human feedback},
  author={Ouyang, Long and Wu, Jeffrey and Jiang, Xu and Almeida, Diogo and Wainwright, Carroll and Mishkin, Pamela and Zhang, Chong and Agarwal, Sandhini and Slama, Katarina and Ray, Alex and others},
  journal={Advances in Neural Information Processing Systems},
  volume={35},
  pages={27730--27744},
  year={2022}
}

@inproceedings{fish2023generative,
author = {Fish, Sara and G\"{o}lz, Paul and Parkes, David C. and Procaccia, Ariel D. and Rusak, Gili and Shapira, Itai and W\"{u}thrich, Manuel},
title = {Generative Social Choice},
year = {2024},
publisher = {Association for Computing Machinery},
address = {New York, NY, USA},
booktitle = {Proceedings of the 25th ACM Conference on Economics and Computation},
pages = {985},
numpages = {1},
location = {New Haven, CT, USA},
series = {EC '24}
}

@inproceedings{santurkar2023whose,
  title={Whose opinions do language models reflect?},
  author={Santurkar, Shibani and Durmus, Esin and Ladhak, Faisal and Lee, Cinoo and Liang, Percy and Hashimoto, Tatsunori},
  booktitle={International Conference on Machine Learning},
  pages={29971--30004},
  year={2023},
  organization={PMLR}
}

@inproceedings{aumann2010pareto,
  title={Pareto efficiency and approximate Pareto efficiency in routing and load balancing games},
  author={Aumann, Yonatan and Dombb, Yair},
  booktitle={International Symposium on Algorithmic Game Theory},
  pages={66--77},
  year={2010},
  organization={Springer}
}

@inproceedings{barman2018finding,
  title={Finding fair and efficient allocations},
  author={Barman, Siddharth and Krishnamurthy, Sanath Kumar and Vaish, Rohit},
  booktitle={Proceedings of the 2018 ACM Conference on Economics and Computation},
  pages={557--574},
  year={2018}
}

@inproceedings{zeng2020fairness,
  title={Fairness-efficiency tradeoffs in dynamic fair division},
  author={Zeng, David and Psomas, Alexandros},
  booktitle={Proceedings of the 21st ACM Conference on Economics and Computation},
  pages={911--912},
  year={2020}
}

@inproceedings{Tripuraneni2020ProvableMO,
  title={Provable Meta-Learning of Linear Representations},
  author={Nilesh Tripuraneni and Chi Jin and Michael I. Jordan},
  booktitle={International Conference on Machine Learning},
  year={2020},
}

@inproceedings{du2021few,
title={Few-Shot Learning via Learning the Representation, Provably},
author={Simon Shaolei Du and Wei Hu and Sham M. Kakade and Jason D. Lee and Qi Lei},
booktitle={International Conference on Learning Representations},
year={2021},
}

@article{baxter2000model,
  title={A model of inductive bias learning},
  author={Baxter, Jonathan},
  journal={Journal of artificial intelligence research},
  volume={12},
  pages={149--198},
  year={2000}
}

@article{maurer2016benefit,
  title={The benefit of multitask representation learning},
  author={Maurer, Andreas and Pontil, Massimiliano and Romera-Paredes, Bernardino},
  journal={Journal of Machine Learning Research},
  volume={17},
  number={81},
  pages={1--32},
  year={2016}
}

@article{Yu2014AUV,
  title={A useful variant of the Davis--Kahan theorem for statisticians},
  author={Yi Yu and Tengyao Wang and Richard J. Samworth},
  journal={Biometrika},
  year={2014},
  volume={102},
  pages={315-323},
}

@book{vershynin2018high,
  title={High-dimensional probability: An introduction with applications in data science},
  author={Vershynin, Roman},
  volume={47},
  year={2018},
  publisher={Cambridge university press}
}

@article{hsu2012tail,
  author = {Daniel Hsu and Sham Kakade and Tong Zhang},
  title = {A tail inequality for quadratic forms of subgaussian random vectors},
  volume = {17},
  journal = {Electronic Communications in Probability},
  publisher = {Institute of Mathematical Statistics and Bernoulli Society},
  pages = {1 -- 6},
  year = {2012},
}

@inproceedings{xiong2023iterative,
  title={Iterative preference learning from human feedback: Bridging theory and practice for {RLHF} under {KL}-constraint},
  author={Xiong, Wei and Dong, Hanze and Ye, Chenlu and Wang, Ziqi and Zhong, Han and Ji, Heng and Jiang, Nan and Zhang, Tong},
  booktitle={ICLR 2024 Workshop on Mathematical and Empirical Understanding of Foundation Models},
  year={2023}
}

@inproceedings{siththaranjan2024distributional,
title={Distributional Preference Learning: Understanding and Accounting for Hidden Context in {RLHF}},
author={Anand Siththaranjan and Cassidy Laidlaw and Dylan Hadfield-Menell},
booktitle={The Twelfth International Conference on Learning Representations},
year={2024}
}

@article{zhou2023beyond,
  title={Beyond one-preference-for-all: Multi-objective direct preference optimization},
  author={Zhou, Zhanhui and Liu, Jie and Yang, Chao and Shao, Jing and Liu, Yu and Yue, Xiangyu and Ouyang, Wanli and Qiao, Yu},
  journal={arXiv preprint arXiv:2310.03708},
  year={2023}
}

@article{wang2024arithmetic,
  title={Arithmetic Control of {LLM}s for Diverse User Preferences: Directional Preference Alignment with Multi-Objective Rewards},
  author={Wang, Haoxiang and Lin, Yong and Xiong, Wei and Yang, Rui and Diao, Shizhe and Qiu, Shuang and Zhao, Han and Zhang, Tong},
  journal={arXiv preprint arXiv:2402.18571},
  year={2024}
}

@inproceedings{chakraborty2024maxmin,
  title={{M}ax{M}in-{RLHF}: Alignment with Diverse Human Preferences},
  author={Chakraborty, Souradip and Qiu, Jiahao and Yuan, Hui and Koppel, Alec and Manocha, Dinesh and Huang, Furong and Bedi, Amrit Singh and Wang, Mengdi},
  booktitle={Proceedings of the 41st International Conference on Machine Learning},
  series={Proceedings of Machine Learning Research},
  volume={235},
  pages={6116--6135},
  year={2024},
  publisher={PMLR},
  url={https://proceedings.mlr.press/v235/chakraborty24b.html}
}

@article{casper2023open,
  title={Open problems and fundamental limitations of reinforcement learning from human feedback},
  author={Casper, Stephen and Davies, Xander and Shi, Claudia and Gilbert, Thomas Krendl and Scheurer, J{\'e}r{\'e}my and Rando, Javier and Freedman, Rachel and Korbak, Tomasz and Lindner, David and Freire, Pedro and others},
  journal={arXiv preprint arXiv:2307.15217},
  year={2023}
}

@inproceedings{siddique2023fairness,
title={Fairness in Preference-based Reinforcement Learning},
author={Umer Siddique and Abhinav Sinha and Yongcan Cao},
booktitle={ICML 2023 Workshop The Many Facets of Preference-Based Learning},
year={2023},
}

@article{mandal2022socially,
  title={Socially fair reinforcement learning},
  author={Mandal, Debmalya and Gan, Jiarui},
  journal={arXiv preprint arXiv:2208.12584},
  year={2022}
}

@inproceedings{ju2024achieving,
title={Achieving Fairness in Multi-Agent {MDP} Using Reinforcement Learning},
author={Peizhong Ju and Arnob Ghosh and Ness Shroff},
booktitle={The Twelfth International Conference on Learning Representations},
year={2024}
}

@InProceedings{pmlr-v119-siddique20a,
  title = 	 {Learning Fair Policies in Multi-Objective (Deep) Reinforcement Learning with Average and Discounted Rewards},
  author =       {Siddique, Umer and Weng, Paul and Zimmer, Matthieu},
  booktitle = 	 {Proceedings of the 37th International Conference on Machine Learning},
  pages = 	 {8905--8915},
  year = 	 {2020},
  editor = 	 {III, Hal Daumé and Singh, Aarti},
  volume = 	 {119},
  series = 	 {Proceedings of Machine Learning Research},
  month = 	 {13--18 Jul},
  publisher =    {PMLR},
}

@article{ogryczak2013compromise,
  title={A compromise programming approach to multiobjective Markov decision processes},
  author={Ogryczak, Wlodzimierz and Perny, Patrice and Weng, Paul},
  journal={International Journal of Information Technology \& Decision Making},
  volume={12},
  number={05},
  pages={1021--1053},
  year={2013},
  publisher={World Scientific}
}

@book{sen2018collective,
  title={Collective choice and social welfare},
  author={Sen, Amartya},
  year={2018},
  publisher={Harvard University Press}
}

@article{rame2023rewarded,
  title={Rewarded soups: towards pareto-optimal alignment by interpolating weights fine-tuned on diverse rewards},
  author={Rame, Alexandre and Couairon, Guillaume and Dancette, Corentin and Gaya, Jean-Baptiste and Shukor, Mustafa and Soulier, Laure and Cord, Matthieu},
  journal={Advances in Neural Information Processing Systems},
  volume={36},
  pages={71095--71134},
  year={2023}
}

@article{bose2024offline,
  title={Offline multi-task transfer {RL} with representational penalization},
  author={Bose, Avinandan and Du, Simon Shaolei and Fazel, Maryam},
  journal={arXiv preprint arXiv:2402.12570},
  year={2024}
}

@InProceedings{pmlr-v235-munos24a,
  title = 	 {{N}ash Learning from Human Feedback},
  author =       {Munos, Remi and Valko, Michal and Calandriello, Daniele and Gheshlaghi Azar, Mohammad and Rowland, Mark and Guo, Zhaohan Daniel and Tang, Yunhao and Geist, Matthieu and Mesnard, Thomas and Fiegel, C\^{o}me and Michi, Andrea and Selvi, Marco and Girgin, Sertan and Momchev, Nikola and Bachem, Olivier and Mankowitz, Daniel J and Precup, Doina and Piot, Bilal},
  booktitle = 	 {Proceedings of the 41st International Conference on Machine Learning},
  pages = 	 {36743--36768},
  year = 	 {2024},
  editor = 	 {Salakhutdinov, Ruslan and Kolter, Zico and Heller, Katherine and Weller, Adrian and Oliver, Nuria and Scarlett, Jonathan and Berkenkamp, Felix},
  volume = 	 {235},
  series = 	 {Proceedings of Machine Learning Research},
  month = 	 {21--27 Jul},
  publisher =    {PMLR},
  url = 	 {https://proceedings.mlr.press/v235/munos24a.html}
}

@inproceedings{park2024rlhf,
title={{RLHF} from Heterogeneous Feedback via Personalization and Preference Aggregation},
author={Chanwoo Park and Mingyang Liu and Dingwen Kong and Kaiqing Zhang and Asuman E. Ozdaglar},
booktitle={ICML 2024 Workshop: Aligning Reinforcement Learning Experimentalists and Theorists},
year={2024},
}

@inproceedings{NEURIPS2024_9328208f,
 author = {Ge, Luise and Halpern, Daniel and Micha, Evi and Procaccia, Ariel D. and Shapira, Itai and Vorobeychik, Yevgeniy and Wu, Junlin},
 booktitle = {Advances in Neural Information Processing Systems},
 editor = {A. Globerson and L. Mackey and D. Belgrave and A. Fan and U. Paquet and J. Tomczak and C. Zhang},
 pages = {80439--80465},
 publisher = {Curran Associates, Inc.},
 title = {Axioms for AI Alignment from Human Feedback},
 url = {https://proceedings.neurips.cc/paper_files/paper/2024/file/9328208f88ec69420031647e6ff97727-Paper-Conference.pdf},
 volume = {37},
 year = {2024}
}

@article{xiao2024algorithmic,
  title={On the Algorithmic Bias of Aligning Large Language Models with {RLHF}: Preference Collapse and Matching Regularization},
  author={Xiao, Jiancong and Li, Ziniu and Xie, Xingyu and Getzen, Emily and Fang, Cong and Long, Qi and Su, Weijie J.},
  journal={Journal of the American Statistical Association},
  volume={120},
  number={552},
  pages={2154--2164},
  year={2025},
  doi={10.1080/01621459.2025.2555067},
  url={https://doi.org/10.1080/01621459.2025.2555067}
}

@inproceedings{NEURIPS2024_7e670825,
 author = {Alamdari, Parand A. and Ebadian, Soroush and Procaccia, Ariel D.},
 booktitle = {Advances in Neural Information Processing Systems},
 editor = {A. Globerson and L. Mackey and D. Belgrave and A. Fan and U. Paquet and J. Tomczak and C. Zhang},
 pages = {68308--68329},
 publisher = {Curran Associates, Inc.},
 title = {Policy Aggregation},
 url = {https://proceedings.neurips.cc/paper_files/paper/2024/file/7e670825a578392891ad40e93931b1e3-Paper-Conference.pdf},
 volume = {37},
 year = {2024}
}

@article{liu2024dual,
  title={Dual active learning for reinforcement learning from human feedback},
  author={Liu, Pangpang and Shi, Chengchun and Sun, Will Wei},
  journal={arXiv preprint arXiv:2410.02504},
  year={2024}
}

@article{lee2024low,
  title={Low-Rank Contextual Reinforcement Learning from Heterogeneous Human Feedback},
  author={Lee, Seong Jin and Sun, Will Wei and Liu, Yufeng},
  journal={Journal of the American Statistical Association},
  volume={121},
  number={554},
  pages={1037--1050},
  year={2026},
  publisher={Taylor \& Francis},
  doi={10.1080/01621459.2026.2674413},
  url={https://doi.org/10.1080/01621459.2026.2674413}
}

@inproceedings{xiong2025projection,
  title={Projection Optimization: A General Framework for Multi-Objective and Multi-Group RLHF},
  author={Xiong, Nuoya and Singh, Aarti},
  booktitle={Proceedings of the 42nd International Conference on Machine Learning},
  series={Proceedings of Machine Learning Research},
  volume={267},
  pages={69017--69058},
  year={2025},
  publisher={PMLR},
  url={https://proceedings.mlr.press/v267/xiong25c.html}
}

@inproceedings{buening2025strategyproof,
  title={Strategyproof Reinforcement Learning from Human Feedback},
  author={Kleine Buening, Thomas and Gan, Jiarui and Mandal, Debmalya and Kwiatkowska, Marta},
  booktitle={Advances in Neural Information Processing Systems},
  volume={38},
  pages={101431--101464},
  year={2025},
  publisher={Curran Associates, Inc.},
  doi={10.52202/085713-3396},
  url={https://proceedings.neurips.cc/paper_files/paper/2025/file/92f43b1d33fae4aa1958f75317f0cec1-Paper-Conference.pdf}
}

@inproceedings{liu2025shared,
  title={A Shared Low-Rank Adaptation Approach to Personalized RLHF},
  author={Liu, Renpu and Wang, Peng and Li, Donghao and Shen, Cong and Yang, Jing},
  booktitle={Proceedings of the 28th International Conference on Artificial Intelligence and Statistics},
  series={Proceedings of Machine Learning Research},
  volume={258},
  pages={1405--1413},
  year={2025},
  publisher={PMLR},
  url={https://proceedings.mlr.press/v258/liu25d.html}
}

@inproceedings{pmlr-v235-swamy24a,
  title={A Minimaximalist Approach to Reinforcement Learning from Human Feedback},
  author={Swamy, Gokul and Dann, Christoph and Kidambi, Rahul and Wu, Steven and Agarwal, Alekh},
  booktitle={Proceedings of the 41st International Conference on Machine Learning},
  series={Proceedings of Machine Learning Research},
  volume={235},
  pages={47345--47377},
  year={2024},
  publisher={PMLR},
  url={https://proceedings.mlr.press/v235/swamy24a.html}
}

@inproceedings{pmlr-v267-wang25af,
  title={{MPO}: An Efficient Post-Processing Framework for Mixing Diverse Preference Alignment},
  author={Wang, Tianze and Gui, Dongnan and Hu, Yifan and Lin, Shuhang and Zhang, Linjun},
  booktitle={Proceedings of the 42nd International Conference on Machine Learning},
  series={Proceedings of Machine Learning Research},
  volume={267},
  pages={62958--62979},
  year={2025},
  publisher={PMLR},
  url={https://proceedings.mlr.press/v267/wang25af.html}
}

\appendix

\newpage


\appendix

\section{Further Details on Methods}\label{app:add}

\subsection{Social Welfare Function Details}\label{ap:swfdef}
Here, we provide additional details regarding social welfare functions to complement the discussion in Section \ref{sec:preswf}. In welfare economics, a social welfare function aggregates individual utilities towards collective welfare and thus guides social decisions to align with the group's preferences. A reasonable social welfare function that satisfies six desirable axioms has been shown to lie in a one-parameter family of isoelastic social welfare functions $W_\alpha$ defined in \eqref{eq:swfWa} (or its monotonically increasing transformation).

Among these functions, the Nash welfare function $(\alpha=0)$ stands out as the unique solution that additionally satisfies \textit{independence of individual scales(IIS)}; and the Utilitarian welfare function $(\alpha=1)$ is the unique one that additionally satisfies \textit{independence of individual zeros (IIZ)}. The definitions of these properties are provided below.

\begin{definition}[Independence of Individual Scales of Utilities (IIS)]
    A solution is independent of individual scales of utilities if, for $\forall m\in [M]$, transforming the $m$-th utility function $u_m$ to $c_mu_m$ with $c_m\in\mathbb R^+$ does not change the solution.
\end{definition} 
\begin{definition}[Independence of Individual Zeros of Utilities (IIZ)]
    A solution is independent of individual zeros of utilities if, 
    for $\forall m\in [M]$, transforming the $m$-th utility function $u_m$ to $c_m+u_m$ with $c_m\in\mathbb R$ does not change the solution.
\end{definition}
Additionally, taking the limit  $\alpha\rightarrow-\infty$ yields Egalitarian welfare function, which can be reduced to consider $\min_m u_m$ for simplicity. 
 
In the main article, we introduce the Nash welfare function as the primary illustrative example. The Nash welfare function is advantageous due to its IIS property, which eliminates the potential issues of dominance arising from differences in utility scales. Moreover, one can also consider the Nash bargaining version which introduces an objective notion of individual ``zero utility". This involves the normalization of individual zeros $\{u_m^0\}_{m=1}^M$, and is formulated as follows:

\begin{equation*}
    \arg\max_u \prod_{m=1}^M(u_m-u_m^0)^{1/M}, \text{ s.t. } u_m>u_m^0 \text{ for } \forall m\in[M].
\end{equation*}
Thus, Nash bargaining solution satisfies the IIZ property. 

\subsection{Connections Between Two Models}\label{ap:connect}
In this section, we outline the connection between the reward-based and general reward-free models.
The average-wise definition of the von Neumann winner in Section \ref{sec:pes-von} is grounded in the Utilitarian welfare function. Here, we provide the rationale behind this choice by considering a voting scenario. 
In the context of a voting election, each voter has an absolute preference ranking over the set of alternatives. For any pair of alternatives $a\neq a'$, we have $N_{s,m}(a,a')\in \{1,-1\}$ for $a \neq a'$. Aggregating over all voters, $N_s(a,a')$ thus represents the disparity between the count of individuals preferring $a$ over $a'$ and the count of those preferring $a'$ over $a$. Consequently, a von Neumann winner represents the outcome favored by a majority of voters. Here the normalization for $N_{s,m}$ is crucial to avoid dominance due to unequal sample sizes or preference weights.

We have claimed that, under the additional assumption of a reward-based model, the von Neumann winner in the general model consistently aligns with the solution of the Utilitarian welfare function when $M=1$ or $2$. We give a brief illustration as follows.
\begin{itemize}
    \item Case $M=1$: In this scenario, there exists a Condorcet winner \citep{fishburn1977condorcet} determined by the optimal reward. Consequently, the Condorcet winner that maximizes the reward function is trivially the von Neumann winner.

    \item Case $M=2$: We first establish the following inequality for any real numbers $x,y$,
    \begin{equation*}
        \dfrac{1}{2} (\dfrac{1}{1+e^{-x}}+\dfrac{1}{1+e^{-y}})\geq \dfrac{1}{2} \Leftrightarrow x+y\geq 0.
    \end{equation*}
    
    Next, we substitute $x=R_1(s,a_1)-R_1(s,a_0),y=R_2(s,a_1)-R_2(s,a_0)$, where $R_m(s,a)$ denotes the reward function for the $m$-th party. Using the result above, we obtain:
    \begin{equation*}
        \dfrac{1}{2}\left[\mathbb P_1(a_1\succ a_0|s)+\mathbb P_2(a_1\succ a_0|s)\right]\geq \dfrac{1}{2} \Leftrightarrow R_1(s,a_1)+R_2(s,a_1)\geq R_1(s,a_0)+R_2(s,a_0).
    \end{equation*}
    This equivalence demonstrates that the aggregate preference between any action pair, based on the average of the rewards across both parties, determines the von Neumann winner.
\end{itemize}

\subsection{Generalization to MDPs}\label{sec:mdp}
We now generalize our results for the reward-based model in Section \ref{sec:rwd} to MDPs where the preference is based on a whole trajectory instead of a single action. 
Examples of such settings include Atari games and robot locomotion \citep{NIPS2017_d5e2c0ad}, where the preference depends on historical behaviors. 
Consider a MDP with $M$ individuals (parties), each of whom has its own reward function. Formally, we consider the environment $E=(\mathcal{S},\mathcal{A},H,\{P_h\}_{h=1}^{H-1},\{R_m\}_{m=1}^M,\rho)$, where $\mathcal{S}$ is a state space, $\mathcal{A}$ is an action space, $H$ is the horizon length, $P_h:\mathcal{S}\times\mathcal{A}\rightarrow\Delta(\mathcal{S})$ for $h\in[H-1]$ are known transition kernels, $R_m:\mathcal{S}\times\mathcal{A}\rightarrow \mathbb R$ is the reward function for the $m$-th individual (or party), and $\rho:\mathcal{S}\rightarrow \Delta(\mathcal{S})$ is an initial state distribution. 
When $H=1$ and there are no transitions, it is reduced to the contextual bandit (CB) setting.
Define the occupancy measure $d^{\pi}:\mathcal{S}\rightarrow \mathbb R$ as $d^{\pi}(s)=\sum_{h=1}^H\mathbb P_h(s_h=s|\pi)$, which becomes $\rho(s)$ when $H=1$.
We are provided with a dataset $\mathcal{D}=\bigcup_{m=1}^M\mathcal{D}_m$, where $\mathcal{D}_m=\{(s_{m}^i,\tau_{m,1}^i,\tau_{m,0}^i,y_{m}^i)\}_{i=1}^{n}$ is the pairwise comparison dataset for the $m$-th individual.
For the $i$-th sample in $\mathcal{D}_m$, an initial state $s_{m}^i$ is sampled from $\rho$ and two trajectories $\tau_{m,1}^i=(a_{m,1}^i,\cdots,s_{m,H}^i,a_{m,H}^i), \tau_{m,0}^i=(a_{m,1}^{i'},\cdots,s_{m,H}^{i'},a_{m,H}^{i'})$  are sampled from a distribution $g_{\tau}(\tau_{m,1}^i;\tau_{m,0}^i|s_{m}^i)$. Thus $(s_m^i,\tau_{m,1}^i,\tau_{m,0}^i)$ is generated by a generation distribution with density $g=g_\tau(\cdot|s)\rho(s)$.
We then observe a binary outcome $y_m^i$ from a Bernoulli distribution $\mathbb P_m(y_m^i=1|\tau_{m,1}^i,\tau_{m,0}^i)$ as follows,
\begin{equation*}
    \mathbb P_m(y=1|\tau_{1},\tau_{0},s)=\Phi(\sum_{h=1}^H R_{m,\theta_m^*}(s_h,a_h)-\sum_{h=1}^H R_{m,\theta_m^*}(s_h',a_h')),
\end{equation*}
where Assumption \ref{as:well},\ref{as:feature} hold with the feature difference in $\Sigma_*$ changed to the cumulative difference over the trajectory, i.e.,
\begin{equation}\label{eq:newsigma}
    \Sigma_*=\mathbb E_g(\sum_{h=1}^H(\phi(s_h,a_h)-\phi(s_h',a_h')))(\sum_{h=1}^H(\phi(s_h,a_h)-\phi(s_h',a_h')))^{\top}.
\end{equation}


The generalized algorithm for MDPs is similar to Algorithm \ref{alg1}. However, here we adopt a trajectory-based preference model and thus optimize a trajectory-wise value function. We are going to discuss the differences in detail below.

Denote $X_{m,i}=\sum_{h=1}^H \left(\phi(s_{m,h}^i,a_{m,h}^i)-\phi(s_{m,h}^{i'},a_{m,h}^{i'})\right)$ now as the difference between the cumulative feature in two trajectories. Define the data covariance matrices as $\Sigma_m=\dfrac{1}{n}\sum X_{m,i}X_{m,i}^\top$ for $\forall m\in [M]$. 
We again minimize the negative log-likelihood:
\begin{align*}
   \hat{U},\hat{\alpha} \in \arg\min_{U\in\mathcal{O}_{d\times r}, \|\alpha_m\|_2\leq B}l_{\mathcal{D}}(U,\alpha),
   \end{align*}
$$l_{\mathcal{D}}(U,\alpha)=-\dfrac{1}{Mn}\sum_{m=1}^M\sum_{i=1}^{n}\Big\{\mathbf{1}(y_m^i=1)\log\big(\Phi(\langle U\alpha_m,X_{m,i}\rangle)\big)    +\mathbf{1}(y_m^i=0)\log\big(\Phi(\langle U\alpha_m,-X_{m,i}\rangle)\big)\Big\}.$$

Subsequently, we use Nash welfare function to aggregate individual preferences.
Here we point out that the key difference between MDPs and CBs is the inclusion of the trajectory-based measure $d^{\pi}$, which stems from the observation \citep{zhu2023principled} that: 
$$\mathbb E_{s\sim\rho}[V_m^\pi(s)]=\mathbb E_{s\sim d^\pi} R_{m,\theta_m^*}(s,{\pi}(s)),$$
where $V_m^\pi(s)=\mathbb E[\sum\limits_{h=1}^H R_m(s_h,a_h)|s_1=s,\pi]$, and this expectation $\mathbb E$ is taken over the trajectory generated according to the transition kernel. Since the transition distribution $P$ is known, we can directly calculate $d^\pi$ for any given $\pi$. Define the trajectory-wise true and pessimistic Nash welfare function $J$ and $\hat{J}$ as follows: 
\begin{align*}
    J({\pi})&=\prod_{m=1}^M[\mathbb E_{s\sim d^\pi}R_{m,\theta_m^*}(s,{\pi}(s))]^{1/M},\\ 
    \hat{J}(\pi)&=\min_{\Theta\in\mathbf{\Theta}_B(\hat{\Theta})}\prod_{m=1}^M[\mathbb E_{s\sim d^\pi}R_{m,\theta_m}(s,{\pi}(s))]^{1/M},
\end{align*}
where $\mathbf{\Theta}_B(\hat{\theta}_m)=\{\theta\in\mathbf{\Theta}_B:\|\theta-\hat{\theta}_m\|_{\Sigma_m}\leq K'$ $\sqrt{\dfrac{r^2}{n}+\dfrac{dr^2\log{d}+r\log(M/\delta)}{Mn}}\}$, $K'$ is a fixed parameter.
Define $C^*=\max_m\mathbb E_{s\sim d^{\pi^*}}\|(\Sigma_m^{-1/2}\phi(s,\pi^*(s)))\|_2$. Similarly, we assume that the optimal policy value for every party is bounded below by a positive constant, i.e., $\mathbb E_{s\sim d^{\pi^*}} R_{m,\theta_m^*}(s,\pi^*(s))\geq r_0$ for $\forall m \in[M]$.  We have the following result, whose proof is deferred to Appendix \ref{pf:mdp}. 
\begin{theorem}\label{th:mdp-pes}
    Under the same condition as Theorem \ref{th:metatheta} (with $\Sigma_*$ replaced by  \eqref{eq:newsigma}), we have, then with probability at least $1-\delta$, for any $m\in [M]$,
    \begin{equation*}
        J(\pi^*)-J(\hat{\pi})\lesssim C^*\sqrt{\dfrac{r^2}{n}+\dfrac{dr^2\log{d}+r\log(M/\delta)}{Mn}}.
    \end{equation*}
\end{theorem}
The sub-optimality scales with the concentratability coefficient $C^*$. The primary distinction between Theorem \ref{th:mdp} and Theorem \ref{th:pes+} lies in the trajectory-wise form of the concentratability coefficient $C^*$. We can also attain efficiency and fairness guarantees similar to Section \ref{sec:social} accordingly.

\section{Technical Details for Reward-Based Model}\label{ap:rwd}
We denote the low-dimensional parameters and the original parameters as follows:
\begin{align*}
    &\alpha=(\alpha_1,\cdots,\alpha_M), \quad \alpha^*=(\alpha_1^*,\cdots,\alpha_M^*), \quad \hat{\alpha}=(\hat{\alpha}_1,\cdots,\hat{\alpha}_M) \in\mathbb R^{r\times M},\\
    &\Theta=(\theta_1,\cdots,\theta_M), \quad \Theta^*=(\theta_1^*,\cdots,\theta_M^*), \quad \hat{\Theta}=(\hat{\theta}_1,\cdots,\hat{\theta}_M)=(\hat{U}\hat{\alpha}_1,\cdots,\hat{U}\hat{\alpha}_M)\in\mathbb R^{d\times M}.
\end{align*}

We denote the feature difference vectors matrices as follows: 
\begin{equation*}
    X_{m,i}=\phi(s_m^i,a_{m,1}^i)-\phi(s_m^i,a_{m,0}^i)\in\mathbb R^{d},\quad 
    X_m=(X_{m,1},\cdots,X_{m,n})^\top \in\mathbb R^{n \times d}, \quad \forall m,i.
\end{equation*}

Through out this section, we define $\bm{\theta},\hat{\bm{\theta}},\bm{\theta}^*\in\mathbb R^{dM}$ as the vectorization of $\Theta,\hat{\Theta},\Theta^*$. We write $\Sigma \succeq(\preceq) \Sigma'$ if $\Sigma-\Sigma'$ is positive (negative) semidefinite.

\subsection{Proof of Theorem \ref{th:metatheta}}\label{pf:meta}
\begin{lemma}[Covariance concentration]\label{le:boundsigma}
    Suppose Assumption \ref{as:feature} holds. If the number of samples satisfies $n\gg d+\log(M/\delta)$ for $\delta\in (0,1)$, then with probability at least $1-\delta/10$, we have: for any $m \in [M]$,
    \begin{equation}\label{eq:boundsigma}
        0.9\Sigma_*\preceq \Sigma_m \preceq 1.1\Sigma_*.
    \end{equation}
\end{lemma}
\begin{proof}
    First, we fix $m \in [M]$ and recall the definition $X_{m,i}:=\phi(s_m^i,a_{m,1}^i)-\phi(s_m^i,a_{m,0}^i)$. It follows that $\Sigma_m=\dfrac{1}{n}\sum_{i=1}^{n}X_{m,i}X_{m,i}^\top=\dfrac{1}{n}X_m^\top X_m$, and its expectation satisfies $\mathbb E_g \Sigma_m=\Sigma_*$. Since $\phi$ is bounded from the assumption, $X_{m,i}$ $i\in[n]$ are independent sub-Gaussian random vectors. Consequently, with probability at least $1-2\exp(-t^2)$, 
    \begin{align*}
        \|\Sigma_m-\Sigma_*\|&=\|\dfrac{1}{n}\sum_{i=1}^{n}X_{m,i}X_{m,i}^\top-\Sigma_*\|\\
        &\lesssim \|\Sigma_*\| \max(\eta,\eta^2), \quad \eta=\sqrt{\dfrac{d}{n}}+\dfrac{t}{\sqrt{n}}.
    \end{align*}
    See Thereom 4.6.1 in \cite{vershynin2018high} or Lemma 7 in \cite{Tripuraneni2020ProvableMO} for reference. Thus, since $\|\Sigma_*\|\leq C_{\text{max}}$, it follows that with probability at least  $1-\delta$,
    \begin{equation*}
        \|\Sigma_m-\Sigma_*\| \lesssim \sqrt{\dfrac{d+\log(1/\delta)}{n}}.
    \end{equation*}
    
    Next, we apply a union bound over all $m \in [M]$. We substitute $\delta$ with $\delta/M$, and obtain that, with probability at least $1-\delta$, for any $m \in [M]$
    \begin{equation*}
        \|\Sigma_m-\Sigma_*\| \lesssim \sqrt{\dfrac{d+\log(M/\delta)}{n}}    
    \end{equation*}
    Thus, the bound $\|\Sigma_m-\Sigma_*\|\leq 0.1 \|\Sigma_*\|$ holds for any $m \in [M]$ given $n\gg d+\log(M/\delta)$. 
\end{proof}

\begin{theorem}[Guarantee on source data]\label{th:source}
    Suppose Assumption \ref{as:well},\ref{as:feature} hold. If the number of samples satisfies $n\gg d+\log(M/\delta)$ for $\delta\in (0,1)$, then with probability at least $1-\delta/5$,
    \begin{equation}
        \sum_{m=1}^M\|X_m\hat{U}\hat{\alpha}_m-X_mU^*\alpha_m^*\|^2\lesssim Mr+rd\log {d}+\log{(1/\delta)}.
    \end{equation}
\end{theorem}
\begin{proof}
    We assume that \eqref{eq:boundsigma} is true, which happens with probability at least $1-\delta/10$ according to Lemma \ref{le:boundsigma}. 
    
    \textbf{Step 1: Optimality of $\hat{\bm{\theta}}$.}
    Consider the following loss function, along with its first and second derivatives, 
    \begin{align*}
        &l_{\mathcal{D}}(\bm{\theta})=-\dfrac{1}{Mn}\sum_{m=1}^M\sum_{i=1}^{n}\log[\mathbf{1}(y_m^i=1)\dfrac{1}{\exp(-\langle \theta_m,X_{m,i}\rangle)+1}+\mathbf{1}(y_m^i=0)\dfrac{\exp(-\langle \theta_m,X_{m,i}\rangle)}{\exp(-\langle \theta_m,X_{m,i}\rangle)+1}],\\
        &\nabla_{\theta_m}l_{\mathcal{D}}(\bm{\theta})=-\dfrac{1}{Mn}\sum_{i=1}^{n}\left[\mathbf{1}(y_m^i=1)\dfrac{\exp(-\langle \theta_m,X_{m,i}\rangle)}{\exp(-\langle \theta_m,X_{m,i}\rangle)+1}-\mathbf{1}(y_m^i=0)\dfrac{1}{\exp(-\langle \theta_m,X_{m,i}\rangle)+1}\right]X_{m,i},\\
        &\nabla_{\theta_m}^2l_{\mathcal{D}}(\bm{\theta})=\dfrac{1}{Mn}\sum_{i=1}^{n}\dfrac{\exp(-\langle \theta_m,X_{m,i}\rangle)}{(\exp(-\langle \theta_m,X_{m,i}\rangle)+1)^2}\cdot X_{m,i}X_{m,i}^\top.
    \end{align*}
    Thus we have,
    \begin{align*}
        &\nabla_{\theta}l_{\mathcal{D}}(\bm{\theta})=(\nabla_{\theta_1}l_{\mathcal{D}}(\bm{\theta})^\top,\cdots,\nabla_{\theta_M}l_{\mathcal{D}}(\bm{\theta})^\top)^\top,\\
        &\nabla_{\theta}^2 l_{\mathcal{D}}(\bm{\theta})=\text{diag}(\nabla_{\theta_1}^2 l_{\mathcal{D}}(\bm{\theta}),\cdots,\nabla_{\theta_M}^2 l_{\mathcal{D}}(\bm{\theta})).
    \end{align*}
    Regarding the second derivative, since  $|\langle \theta_m,X_{m,i}\rangle|\leq 2BL$, we have $\dfrac{\exp(-\langle \theta_m,X_{m,i}\rangle)}{(\exp(-\langle \theta_m,X_{m,i}\rangle)+1)^2}\geq \gamma$ where $\gamma=\dfrac{1}{2+\exp(-2BL)+\exp(2BL)}$. Therefore, we obtain the lower bound
    \begin{equation*}
        \nabla_{\theta_m}^2l_{\mathcal{D}}(\bm{\theta})\succeq \dfrac{\gamma}{Mn}\sum_{i=1}^{n}X_{m,i}X_{m,i}^\top= \dfrac{\gamma}{M} \Sigma_m.
    \end{equation*}
    Define $\Sigma_{\text{sum}}=\text{diag}(\Sigma_1,\cdots,\Sigma_M)\in\mathbb R^{Mn\times Mn}$,  we conclude from strong convexity that 
    \begin{equation*}
        l_{\mathcal{D}}(\hat{\bm{\theta}})-l_{\mathcal{D}}(\bm{\theta}^*)-\langle \nabla l_{\mathcal{D}}(\bm{\theta}^*),\hat{\bm{\theta}}-\bm{\theta}^*\rangle\geq \dfrac{\gamma}{2M} (\hat{\bm{\theta}}-\bm{\theta}^*)^\top\Sigma_{\text{sum}}(\hat{\bm{\theta}}-\bm{\theta}^*)=\dfrac{\gamma}{2M}\|\hat{\bm{\theta}}-\bm{\theta}^*\|_{\Sigma_{\text{sum}}}^2.
    \end{equation*}
    Here the matrix-induced norm is defined as $\|x\|_\Sigma=\sqrt{x^{\top}\Sigma x}$. In the following, we also use $\|Z\|_{\Sigma_{\text{sum}}}$ to denote the $\Sigma_{\text{sum}}$-norm of $Z$'s vectorization when $Z$ is a matrix. From the optimality of $\hat{\bm{\theta}}$, we have $l_{\mathcal{D}}(\hat{\bm{\theta}})-l_{\mathcal{D}}(\bm{\theta}^*)\leq 0$, it follows that 
    \begin{equation}\label{eq:convexl}
        -\langle \nabla l_{\mathcal{D}}(\bm{\theta}^*),\hat{\bm{\theta}}-\bm{\theta}^*\rangle\geq \dfrac{\gamma}{2M}\|\hat{\bm{\theta}}-\bm{\theta}^*\|_{\Sigma_{\text{sum}}}^2.
    \end{equation}
    Using Cauchy-Schwartz inequality immediately implies that
    \begin{equation}\label{eq:theta_gra_hes}
        \|l_{\mathcal{D}}(\bm{\theta}^*)\|_{\Sigma_{\text{sum}}^{-1}} \|\hat{\bm{\theta}}-\bm{\theta}^*\|_{\Sigma_{\text{sum}}}\geq \dfrac{\gamma}{2M}\|\hat{\bm{\theta}}-\bm{\theta}^*\|_{\Sigma_{\text{sum}}}^2.
    \end{equation}
    Regarding the first derivative, notice that $y_{m}^i|X_{m,i}\sim \texttt{Ber}\left(\dfrac{1}{\exp(-\langle \theta_m^*,X_{m,i}\rangle)+1}\right)$, we define a random vector $V_m\in \mathbb R^{n}$ with components $V_{m,i}$ as follows,
    \begin{align}\label{eq:defv}
        V_{m,i}&=\mathbf{1}(y_m^i=1)\dfrac{\exp(-\langle \theta_m^*,X_{m,i}\rangle)}{\exp(-\langle \theta_m^*,X_{m,i}\rangle)+1}-\mathbf{1}(y_m^i=0)\dfrac{1}{\exp(-\langle \theta_m^*,X_{m,i}\rangle)+1}\nonumber\\
        &=\left\{
        \begin{aligned}
        &\dfrac{\exp(-\langle \theta_m^*,X_{m,i}\rangle)}{\exp(-\langle \theta_m^*,X_{m,i}\rangle)+1} \quad \text{w.p. } \dfrac{1}{\exp(-\langle \theta_m^*,X_{m,i}\rangle)+1},\\
        &\dfrac{-1}{\exp(-\langle \theta_m^*,X_{m,i}\rangle)+1} \quad \text{w.p. } \dfrac{\exp(-\langle \theta_m^*,X_{m,i}\rangle}{\exp(-\langle \theta_m^*,X_{m,i}\rangle)+1}.
        \end{aligned}
        \right.
    \end{align}
    Here $V_m$ solely depends on $y_m$ when conditioned on $X_m$. From the definition of $V_m$, we have $\nabla_{\theta_m}l_{\mathcal{D}}(\bm{\theta}^*)=-\dfrac{1}{Mn}X_m^\top V_m$. So conditioned on $X_m$, $V_m$ is a random vector with independent components that satisfy $\mathbb E V_i=0, \|V_{m,i}\|\leq 1$, and is therefore sub-Gaussian. Denote $A_m=\dfrac{1}{n^2}X_m \Sigma_m^{-1}X_m^\top$, we can then establish the following bound \citep[see, Appendix B.1, ][]{zhu2023principled}:
    \begin{equation*}
        \text{tr}(A_m)\leq \dfrac{d}{n},\quad\text{tr}(A_m^2)\leq \dfrac{d}{n^2}, \quad \|A_m\|\leq \dfrac{1}{n}.
    \end{equation*}
    By applying Bernstein's inequality for sub-Gaussian random variables in a quadratic form with $x=(V_1^\top,\cdots,V_m^\top)^\top$ and $A=\text{diag}(A_1,\cdots,A_M)$ \citep[see, Thereom 2.1, ][]{hsu2012tail}, we directly obtain that, with probability at least $1-\delta/20$, 
    \begin{equation}\label{eq:gradi}
        M^2\|l_{\mathcal{D}}(\bm{\theta}^*)\|_{\Sigma_{\text{sum}}^{-1}}^2=\sum_{m=1}^M\dfrac{1}{n^2}V_m^\top X_m \Sigma_m^{-1}X_m^\top V_m  \lesssim \dfrac{Md+\log(1/\delta)}{n}.
    \end{equation}
    Combining \eqref{eq:gradi} with \eqref{eq:theta_gra_hes}, we obtain that, with probability at least $1-\delta/20$, 
    \begin{equation}
        \|\hat{\bm{\theta}}-\bm{\theta}^*\|_{\Sigma_{\text{sum}}}\lesssim \sqrt{\dfrac{Md+\log(1/\delta)}{n} }.
    \end{equation}
    This implies that
    \begin{align*} 
        Md+\log(1/\delta) &\gtrsim \sum_{m=1}^M(\hat{\theta}_m-\theta_m^*)^\top X_m^\top X_m(\hat{\theta}_m-\theta_m^*) \\
        &\geq n \sum_{m=1}^M(\hat{\theta}_m-\theta_m^*)^\top 0.9\Sigma_*(\hat{\theta}_m-\theta_m^*)   &(\text{using \eqref{eq:boundsigma}})\\
        &\geq 0.9nC_{\text{min}}\|\hat{\bm{\theta}}-\bm{\theta}^*\|^2  &(\text{using Assumption }\ref{as:feature})\\
        &\gtrsim  n  \|\hat{\bm{\theta}}-\bm{\theta}^*\|^2.
    \end{align*}
    Thus, with probability at least $1-\delta/20$, 
    \begin{equation}\label{eq:bound_theta_F}
        \|\hat{\Theta}-\Theta^*\|_F^2=\|\hat{\bm{\theta}}-\bm{\theta}^*\|^2\lesssim \sqrt{\dfrac{Md+\log(1/\delta)}{n} }.
    \end{equation}
    Here we use $\|\cdot\|_F$ to denote the Frobenius norm of a matrix. 

    \textbf{Step 2: Combining with low-dimensional assumption.}
    Notice that $\text{rank}(\hat{\Theta}-\Theta^*)\leq 2r$, so we can rewrite it as $\hat{\Theta}-\Theta^*=QR=(Q\mathbf{r}_1,\cdots,Q\mathbf{r}_M)$ where $Q\in\mathcal{O}_{d\times 2r},R=(\mathbf{r}_1,\cdots,\mathbf{r}_M)\in \mathbb R^{2r\times M}$. Thus the vectorization can be rewriten as $\hat{\bm{\theta}}-\bm{\theta}^*=[(Q\mathbf{r}_1)^\top,\cdots,(Q\mathbf{r}_M)^\top]^\top$
    For each $m\in [M]$ we further rewrite $X_mQ=Z_mW_m$ where $Z_m\in\mathcal{O}_{n\times 2r},W_m\in \mathbb R^{2r\times 2r}$. Then we have
    \begin{align}\label{eq:twosqrt} 
        -\langle \nabla l_{\mathcal{D}}(\bm{\theta}^*),\hat{\bm{\theta}}-\bm{\theta}^*\rangle&=\sum_{m=1}^M\dfrac{1}{Mn}V_m^\top X_m Q\mathbf{r}_m\nonumber\\
        &=\dfrac{1}{Mn}\sum_{m=1}^MV_m^\top Z_mW_m\mathbf{r}_m\nonumber\\
        &\leq\dfrac{1}{Mn}\sum_{m=1}^M\|Z_m^\top V_m\| \|W_m\mathbf{r}_m\|\nonumber\\
        &\leq\dfrac{1}{Mn}\sqrt{\sum_{m=1}^M\|Z_m^\top V_m\|^2} \sqrt{\sum_{m=1}^M\|W_m\mathbf{r}_m\|^2}\nonumber\\
        &=\dfrac{1}{Mn}\sqrt{\sum_{m=1}^M\|Z_m^\top V_m\|^2} \sqrt{\sum_{m=1}^M\|Z_mW_m\mathbf{r}_m\|^2}\nonumber\\
        &=\dfrac{1}{Mn}\sqrt{\sum_{m=1}^M\|Z_m^\top V_m\|^2} \sqrt{\sum_{m=1}^M\|X_mQ\mathbf{r}_m\|^2}
    \end{align}
    where the second term can be calculated as $\sum_{m=1}^M\|X_mQ\mathbf{r}_m\|^2=\sum_{m=1}^M\|X_m(\hat{\theta}_m-\theta_m^*)\|^2=n \|\hat{\bm{\theta}}-\bm{\theta}^*\|_{\Sigma_{\text{sum}}}$.

    \textbf{Step 3: Applying $\epsilon$-net argument.}
    Next, we are going to give a high-probability upper bound on $\sum_{m=1}^M\|Z_m^\top V_m\|^2$ using the randomness of $V_m$. Since $Z_m$ depends on $Q$ which depends on $V$, we use an $\epsilon$-net argument to cover all possible $Q\in\mathcal{O}_{d\times 2r}$.
    First, for any fixed $\Bar{Q}\in\mathcal{O}_{d\times 2r}$, we let $X_m\Bar{Q}=\Bar{Z}_m\Bar{W}_m$ where $\Bar{Z}_m\in\mathcal{O}_{n\times 2r}$. The $\Bar{Z}_m$'s defined in this way are independent of $V$.

    From the orthonormality of $\Bar{Z}_m$, we can establish the following equation:
    \begin{equation*}
        \text{tr}(\Bar{Z}_m\Bar{Z}_m^\top)=r,\quad\text{tr}(\Bar{Z}_m\Bar{Z}_m^\top)^2=r,\quad\|\Bar{Z}_m\Bar{Z}_m^\top\|=1
    \end{equation*}
    Again, use Bernstein’s inequality for sub-Gaussian random variables in the quadratic form as we did in \eqref{eq:gradi}, we obtain that, with probability at least $1-\delta'$, 
    \begin{equation*}
        \sum_{m=1}^M\|\Bar{Z}_m^\top V_m\|^2=\sum_{m=1}^M V_m \Bar{Z}_m\Bar{Z}_m^\top V_m\lesssim Mr+\log(1/\delta').
    \end{equation*}
    Combining this with \eqref{eq:twosqrt}, we obtain that 
    \begin{equation*}
        -\langle \nabla l_{\mathcal{D}}(\bm{\theta}^*),\Bar{Q}R\rangle\leq\sqrt{\dfrac{Mr+\log(1/\delta')}{M^2n}}\|\Bar{Q}R\|_{\Sigma_{\text{sum}}}.
    \end{equation*}

    Lemma A.5 in \cite{du2021few} implies that: there exists an $\epsilon$-net $\mathcal{N}$ of $\mathcal{O}_{d\times 2r}$ in Frobenius norm such that $|\mathcal{N}|\leq(\dfrac{6\sqrt{2r}}{\epsilon})^{2rd}$. Applying a union bound over $\mathcal{N}$, we obtain that, with probability at least $1-|\mathcal{N}|\delta'$,
    \begin{equation}\label{eq:unionz}
        -\langle \nabla l_{\mathcal{D}}(\bm{\theta}^*),\Bar{Q}R\rangle\leq\sqrt{\dfrac{Mr+\log(1/\delta')}{M^2n}}\|\Bar{Q}R\|_{\Sigma_{\text{sum}}}, \quad\forall \Bar{Q}\in\mathcal{N}.
    \end{equation}
    Choosing $\delta'=\dfrac{\delta}{20(\frac{6\sqrt{2r}}{\epsilon})^{2rd}}$, we obtain that \eqref{eq:unionz} holds with probability at least $1-\delta/20$.

    Next, we proceed to apply $\epsilon$-net. Let $\Bar{Q}\in\mathcal{N}$ such that $\|\Bar{Q}-Q\|_F\leq\epsilon$, then we have
    \begin{align}\label{eq:q-qr}
        \|(\Bar{Q}-Q)R\|_{\Sigma_{\text{sum}}}^2
        =&\dfrac{1}{n}\sum_{m=1}^M \|X_m(\Bar{Q}-Q)\mathbf{r}_m\|^2\nonumber\\
        \leq & \dfrac{1}{n}\sum_{m=1}^M \|X_m\|^2\|\Bar{Q}-Q\|^2\|\mathbf{r}_m\|^2\nonumber\\
        \leq &\|\Sigma_*\|\epsilon^2\sum_{m=1}^M \|\mathbf{r}_m\|^2\nonumber\\
        \leq & C_{\text{max}}\epsilon^2\|R\|_F^2\nonumber\\
        =&C_{\text{max}}\epsilon^2\|QR\|_F^2\nonumber\\
        \lesssim &\dfrac{Md+\log(1/\delta)}{n} \epsilon^2  & (\text{using \eqref{eq:bound_theta_F}})
    \end{align}

    \textbf{Step 4: Finishing the proof.} We have the following inequalities,
    \begin{align*}
        &\dfrac{\gamma}{2M}\|\hat{\bm{\theta}}-\bm{\theta}^*\|_{\Sigma_{\text{sum}}}^2\\
        \leq& -\langle \nabla l_{\mathcal{D}}(\bm{\theta}^*),QR \rangle  &(\text{using \eqref{eq:convexl}})\\
        =&-\langle \nabla l_{\mathcal{D}}(\bm{\theta}^*),\Bar{Q}R\rangle-\langle \nabla l_{\mathcal{D}}(\bm{\theta}^*),(\Bar{Q}-Q)R \rangle\\
        \lesssim& \sqrt{\dfrac{Mr+\log(1/\delta')}{M^2n}}\|\Bar{Q}R\|_{\Sigma_{\text{sum}}}+\|\nabla l_{\mathcal{D}}(\bm{\theta}^*)\|_{\Sigma_{\text{sum}}}^{-1}\|(\Bar{Q}-Q)R\|_{\Sigma_{\text{sum}}} & (\text{using \eqref{eq:unionz}})\\
        \lesssim &\sqrt{\dfrac{Mr+\log(1/\delta')}{M^2n}}(\|QR\|_{\Sigma_{\text{sum}}}+\|(\Bar{Q}-Q)R\|_{\Sigma_{\text{sum}}} )\\
        +& \sqrt{\dfrac{Md+\log(1/\delta)}{M^2n}}\|(\Bar{Q}-Q)R\|_{\Sigma_{\text{sum}}} & (\text{using \eqref{eq:gradi}})\\
        \leq& \sqrt{\dfrac{Mr+\log(1/\delta')}{M^2n}}\|QR\|_{\Sigma_{\text{sum}}}+\sqrt{\dfrac{Md+\log(1/\delta')}{M^2n}}\|(\Bar{Q}-Q)R\|_{\Sigma_{\text{sum}}} \quad(\text{using }r<d,\delta'\leq \delta)\\
        \lesssim &\sqrt{\dfrac{Mr+\log(1/\delta')}{M^2n}}\|QR\|_{\Sigma_{\text{sum}}}+\sqrt{\dfrac{Md+\log(1/\delta')}{M^2n}}\sqrt{\dfrac{Md+\log(1/\delta)}{n}}\epsilon & (\text{using \eqref{eq:q-qr}})\\
        \leq&\sqrt{\dfrac{Mr+\log(1/\delta')}{M^2n}}\|QR\|_{\Sigma_{\text{sum}}}+\dfrac{Md+\log(1/\delta')}{Mn}\epsilon.  &(\text{using }\delta'\leq \delta)
    \end{align*}
    Thus, we have
    \begin{equation*}
        \|QR\|_{\Sigma_{\text{sum}}}^2\lesssim M\left(\sqrt{\dfrac{Mr+\log(1/\delta')}{M^2n}}\|QR\|_{\Sigma_{\text{sum}}}+\dfrac{Md+\log(1/\delta')}{Mn}\epsilon \right).
    \end{equation*}
    Therefore, we let $\epsilon=r/d$ and recall that $\delta'=\dfrac{\delta}{20(\frac{6\sqrt{2r}}{\epsilon})^{2rd}}$, it follows that
    \begin{align*}
        \|QR\|_{\Sigma_{\text{sum}}}&\lesssim \max\left(\sqrt{\dfrac{Mr+\log(1/\delta')}{n}},\sqrt{\dfrac{Md+\log(1/\delta')}{n}\epsilon} \right)\\
        &\leq \sqrt{\dfrac{Mr+\log(1/\delta')}{n}}\\ 
        &=\sqrt{\dfrac{Mr+rd\log{(r/\epsilon)}+\log(1/\delta)}{n}}\\
        &\leq \sqrt{\dfrac{Mr+rd\log{d}+\log(1/\delta)}{n}}.
    \end{align*}
    Finally, notice that
    \begin{equation*}
        \|QR\|_{\Sigma_{\text{sum}}}^2=\sum_{m=1}^M(\hat{\theta}_m-\theta_m^*)^\top \Sigma_m(\hat{\theta}_m-\theta_m^*)=\dfrac{1}{n} \sum_{m=1}^M\|X_m\hat{U}\hat{\alpha}_m-X_mU^*\alpha_m^*\|^2,
    \end{equation*}
    and the high-probability results used in \eqref{eq:boundsigma}, \eqref{eq:bound_theta_F}, \eqref{eq:unionz}, we obtain the needed inequality with a failure probability at most $\frac{\delta}{10}+\frac{\delta}{20}+\frac{\delta}{20}=\frac{\delta}{5}$. This concludes our proof.
\end{proof}

\begin{theorem}[Guarantee on target data]\label{th:target}
    Suppose Assumption \ref{as:well},\ref{as:feature} hold. If the number of samples satisfies $n\gg d+\log(M/\delta)$ for $\delta\in (0,1)$, then with probability at least $1-\delta$, for any $m\in [M]$,
    \begin{equation*}
        \|\hat{\theta}_m-\theta_m^*\|_{\Sigma_m}=\|\hat{U}\hat{\alpha}_m-U^*\alpha_m^*\|_{\Sigma_m}\lesssim \sqrt{\dfrac{r^2}{n}+\dfrac{dr^2\log{d}+r\log(M/\delta)}{Mn}}.
    \end{equation*}
\end{theorem}
\begin{proof}
    We assume Lemma \ref{le:boundsigma} and Theorem \ref{th:source} are true, which happens with probability at least $1-2\delta/5$. 

    \textbf{Step 1: Finding a rotation of $\hat{U}$ near $U^*$.}
    Theorem \ref{th:source} implies that
    \begin{equation*}
        \sum_{m=1}^M(\hat{\theta}_m-\theta_m^*)^\top \Sigma_m(\hat{\theta}_m-\theta_m^*)\lesssim \dfrac{Mr+rd\log {d}+\log{(1/\delta)}}{n}.
    \end{equation*}
    Thus, Lemma \ref{le:boundsigma} further shows that
    \begin{equation*}
        \sum_{m=1}^M(\hat{\theta}_m-\theta_m^*)^\top \Sigma_*(\hat{\theta}_m-\theta_m^*)\lesssim \dfrac{Mr+rd\log {d}+\log{(1/\delta)}}{n}.
    \end{equation*}
    Therefore, combining this with Assumption \ref{as:feature} implies that 
    \begin{equation*}
        \|\hat{\Theta}-\Theta^*\|_F^2\leq \dfrac{Mr+rd\log {d}+\log{(1/\delta)}}{n}.
    \end{equation*}

    Then, we bound the singular values of $\Theta^*=U^*\alpha^*$. Note that $\alpha^*$ is a $r\times M$ matrix, by definition, we have the smallest singular value $\nu=\sigma_r(\frac{\Theta^{*\top} \Theta^*}{M})=\sigma_r\left(\frac{\alpha^{*\top}\alpha^*}{M}\right)$, and the worst-case condition number as $\kappa=\sigma_1(\frac{\Theta^{*\top} \Theta^*}{M})/\nu=\sigma_1\left(\frac{\alpha^{*\top}\alpha^*}{M}\right)/\nu$. Note that when $\Theta$ is well-conditioned in the sense that $\nu>0$ and $\kappa\leq O(1)$, the assumption $\|\alpha_m^*\|=\Theta(1)$ implies that $\nu\geq \Omega(\frac{1}{r})$ \citep{Tripuraneni2020ProvableMO}. Therefore, we have the following inequality:
    \begin{equation*}
        \sqrt{\dfrac{\sigma_1(\Theta^{*\top} \Theta^*)}{\sigma_r^2(\Theta^{*\top} \Theta^*)}}=\dfrac{\sqrt{M\nu\kappa}}{M\nu}\leq \sqrt{\dfrac{r}{M}}.
    \end{equation*}
    
    Notice that $\hat{U}$ and $U^*$ are orthogonal. By applying the general Davis-Kahan $\sin \theta$ theorem \citep[see, Theorem 4, ][]{Yu2014AUV} for matrices $\hat{\Theta}$ and $\Theta^*$, it follows that, there exists an orthogonal matrix $\hat{O}\in\mathbb R^{r\times r}$ such that 
    \begin{align}\label{eq:rotate}
        \|\hat{U}\hat{O}-U^*\|_F&\lesssim  \sqrt{\dfrac{\sigma_1(\Theta^{*\top} \Theta^*)}{\sigma_r^2(\Theta^{*\top} \Theta^*)}} \min\left(r^{1/2}\|\hat{\Theta}-\Theta^*\|,\|\hat{\Theta}-\Theta^*\|_F\right) \nonumber\\
        &\lesssim\sqrt{\dfrac{r}{M}}  \sqrt{\dfrac{Mr+rd\log {d}+\log{(1/\delta)}}{n}}\nonumber\\
        &=\sqrt{\dfrac{r^2}{n}+\dfrac{dr^2\log{d}+r\log(1/\delta)}{Mn}}.
    \end{align}
    \textbf{Step 2: Bounding $\hat{\alpha}_m$.} For simplicity, suppose $\|\hat{U}-U^*\|_F\lesssim \sqrt{\dfrac{r^2}{n}+\dfrac{dr^2\log{d}+r\log(1/\delta)}{Mn}}$. We can rotate $\hat{U}$ to achieve this. Then $\hat{\alpha}_m$ is obtained by minimizing the following loss function:
    \begin{equation*}
        l_{\mathcal{D}_m}(\alpha_m)=-\dfrac{1}{n}\sum_{i=1}^{n}\log[\mathbf{1}(y_m^i=1)\dfrac{1}{\exp(-\langle \hat{U}\alpha_m,X_{m,i}\rangle)+1}+\mathbf{1}(y_m^i=0)\dfrac{\exp(-\langle \hat{U}\alpha_m,X_{m,i}\rangle)}{\exp(-\langle \hat{U}\alpha_m,X_{m,i}\rangle)+1}].
    \end{equation*}
    Its first and second derivatives are given by, 
    \begin{align*}
        &\nabla l_{\mathcal{D}_m}(\alpha_m)=-\dfrac{1}{n}\sum_{i=1}^{n}\left[\mathbf{1}(y_m^i=1)\dfrac{\exp(-\langle \hat{U}\alpha_m,X_{m,i}\rangle)}{\exp(-\langle \hat{U}\alpha_m,X_{m,i}\rangle)+1}-\mathbf{1}(y_m^i=0)\dfrac{1}{\exp(-\langle U\alpha_m,X_{m,i}\rangle)+1}\right]\hat{U}^\top X_{m,i},\\
        &\nabla^2 l_{\mathcal{D}_m}(\alpha_m)=\dfrac{1}{n}\sum_{i=1}^{n}\dfrac{\exp(-\langle \hat{U}\alpha_m,X_{m,i}\rangle)}{(\exp(-\langle \hat{U}\alpha_m,X_{m,i}\rangle)+1)^2}\cdot  \hat{U}^\top X_{m,i} X_{m,i}^\top \hat{U}.
    \end{align*}
    Similar to \eqref{eq:theta_gra_hes} in the proof of Theorem \ref{th:source}, we have the following bound
    \begin{equation}\label{eq:bounda}
        \|\hat{\alpha}_m-\alpha_m^*\|_{\hat{U}^\top\Sigma_m\hat{U}}\lesssim \|\nabla l_{\mathcal{D}_m}(\alpha_m^*)\|_{(\hat{U}^\top\Sigma_m\hat{U})^{-1}}.
    \end{equation}
    Here, the gradient is given by $\nabla l_{\mathcal{D}_m}(\alpha_m^*)=-\dfrac{1}{n}\hat{U}^\top X_m^\top\hat{V}_m$, where $\hat{V}_{m,i}=V_{m,i}+\Delta_i$, $V_{m,i}$ is defined in \eqref{eq:defv}, and $\Delta_i$ is defined as follows:
    \begin{equation*}
        \Delta_i=f_i(\hat{U})-f_i(U^*),\quad 
        f_i(U)=\dfrac{\exp(-\langle U\alpha_m^*,X_{m,i}\rangle)}{\exp(-\langle U\alpha_m^*,X_{m,i}\rangle)+1}.
    \end{equation*}
    We aim to bound $\Delta_i$ by considering the gradient of $f_i(U)$:
    \begin{equation*}
        \nabla f_i(U)=\dfrac{\exp(-\langle \hat{U}\alpha_m,X_{m,i}\rangle)}{(\exp(-\langle \hat{U}\alpha_m,X_{m,i}\rangle)+1)^2}\cdot   X_{m,i}\alpha_m^\top
    \end{equation*}
    Notice that $\|X_{m,i}\|_2\leq B$ and $\|\alpha_m\|_2=\Theta(1)$, therefore $\|X_{m,i}\alpha_m^\top\|_F^2=\|X_{m,i}\|_2^2\|\alpha_m\|_2^2$ is bounded, implying that the gradient $ \nabla f_i(U)$ is also bounded
    Thus,
    \begin{equation}\label{eq:deltai}
        |\Delta_i|=|f_i(\hat{U})-f_i(U^*)|\leq \|\hat{U}-U^*\|_F\|X_{m,i}\alpha_m^\top\|_F\lesssim \|\hat{U}-U^*\|_F.
    \end{equation}
    Therefore, it follows that
    \begin{align*}
        \|\nabla l_{\mathcal{D}_m}(\alpha_m^*)\|_{(\hat{U}^\top\Sigma_m\hat{U})^{-1}}&=\|-\dfrac{1}{n}\hat{U}^\top X_m^\top\hat{V}_m\|_{(\hat{U}^\top\Sigma_m\hat{U})^{-1}}\\
        &=\|-\dfrac{1}{n}\hat{U}^\top X_m^\top V_m\|_{(\hat{U}^\top\Sigma_m\hat{U})^{-1}}+\|-\dfrac{1}{n}\hat{U}^\top X_m^\top\Delta\|_{(\hat{U}^\top\Sigma_m\hat{U})^{-1}}.
    \end{align*}
    
    Regarding the first term, note that $c_{\min}I_{d}\preceq\Sigma_{m}\preceq c_{\max}I_{d}$, we have 
    \begin{equation*}
    \|-\dfrac{1}{n}\hat{U}^\top X_m^\top V_m\|_{(\hat{U}^\top\Sigma_m\hat{U})^{-1}}\lesssim \|\frac{1}{n}\hat{U}^{\top}X_{m}^{\top}V_{m}\|_{2},
    \end{equation*}
    and thus 
    \begin{equation*}
        \begin{aligned}
            &\|-\dfrac{1}{n}\hat{U}^\top X_m^\top V_m\|_{(\hat{U}^\top\Sigma_m\hat{U})^{-1}}\\
            \lesssim & \|\frac{1}{n}\hat{U}^{\top}X_{m}^{\top}V_{m}\|_{2}
            \\ \le& \frac{1}{n}\|U^{*\top}X_{m}^\top V_{m}\|_{2}+\frac{1}{n}\|(\hat{U}-U^*)^{\top}X_m^\top V_m\|_{2},
            \\ \le& \frac{1}{n}\|U^{*\top}X_{m}^\top V_{m}\|_{2}+\frac{1}{n}\|\hat{U}-U^*\|_{F}\|X_{m}^{\top}V_{m}\|_{2}
            \\ \lesssim & \sqrt{\dfrac{r+\log(1/\delta)}{n}} +\|\hat{U}-U^*\|_{F}\sqrt{\dfrac{d+\log(1/\delta)}{n}} 
            \qquad \text{(By Theorem 2.1 in \citep{hsu2012tail})}
            \\ \lesssim &\sqrt{\dfrac{r+\log(1/\delta)}{n}} + \sqrt{\dfrac{r^2}{n}+\dfrac{dr^2\log{d}+r\log(1/\delta)}{Mn}}\sqrt{\dfrac{d+\log(1/\delta)}{n}}
        \end{aligned}
    \end{equation*}
    Under the sample-size condition $n\gg d+\log(M/\delta)$, it is dominated by the first term.

    Regarding the second term, denote $\Tilde{X}_m=X_m\hat{U}$, it follows that
    \begin{align*}
        \|-\dfrac{1}{n}\hat{U}^\top X_m^\top\Delta\|_{(\hat{U}^\top\Sigma_m\hat{U})^{-1}}&=\sqrt{\dfrac{1}{n}\Delta^\top\Tilde{X}_m^\top (\Tilde{X}_m^\top\Tilde{X}_m)^{-1}\Tilde{X}_m\Delta}\\
        &\leq \frac{1}{\sqrt{n}}\|\Delta\|_2\\
        & \leq \max_{i}|\Delta_{i}|
        \\ &\lesssim {\|\hat{U}-U^*\|_F} &(\text{using \eqref{eq:deltai}})
    \end{align*}
    Thus, \eqref{eq:bounda} implies that
    \begin{equation}\label{eq:furfea}
        \|\hat{\alpha}_m-\alpha_m^*\|_{\hat{U}^\top\Sigma_m\hat{U}}\lesssim\|\nabla l_{\mathcal{D}_m}(\alpha_m^*)\|_{(\hat{U}^\top\Sigma_m\hat{U})^{-1}} \lesssim \sqrt{\dfrac{r+\log(1/\delta)}{n}}+ {\|\hat{U}-U^*\|_F}.
    \end{equation}

    \textbf{Step 3: Finishing the proof.}
    Therefore, we have the following inequalities,
    \begin{align*}
        \|\hat{U}\hat{\alpha}_m-U^*\alpha_m^*\|_{\Sigma_m}&\leq  \|\hat{U}\hat{\alpha}_m-\hat{U}\alpha_m^*\|_{\Sigma_m}+\|\hat{U}\alpha_m^*-U^*\alpha_m^*\|_{\Sigma_m}\\
        &=\|\hat{\alpha}_m-\alpha_m^*\|_{\hat{U}^\top\Sigma_m\hat{U}}+\sqrt{\alpha_m^{*\top}(\hat{U}-U^*)^\top\Sigma_m(\hat{U}-U^*)\alpha_m^*}\\
        &\lesssim \|\hat{\alpha}_m-\alpha_m^*\|_{\hat{U}^\top\Sigma_*\hat{U}}+\|\alpha_m^*\|_2\|(\hat{U}-U^*)^\top\Sigma_*(\hat{U}-U^*)\|^{1/2} & (\text{using \eqref{eq:boundsigma}})\\
        &\lesssim \|\hat{\alpha}_m-\alpha_m^*\|_{\hat{U}^\top\Sigma_*\hat{U}}+\|\hat{U}-U^*\|_F & (\text{using Assumption \ref{as:feature}})\\
        &\lesssim \sqrt{\dfrac{r+\log(1/\delta)}{n}}+ \|\hat{U}-U^*\|_F & (\text{using \eqref{eq:furfea}})\\
        &\lesssim \sqrt{\dfrac{r+\log(1/\delta)}{n}}+\sqrt{\dfrac{r^2}{n}+\dfrac{dr^2\log{d}+r\log(1/\delta)}{Mn}}& (\text{using \eqref{eq:rotate}})
        \\&{\lesssim \sqrt{\frac{1}{n}\big({r^2}+\dfrac{dr^2\log{d}}{M}+\log(1/\delta)\big)}}
    \end{align*}
    
    Notice that the high-probability results are used in Lemma \ref{le:boundsigma}, Theorem \ref{th:source}, and \eqref{eq:furfea}, thus we obtain the needed inequality with a failure probability at most $\frac{2\delta}{5}+\frac{\delta}{20}\leq\frac{3\delta}{5}$ for a fixed $m$. Finally, applying a union bound for $\forall m\in [M]$ and substituting $\delta$ with $\delta/M$ finish the proof.
\end{proof}

\subsection{Proof of Theorem \ref{th:pes+}}\label{pf:pes+}

\begin{lemma}\label{lm:geo}
    If $B\geq x_i,y_i\geq \epsilon>0$ for $\forall i\in[M]$, then
    \begin{equation*}
        |(x_1\cdots x_M)^{1/M}-(y_1\cdots y_M)^{1/M}|\leq \dfrac{B}{M\epsilon} \sum_{m=1}^M |x_m-y_m|.
    \end{equation*}
\end{lemma}

\begin{proof}
    Without loss of generality, we assume that $y_1\cdots y_M \geq x_1\cdots x_M$.
    \begin{align*}
        &(y_1\cdots y_M)^{1/M} - (x_1\cdots x_M)^{1/M}\\
        =&(x_1\cdots x_M)^{1/M}\left\{\left((1+\dfrac{y_1-x_1}{x_1})\cdots (1+\dfrac{y_M-x_M}{x_M})\right)^{1/M}-1\right\}\\
        \leq &(x_1\cdots x_M)^{1/M} \left\{\left((1+\dfrac{y_1-x_1}{x_1})+\cdots (1+\dfrac{y_M-x_M}{x_M})\right)/M-1\right\}& (\text{using mean inequality})\\
        \leq &\dfrac{B}{M} \left(\dfrac{y_1-x_1}{x_1}+\cdots +\dfrac{y_M-x_M}{x_M}\right)& (\text{using } x_i\leq B)\\
        \leq &\dfrac{B}{M}  \left((\dfrac{|y_1-x_1|}{x_1}+\cdots +\dfrac{|y_M-x_M|}{x_M})\right)\\
        \leq & \dfrac{B}{M}  \left((\dfrac{|y_1-x_1|}{\epsilon}+\cdots +\dfrac{|y_M-x_M|}{\epsilon})\right) & (\text{using } x_i\geq \epsilon)\\
        = & \dfrac{B}{M\epsilon}\sum_{m=1}^M |x_m-y_m|.
    \end{align*}
    We conclude the proof.
\end{proof}

We then provided the detailed proof of Theorem \ref{th:pes+}.

\begin{proof}
    Decompose the sub-optimality into three terms:
    \begin{equation*}
        J(\pi^*)-J(\hat{\pi})=\left(J(\pi^*)-\hat{J}(\pi^*)\right)+\left(\hat{J}(\pi^*)-\hat{J}(\hat{\pi})\right)+\left(\hat{J}(\hat{\pi})-J(\hat{\pi})\right).
    \end{equation*}
    The second term, $\hat{J}(\pi^*)-\hat{J}(\hat{\pi})\leq 0$, holds directly by the definition of $\hat{\pi}$, as $\hat{\pi}$ maximizes $\hat{J}$. From Theorem \ref{th:metatheta}, with probability at least $1-\delta$, $\theta_m^*\in\mathbf{\Theta}_B(\hat{\theta}_m)$ for $\forall m$. This allows us to bound the third term as follows:
    \begin{align*}
        \hat{J}(\hat{\pi})-J(\hat{\pi}) =\min_{\Theta\in\mathbf{\Theta}_B(\hat{\Theta})}\prod_{m=1}^M [\mathbb E_{s\sim\rho}R_{m,{\theta}_m}(s,\hat{\pi}(s))]^{1/M}-\prod_{m=1}^M [\mathbb E_{s\sim\rho} R_{m,\theta_m^*}(s,\hat{\pi}(s))]^{1/M}\leq 0.
    \end{align*}

    Thus we can focus on the first term. Let $(\Tilde{\theta}_1,\cdots,(\Tilde{\theta}_M)= \arg\min_{\Theta\in\mathbf{\Theta}_B(\hat{\Theta})}\prod_{m=1}^M[\mathbb E_{s\sim\rho} R_{m,{\theta}_m}(s,\pi^*(s))]^{1/M}$, which represents the pessimistic parameter chosen from lower confidence bound for the optimal policy $\pi^*$.

    From the assumption, we have $\mathbb E_{s\sim\rho} R_{m,{\theta}_m^*}(s,\pi^*(s))\geq r_0$. Besides, with probability at least $1-\delta$, we have $\Tilde{\theta}_m,\theta_m^*\in\mathbf{\Theta}_B(\hat{\theta}_m)$ for $\forall m$. Using this, we can compare the reward functions evaluated at $\Tilde{\theta}_m$ and $\theta_m^*$:
    \begin{align*}
        |R_{m,\Tilde{\theta}_m}(s,\pi^*(s))-R_{m,{\theta}_m^*}(s,\pi^*(s))|
        &=|({\theta}_m^*-\Tilde{\theta}_m)^\top \phi(s,\pi^*(s))|\\
        &\leq \|{\theta}_m^*-\Tilde{\theta}_m\|_2 \|\phi(s,\pi^*(s))\|_2\\
        &\lesssim  \|{\theta}_m^*-\Tilde{\theta}_m\|_{\Sigma_m} \|\phi(s,\pi^*(s))\|_2\\
        &\lesssim \sqrt{\dfrac{r^2}{n}+\dfrac{dr^2\log{d}+r\log(M/\delta)}{Mn}}.
    \end{align*}
    
    The last inequality holds based on the fact that $\Tilde{\theta}_m\in\mathbf{\Theta}_B(\hat{\theta}_m)$ and $\theta_m^*\in\mathbf{\Theta}_B(\hat{\theta}_m)$ with probability at least $1-\delta$ for $\forall m$, as well as the boundedness of $\phi$. Thus, for sufficiently large $n$, we also have 
    $\mathbb E_{s\sim\rho} R_{m,\Tilde{\theta}_m}(s,\pi^*(s))\geq r_0/2$.
    Therefore, with probability at least $1-\delta$, we can conclude that
    \begin{align*}
        &J(\pi^*)-J(\hat{\pi})\leq J(\pi^*)-\hat{J}(\pi^*)\\    =&\max_{\Theta\in\mathbf{\Theta}_B(\hat{\Theta})}\left(\prod_{m=1}^M [\mathbb E_{s\sim\rho} R_{m,{\theta}_m^*}(s,\pi^*(s))]^{1/M}-\prod_{m=1}^M [\mathbb E_{s\sim\rho}R_{m,{\theta}_m}(s,\pi^*(s))]^{1/M}\right) \\
        =&\prod_{m=1}^M [\mathbb E_{s\sim\rho} R_{m,{\theta}_m^*}(s,\pi^*(s))]^{1/M}-\prod_{m=1}^M [\mathbb E_{s\sim\rho}R_{m,\Tilde{\theta}_m}(s,\pi^*(s))]^{1/M}\\
        \lesssim& \dfrac{1}{M}\sum_{m=1}^M |\mathbb E_{s\sim\rho} R_{m,{\theta}_m^*}(s,\pi^*(s))-\mathbb E_{s\sim\rho}R_{m,\Tilde{\theta}_m}(s,\pi^*(s))| \qquad \text{(using Lemma \ref{lm:geo})}\\
        =&\dfrac{1}{M} \sum_{m=1}^M \left|({\theta}_m^*-\Tilde{\theta}_m)^{\top}\mathbb E_{s\sim\rho}\phi(s,\pi^*(s))\right|.
    \end{align*}
    
    Separately, for each $m$ we have
    \begin{align*}
        &\left|({\theta}_m^*-\Tilde{\theta}_m)^{\top}\mathbb E_{s\sim\rho}\phi(s,\pi^*(s))\right|\\
        =&\left|({\theta}_m^*-\hat{\theta}_m+\hat{\theta}_m-\Tilde{\theta}_m)^{\top}\mathbb E_{s\sim\rho}\phi(s,\pi^*(s))\right|\\
        \leq&\left|({\theta}_m^*-\hat{\theta}_m)^{\top}\mathbb E_{s\sim\rho}\phi(s,\pi^*(s))\right|+\left|(\hat{\theta}_m-\Tilde{\theta}_m)^{\top}\mathbb E_{s\sim\rho}\phi(s,\pi^*(s))\right|\\
        \leq&\|{\theta}_m^*-\hat{\theta}_m\|_{\Sigma_m}\|\mathbb E_{s\sim\rho}\phi(s,\pi^*(s))\|_{\Sigma_m^{-1}}+\|\hat{\theta}_m-\Tilde{\theta}_m\|_{\Sigma_m}\|\mathbb E_{s\sim\rho}\phi(s,\pi^*(s))\|_{\Sigma_m^{-1}} \quad \text{(Cauchy-Schwartz inequality)}
        \\ \lesssim&  {\sqrt{\frac{1}{n}\big({r^2}+\dfrac{dr^2\log{d}}{M}+\log(M/\delta)\big)}}\|(\Sigma_m^{-1/2}\mathbb E_{s\sim\rho}\phi(s,\pi^*(s)))\|_2
    \end{align*}
    The last inequality holds because $\Tilde{\theta}_m\in\mathbf{\Theta}_B(\hat{\theta}_m)$ and $\theta_m^*\in\mathbf{\Theta}_B(\hat{\theta}_m)$ with probability at least $1-\delta$ for $\forall m$. 
   Therefore, we conclude that
   \begin{equation*}
       J(\pi^*)-J(\hat{\pi}) 
       {\lesssim \sqrt{\frac{1}{n}\big({r^2}+\dfrac{dr^2\log{d}}{M}+\log(M/\delta)\big)}}C^*.
   \end{equation*}
   where $C^*=\max_m\|\Sigma_m^{-1/2}\mathbb E_{s\sim\rho} \phi(s,\pi^*(s))\|_2.$.
\end{proof}
Alternatively, one can consider another logarithm form of the Nash welfare function as in \cite{ju2024achieving}.

\subsection{Proof of Theorem \ref{th:nash-lb}}\label{pf:nash-lb}
\begin{proof}
    The proof is adapted from Theorem 3.10 of \cite{zhu2023principled}.
    In this context, we consider $\Tilde{\phi}=U^{*\top} \phi\in \mathbb R^r$ for $U^{*}=(\mathbf{I}_r \quad \mathbf{0})^\top\in\mathcal{O}_{d\times r}$, and the original feature $\phi$ can be directly reconstructed from $\Tilde{\phi}$. Without loss of generality, assume that $r/3$ is some integer. We set $\mathcal{S}=\{1,\cdots,r/3\},\mathcal{A}=\{a_1,a_2,a_3\}$, and $\rho=\texttt{Unif}([1,2,\cdots,|\mathcal{S}|])$. The feature functions and the number of observations for each state-action pair are defined as follows:
    \begin{align*}
        &\Tilde{\phi}(s,a_1)=e_{3s-1},\quad\Tilde{\phi}(s,a_2)=e_{3s-2},\quad\Tilde{\phi}(s,a_3)=0, \\
        &n(s,a_1,a_3)=(n/|\mathcal{S}|)(1-2/\mathcal{C}^2), \quad n(s,a_2,a_3)=(n/|\mathcal{S}|)(2/\mathcal{C}^2).
    \end{align*}
    Let $v_{-1}=(1/r,1/r+\Delta,-2/r-\Delta),v_1=(1/r+2\Delta,1/r+\Delta,-2/r-3\Delta)$ where $\Delta=\mathcal{C}\sqrt{|\mathcal{S}|/n}$. We construct $2^{|\mathcal{S}|}$ instances, indexed by $\tau\in\{-1,1\}^{|\mathcal{S}|}$, where each $\alpha_\tau^*=( v_{\tau_1},\cdots,v_{\tau_{|\mathcal{S}|}})^\top$. Under each $\alpha_\tau$, the optimal policy $\pi^*(s)$ is either $a_1$ or $a_2$. It can be verified that $ \|\Sigma^{-\frac{1}{2}} \mathbb E_{s\sim\rho} \phi(s,\pi^*(s))\|_2\leq \mathcal{C}$ and $\|\theta_\tau^*\|=\|\alpha_{\tau}^*\|$ is bounded, given that $r>6,n\gtrsim r\mathcal{C}^2,\mathcal{C}\geq 2$. Here $\Sigma^{-\frac{1}{2}} $ is regarded as the Moore–Penrose pseudoinverse, which is well-defined here because $\mathbb E_{s\sim\rho} \phi(s,\pi^*(s))$ lies in the column space of $\Sigma$.
    By applying Assouad’s lemma as outlined in \cite{zhu2023principled}, we can conclude that there exists some $\alpha_{\tau},\tau\in\{-1,1\}^{|\mathcal{S}|}$ that achieves the lower bound for the case of $M=1$. 

    Now we construct $M$ copies of the above contextual bandit instance, denoted as $\mathcal{Q}_m, m\in [M]$. {For this lower-bound construction, the \(M\) party-specific datasets are perfectly coupled. Specifically, we first generate a single dataset \(\mathcal{D}\) containing \(n\) comparison observations, including their realized preference outcomes, and then set \(\mathcal{D}_1=\cdots=\mathcal{D}_M=\mathcal{D}\). Consequently, the learner observes only \(n\) distinct random comparisons rather than \(Mn\). This coupled data-generation paradigm is used exclusively for the lower-bound proof; throughout the rest of the paper, the party-specific datasets are generated independently.} We have shown that $C^*\leq \mathcal{C}$ and $\|\theta_m^*\|=\|\alpha_m^*\|$ are bounded. Thus, we can define an instance $\mathcal{Q}(M)\in\text{CB}(M,\mathcal{C}) $ where the multi-party instance $\mathcal{Q}(M)=\mathcal{Q}_1\times\cdots\times\mathcal{Q}_M$ represents the combination of each individual instance for a single party. For any $\hat{\pi}$, we have
    \begin{align*}
        \texttt{SubOpt}_{\mathcal{Q}(M)}(\hat{\pi}) 
        =&\prod_{m=1}^M [\mathbb E_{s\sim\rho} R_{m,{\theta}_m^*}(s,\pi^*(s))]^{1/M}- \prod_{m=1}^M [\mathbb E_{s\sim\rho} R_{m,{\theta}_m^*}(s,\hat{\pi}(s))]^{1/M}\\
        =&\mathbb E_{s\sim\rho} R_{{\theta}^*}(s,\pi^*(s))- \mathbb E_{s\sim\rho} R_{{\theta}^*}(s,\hat{\pi}(s))
         \gtrsim \mathcal{C}\sqrt{\dfrac{r}{n}}.
    \end{align*}
    which is exactly the result we aim to achieve. To ensure that the reward exceeds $r_0$, we can add one additional dimension to both the feature $\Tilde{\phi}$ and the parameter $\alpha$.
\end{proof}

\subsection{Proof of Theorem~\ref{th:pareto}}\label{pf:pareto}
\begin{proof}
    For simplicity, we will use $R(s,\pi)$ to denote $R(s,\pi(s))$. Consider another policy $\Tilde{\pi}$ which satisfies the condition $\mathbb E_{s\sim\rho}R_{m,\theta_m^*}(s,\Tilde{\pi})\geq \mathbb E_{s\sim\rho}R_{m,\theta_m^*}(s,\hat{\pi})$ for $\forall m<M$, then
    \begin{align*}
        &[\mathbb E_{s\sim\rho}R_{M,\theta_M^*}(s,\Tilde{\pi})]^{1/M}\\
        =&\dfrac{\prod_{m=1}^M[\mathbb E_{s\sim\rho}R_{m,\theta_m^*}(s,\Tilde{\pi})]^{1/M}}{\prod_{m=1}^{M-1}[\mathbb E_{s\sim\rho}R_{m,\theta_m^*}(s,\Tilde{\pi})]^{1/M}}\\
        \leq&\dfrac{\prod_{m=1}^M[\mathbb E_{s\sim\rho}R_{m,\theta_m^*}(s,{\pi}^*)]^{1/M}}{\prod_{m=1}^{M-1}[\mathbb E_{s\sim\rho}R_{m,\theta_m^*}(s,\hat{\pi})]^{1/M}} \qquad (\text{using optimality of $\pi^*$ and relationship of $\Tilde{\pi},\hat{\pi}$})\\
        \leq& \dfrac{\prod_{m=1}^M[\mathbb E_{s\sim\rho}R_{m,\theta_m^*}(s,{\pi}^*)]^{1/M}-\prod_{m=1}^M[\mathbb E_{s\sim\rho}R_{m,\theta_m^*}(s,\hat{\pi})]^{1/M}+\prod_{m=1}^M[\mathbb E_{s\sim\rho}R_{m,\theta_m^*}(s,\hat{\pi})]^{1/M}}{\prod_{m=1}^{M-1}[\mathbb E_{s\sim\rho}R_{m,\theta_m^*}(s,\hat{\pi})]^{1/M}}\\
        =&[\mathbb E_{s\sim\rho}R_{M,\theta_M^*}(s,\hat{\pi})]^{1/M}\left(1+\dfrac{\prod_{m=1}^M[\mathbb E_{s\sim\rho}R_{m,\theta_m^*}(s,{\pi}^*)]^{1/M}-\prod_{m=1}^M[\mathbb E_{s\sim\rho}R_{m,\theta_m^*}(s,\hat{\pi})]^{1/M}}{\prod_{m=1}^{M}[\mathbb E_{s\sim\rho}R_{m,\theta_m^*}(s,\hat{\pi})]^{1/M}}\right)\\
        \leq& [\mathbb E_{s\sim\rho}R_{M,\theta_M^*}(s,\hat{\pi})]^{1/M}\left(1+\dfrac{J(\pi^*)-J(\hat{\pi})}{J(\hat{\pi})}\right)\\
        \leq& [\mathbb E_{s\sim\rho}R_{M,\theta_M^*}(s,\hat{\pi})]^{1/M}\left(1+\dfrac{J(\pi^*)-J(\hat{\pi})}{J(\pi^*)-(J(\pi^*)-J(\hat{\pi}))}\right)\\
        \leq & {[\mathbb E_{s\sim\rho}R_{M,\theta_M^*}(s,\hat{\pi})]^{1/M}\left(1+\dfrac{K\sqrt{\frac{1}{n}\big({r^2}+\frac{dr^2\log{d}}{M}+\log(M/\delta)\big)}C^*}{r_0/2}\right)}.
    \end{align*}
    The last inequality holds due to Theorem \ref{th:pes+} and the fact that $J(\pi^*)\geq r_0$, for sufficiently large $n$. By raising both sides to the power of $M$, we obtained the desired conclusion.
\end{proof}

We further show that $\hat{\pi}$ satisfies the approximate Pigou-Dalton principle, which is an approximation of the traditional Pigou-Dalton principle \cite{moulin2004fair}.
\begin{definition}[$\tau$-approximate Pigou-Dalton Principle]
    A solution $\pi$ satisfies the $\tau$-approximate Pigou-Dalton principle, if for any policy $\pi'$ such that 
    \begin{itemize}
        \item $\mathbb E_{s\sim\rho}R_{i,\theta_i^*}(s,\pi'(s))+\mathbb E_{s\sim\rho}R_{j,\theta_j^*}(s,\pi'(s))=\mathbb E_{s\sim\rho}R_{i,\theta_i^*}(s,\pi(s))+\mathbb E_{s\sim\rho}R_{j,\theta_j^*}(s,\pi(s))$,
        \item $|\mathbb E_{s\sim\rho}R_{i,\theta_i^*}(s,\pi'(s))-\mathbb E_{s\sim\rho}R_{j,\theta_j^*}(s,\pi'(s))|=|\mathbb E_{s\sim\rho}R_{i,\theta_i^*}(s,\pi(s))-\mathbb E_{s\sim\rho}R_{j,\theta_j^*}(s,\pi(s))|$,
        \item $\mathbb E_{s\sim\rho}R_{k,\theta_k^*}(s,\pi'(s))=\mathbb E_{s\sim\rho}R_{k,\theta_k^*}(s,\pi(s))$ for any $k\neq i,j$,
    \end{itemize} 
    it holds that $J(\pi')-J(\pi)\leq \tau$.
    Equivalently, the welfare function increases by at most $\tau$ in a move reducing the gap (difference) between any two individual values $\mathbb E_{s\sim\rho}R_{i,\theta_i^*},\mathbb E_{s\sim\rho}R_{j,\theta_j^*}$ and retaining their total value.
\end{definition}

Using the bound on $\prod_{m=1}^M [\mathbb E_{s\sim\rho} R_{m,{\theta}_m^*}(s,\pi^*(s))]^{1/M}-\prod_{m=1}^M [\mathbb E_{s\sim\rho}R_{m,{\theta}_m}(s,\hat{\pi}(s))]^{1/M}$, it directly follows that:
\begin{corollary}\label{th:pigou}
     Suppose Assumption \ref{as:well},\ref{as:feature} hold. If the number of samples satisfies $n\gg d+\log(M/\delta)$ for $\delta\in (0,1)$, then with probability at least $1-\delta$, $\hat{\pi}$ is $\tau$-approximate Pigou-Dalton principle, where $\tau
     \lesssim C^*{\sqrt{\frac{1}{n}\big({r^2}+\dfrac{dr^2\log{d}}{M}+\log(M/\delta)\big)}}$.
\end{corollary}

\subsection{Proofs in Section \ref{ap:moreswf}}\label{pf:swf}
We begin with the \textbf{Utilitarian} welfare function. The proof itself is straightforward to add up individual errors for each party. Here we mainly demonstrate the most significant difference of the proof.
\begin{proof}
    Again, we consider only the first term in the expression:
    $J(\pi^*)-J(\hat{\pi})=(J(\pi^*)-\hat{J}(\pi^*))+(\hat{J}(\pi^*)-\hat{J}(\hat{\pi}))+(\hat{J}(\hat{\pi})-J(\hat{\pi}))$ as in the proof of Theorem \ref{th:pes+}. Define $\Tilde{\theta}_m$ in a similar manner. We proceed as follows:
    \begin{align*}
        &J(\pi^*)-J(\hat{\pi})\leq J(\pi^*)-\hat{J}(\pi^*)\\
        =&\max_{\Theta\in\mathbf{\Theta}_B(\hat{\Theta})}\left\{\dfrac{1}{M}\sum_{m=1}^M\mathbb E_{s\sim\rho} R_{m,{\theta}_m^*}(s,\pi^*(s))-\dfrac{1}{M}\sum_{m=1}^M\mathbb E_{s\sim\rho} R_{m,{\theta}_m}(s,\pi^*(s)) \right\}\\
        =&\dfrac{1}{M}\sum_{m=1}^M\mathbb E_{s\sim\rho} R_{m,{\theta}_m^*}(s,\pi^*(s))-\dfrac{1}{M}\sum_{m=1}^M\mathbb E_{s\sim\rho} R_{m,\Tilde{\theta}_m}(s,\pi^*(s)) \\
        \leq&\dfrac{1}{M}\sum_{m=1}^M \left|\mathbb E_{s\sim\rho}\left(R_{m,{\theta}_m^*}(s,\pi^*(s))-R_{m,\Tilde{\theta}_m}(s,\pi^*(s))\right)\right|\\
        = &\dfrac{1}{M}\sum_{m=1}^M \left|\mathbb E_{s\sim\rho}({\theta}_m^*-\Tilde{\theta}_m)^{\top}\phi(s,\pi^*(s))\right|.
    \end{align*}
    Thus, the remaining steps follow similarly to the proof of Theorem \ref{th:pes+}.
\end{proof}

Then, we move to the \textbf{Egalitarian} welfare function. We introduce a lemma first. 
\begin{lemma}\label{le:leximin}
    Suppose $a_1,\cdots,a_n$ and $b_1,\cdots,b_n$ are two groups of real values, sorted as $a_{(1)}\leq\cdots \leq a_{(n)}$ and $b_{(1)}\leq\cdots\leq b_{(n)}$, respectively. Then, the following inequality holds:
    \begin{equation*}
        a_{(i)}-b_{(i)}\leq \max_{1\leq k\leq n}(a_k-b_k).
    \end{equation*}
\end{lemma}
\begin{proof}
    Let $a_{(i)}=a_{s_i},b_{(j)}=b_{t_j}$, where $s_i,t_j \in [n]$. For $i=1$, we have
    \begin{equation*}
        a_{(1)}-b_{(1)}=a_{s_1}-b_{t_1}=(a_{s_1}-a_{t_1})+(a_{t_1}-b_{t_1})\leq 0+\max_k(a_k-b_k)=\max_k(a_k-b_k).
    \end{equation*}
    We complete the case for $i=1$. 
    In general, for any $i\in [n]$, we consider two cases:
    
    \textbf{Case 1}: If $a_{s_i}\leq a_{t_i}$, we have 
    \begin{equation*}
        a_{(i)}-b_{(i)}=a_{s_i}-b_{t_i}=(a_{s_i}-a_{t_i})+(a_{t_i}-b_{t_i})\leq 0+\max_k(a_k-b_k)=\max_k(a_k-b_k).
    \end{equation*}
    \textbf{Case 2}: Otherwise $a_{s_i}> a_{t_i}$. We know that there are only $i-1$ values smaller than $a_{s_i}$, i.e., $a_{(1)},\cdots,a_{(i-1)}\leq a_{(i)}=a_{s_i}$. Thus there exists $l\leq i$ such that $a_{t_l}\geq a_{s_i}$. Thus, we have
    \begin{equation*}
        a_{(i)}-b_{(i)}=a_{s_i}-b_{(i)}\leq a_{s_i}-b_{(l)}\leq a_{t_l}-b_{t_l}\leq \max_k (a_k-b_k).
    \end{equation*}
    We conclude the proof.
\end{proof}

With such a lemma, the proof of the Egalitarian case is similar to Theorem \ref{th:pes+}. Here we mainly demonstrate the most significant difference of the proof.
\begin{proof}
    Again, we consider only the first term in the expression:
    $J(\pi^*)-J(\hat{\pi})=(J(\pi^*)-\hat{J}(\pi^*))+(\hat{J}(\pi^*)-\hat{J}(\hat{\pi}))+(\hat{J}(\hat{\pi})-J(\hat{\pi}))$ as in the proof of Theorem \ref{th:pes+}. Define $\Tilde{\theta}_m$ in a similar manner. We proceed as follows:
    \begin{align*}
        &J(\pi^*)-J(\hat{\pi})\leq J(\pi^*)-\hat{J}(\pi^*)\\
        =&\max_{\Theta\in\mathbf{\Theta}_B(\hat{\Theta})}\left\{\min_m \mathbb E_{s\sim\rho}R_{m,\theta_m^*}(s,{\pi}(s))-\min_m \mathbb E_{s\sim\rho}R_{m,\theta_m}(s,{\pi}(s)) \right\}\\
        =&\min_m \mathbb E_{s\sim\rho}R_{m,\theta_m^*}(s,{\pi}(s))-\min_m \mathbb E_{s\sim\rho}R_{m,\Tilde{\theta}_m}(s,{\pi}(s))\\
        \leq&\max_m  \left( \mathbb E_{s\sim\rho}R_{m,\theta_m^*}(s,{\pi}(s))- \mathbb E_{s\sim\rho}R_{m,\Tilde{\theta}_m}(s,{\pi}(s))\right) \quad (\text{using Lemma \ref{le:leximin}})\\
        \leq&\max_m \left|\mathbb E_{s\sim\rho}({\theta}_m^*-\Tilde{\theta}_m)^{\top}\phi(s,\pi^*(s))\right|.
    \end{align*}
    Thus, the remaining steps follow similarly to the proof of Theorem \ref{th:pes+}.
\end{proof}

\subsection{Proof of Theorem \ref{th:mdp-pes}}\label{pf:mdp}
We begin with an estimation error bound similar to Theorem \ref{th:metatheta}.
\begin{theorem}\label{th:mdp}
    Suppose Assumption \ref{as:well},\ref{as:feature} hold, with the feature difference in $\Sigma_*$ of Assumption \ref{as:feature} changed to cumulative difference over the trajectory. If the number of samples satisfies $n\gg d+\log(M/\delta)$ for $\delta\in (0,1)$, then with probability at least $1-\delta$, for any $m\in [M]$,
    \begin{equation*}
        \|\theta_m^*-\hat{\theta}_m\|_{\Sigma_m}\lesssim \sqrt{\dfrac{r^2}{n}+\dfrac{dr^2\log{d}+r\log(M/\delta)}{Mn}}.
    \end{equation*}
\end{theorem}
\begin{proof}
    This proof is the same as Theorem \ref{th:metatheta}. In the MDP setting, we just consider $X_{m,i}=\sum_{h=1}^H \left(\phi(s_{m,h}^i,a_{m,h}^i)-\phi(s_{m,h}^{i'},a_{m,h}^{i'})\right)$ with $\|X_{m,i}\|_2\leq 2BLH$ instead.
\end{proof}

Then, we give the proof of Theorem \ref{th:mdp-pes}.
\begin{proof}
    Decompose the sub-optimality into three terms:
    \begin{equation*}
        J(\pi^*)-J(\hat{\pi})=\left(J(\pi^*)-\hat{J}(\pi^*)\right)+\left(\hat{J}(\pi^*)-\hat{J}(\hat{\pi})\right)+\left(\hat{J}(\hat{\pi})-J(\hat{\pi})\right).
    \end{equation*}
    We can again focus on the first term. Let $(\Tilde{\theta}_1,\cdots,\Tilde{\theta}_M)= \arg\min_{\Theta\in\mathbf{\Theta}_B(\hat{\Theta})}\prod_{m=1}^M[\mathbb E_{s\sim\rho} R_{m,{\theta}_m}(s,\pi^*(s))]^{1/M}$, which represents the pessimistic parameter chosen from lower confidence bound for the optimal policy $\pi^*$.

     From the assumption, we have $\mathbb E_{s\sim d^{\pi^*}} R_{m,{\theta}_m^*}(s,\pi^*(s))\geq r_0$. Besides, with probability at least $1-\delta$, we have $\Tilde{\theta}_m,\theta_m^*\in\mathbf{\Theta}_B(\hat{\theta}_m)$ for $\forall m$. Using this, we can compare the reward functions evaluated at $\Tilde{\theta}_m$ and $\theta_m^*$:
    \begin{align*}
        |R_{m,\Tilde{\theta}_m}(s,\pi^*(s))-R_{m,{\theta}_m^*}(s,\pi^*(s))|
        &=|({\theta}_m^*-\Tilde{\theta}_m)^\top \phi(s,\pi^*(s))|\\
        &\leq \|{\theta}_m^*-\Tilde{\theta}_m\|_2 \|\phi(s,\pi^*(s))\|_2\\
        &\lesssim  1/H\|{\theta}_m^*-\Tilde{\theta}_m\|_{\Sigma_m} \|\phi(s,\pi^*(s))\|_2\\
        &\lesssim 1/H\sqrt{\dfrac{r^2}{n}+\dfrac{dr^2\log{d}+r\log(M/\delta)}{Mn}}
    \end{align*}
    The last inequality holds based on the fact that $\Tilde{\theta}_m\in\mathbf{\Theta}_B(\hat{\theta}_m)$ and $\theta_m^*\in\mathbf{\Theta}_B(\hat{\theta}_m)$ with probability at least $1-\delta$ for $\forall m$, as well as the boundedness of $\phi$.
    Thus, for sufficiently large $n$, we also have that
    $\mathbb E_{s\sim d^{\pi^*}} R_{m,\Tilde{\theta}_m}(s,\pi^*(s))\geq r_0/2$.
     Therefore, with probability at least $1-\delta$, we can conclude that
    \begin{align*}
        &J(\pi^*)-J(\hat{\pi})\leq J(\pi^*)-\hat{J}(\pi^*)\\
        =&\max_{\Theta\in\mathbf{\Theta}_B(\hat{\Theta})}\left(\prod_{m=1}^M [\mathbb E_{s\sim d^{\pi^*}} R_{m,{\theta}_m^*}(s,\pi^*(s))]^{1/M}-\prod_{m=1}^M [\mathbb E_{s\sim d^{\pi^*}}R_{m,{\theta}_m}(s,\pi^*(s))]^{1/M}\right) \\
        =&\prod_{m=1}^M [\mathbb E_{s\sim d^{\pi^*}} R_{m,{\theta}_m^*}(s,\pi^*(s))]^{1/M}-\prod_{m=1}^M [\mathbb E_{s\sim d^{\pi^*}}R_{m,\Tilde{\theta}_m}(s,\pi^*(s))]^{1/M}\\
        \lesssim& \dfrac{1}{M}\sum_{m=1}^M |\mathbb E_{s\sim d^{\pi^*}} R_{m,{\theta}_m^*}(s,\pi^*(s))-\mathbb E_{s\sim d^{\pi^*}}R_{m,\Tilde{\theta}_m}(s,\pi^*(s))| \qquad \text{(using Lemma \ref{lm:geo})}\\
        =&\dfrac{1}{M} \sum_{m=1}^M \left|({\theta}_m^*-\Tilde{\theta}_m)^{\top}\mathbb E_{s\sim d^{\pi^*}}\phi(s,\pi^*(s))\right|
    \end{align*}

    Separately, for each $m$ we have
    \begin{align*}
        &\left|({\theta}_m^*-\Tilde{\theta}_m)^{\top}\mathbb E_{s\sim d^{\pi^*}}\phi(s,\pi^*(s))\right|\\
        =&\left|({\theta}_m^*-\hat{\theta}_m+\hat{\theta}_m-\Tilde{\theta}_m)^{\top}\mathbb E_{s\sim d^{\pi^*}}\phi(s,\pi^*(s))\right|\\
        \leq&\left|({\theta}_m^*-\hat{\theta}_m)^{\top}\mathbb E_{s\sim d^{\pi^*}}\phi(s,\pi^*(s))\right|+\left|(\hat{\theta}_m-\Tilde{\theta}_m)^{\top}\mathbb E_{s\sim d^{\pi^*}}\phi(s,\pi^*(s))\right|\\
        \leq&\|{\theta}_m^*-\hat{\theta}_m\|_{\Sigma_m}\|\mathbb E_{s\sim d^{\pi^*}}\phi(s,\pi^*(s))\|_{\Sigma_m^{-1}}+\|\hat{\theta}_m-\Tilde{\theta}_m\|_{\Sigma_m}\|\mathbb E_{s\sim d^{\pi^*}}\phi(s,\pi^*(s))\|_{\Sigma_m^{-1}} \quad \text{(Cauchy-Schwartz inequality)}\\
        \lesssim&  \sqrt{\dfrac{r^2}{n}+\dfrac{dr^2\log{d}+r\log(M/\delta)}{Mn}}\|(\Sigma_m^{-1/2}\mathbb E_{s\sim d^{\pi^*}}\phi(s,\pi^*(s)))\|_2
    \end{align*}
    The last inequality holds because $\Tilde{\theta}_m\in\mathbf{\Theta}_B(\hat{\theta}_m)$ and $\theta_m^*\in\mathbf{\Theta}_B(\hat{\theta}_m)$ with probability at least $1-\delta$ for $\forall m$. 
   Therefore, we conclude that
   \begin{equation*}
       J(\pi^*)-J(\hat{\pi}) \lesssim \sqrt{\dfrac{r^2}{n}+\dfrac{dr^2\log{d}+r\log(M/\delta)}{Mn}}C^*.
   \end{equation*}
   where $C^*=\max_m\mathbb E_{s\sim d^{\pi^*}}\|(\Sigma_m^{-1/2}\phi(s,\pi^*(s)))\|_2$.
\end{proof}

\section{Technical Details for Reward-Free Model}\label{ap:von}
\subsection{Proof of Theorem \ref{th:von-pes}}\label{pf:von-pes}

We begin by introducing two fundamental concentration inequalities that will be useful in our analysis.
\begin{proposition}[McDiarmid inequality, \cite{vershynin2018high} ]\label{le:mc}
    Let $X_1,\cdots,X_n$ be independent random variables in $\mathcal{X}$, let $g:\mathcal{X}\rightarrow\mathbb{R}$ be a function of $X_1,\cdots,X_n$ s.t. $\forall x_1,\cdots,x_n,x_i'\in \mathcal{X}$, $|h(x_1,\cdots,x_i,\cdots,x_n)-h(x_1,\cdots,x_i',\cdots,x_n)|\leq c$.
    Then for all $\delta>0$, with probability at least $1-\delta$, we have
    $$|h(X_1,\cdots,X_n)-E(h(X_1,\cdots,X_n))|\leq c \sqrt{\dfrac{n\log{(2/\delta)}}{2}}.$$
\end{proposition}
\begin{proposition}[Binomial concentration inequality, \cite{xie2021policy}]\label{le:bino}
    Suppose $N \sim \texttt{Ber}(n, p)$, which is a Bernoulli distribution with parameter $n\geq 1$ and $p\in[0, 1]$. Then with probability at least $1-\delta$, we have
    \begin{equation*}
        \dfrac{p}{N\lor 1}\leq \dfrac{8\log(1/\delta)}{n}.
    \end{equation*}
\end{proposition}

\begin{lemma}\label{le:mc-von}
    With probability at least $1-\delta$, 
    
    (1) $|N_{s,m}(a,a')-N_{s,m}^*(a,a')|\leq  \sqrt{\dfrac{2\log{(2MSA^2/\delta)}}{n_0(m,a,a',s)\lor 1}}$,  for $\forall m,s,a,a'$ ;
    
    (2) $|N_{s}(a,a')-N_{s}^*(a,a')|\leq  \sqrt{\dfrac{2\log{(2MSA^2/\delta)}}{\min_m n_0(m,a,a',s)\lor 1}}$, for $\forall s,a,a'$.
\end{lemma}
\begin{proof} 
    (1) For fixed $m,s,a,a'$, define $n'=n_0(m,a,a',s)$.
    The cases where $n'=0$ is trivial. For $n'\geq 1$, let $X_{m,i}$ for $i\in [n']$ represent all the binary responses corresponding to the pair $(s,a,a')$ in $\mathcal{D}_m$.
    Let $h(X)=\sum_{i=1}^{n'} X_i/{n'}$. Since $|X_i-X_i'|/{n'} \leq 1/{n'}$. we apply Proposition \ref{le:mc} and substitute $c$ with $1/n$, then with probability at least $1-\delta$, we have
    \begin{align*}
        &\left|\dfrac{\#_m(a\succ a';s)}{\#_m(a\succ a';s)+\#_m(a'\succ a;s)}-\dfrac{N_{s,m}^*(a,a')+1}{2}\right|\\
        =&\left|\dfrac{\#_m(a\succ a';s)}{n'}-\dfrac{N_{s,m}^*(a,a')+1}{2}\right|\\
        =&\left|\dfrac{\#_m(a\succ a';s)}{n'}-\mathbb P_m(y=1|s,a,a')\right|\\
        \leq& \dfrac{1}{n'}\sqrt{\dfrac{n'\log{(2/\delta)}}{2}}
    \end{align*}
    
    So for fixed $m,s,a,a'$, with probability at least $1-\delta$,
    \begin{equation*}
        |N_{s,m}(a,a')-N_{s,m}^*(a,a')|\leq 2\left|\dfrac{N_{s,m}(a,a')+1}{2}-\dfrac{N_{s,m}^*(a,a')+1}{2}\right|\leq  \sqrt{\dfrac{2\log{(2/\delta)}}{n_0(m,a,a',s)}}.
    \end{equation*}
    
    Then with probability at least $1-MSA^2\delta$, 
    $$|N_{s,m}(a,a')-N_{s,m}^*(a,a')|\leq  \sqrt{\dfrac{2\log{(2/\delta)}}{n_0(m,a,a',s)}}\quad \forall s,m,a,a'.$$ 
    
    Substitute $\delta$ with $\delta/(MSA^2)$, we complete the first part. 

    (2) With probability at least $1-\delta$, for $\forall a,a',s$,
    \begin{align*}
        |N_{s}(a,a')-N_{s}^*(a,a')|&=|\dfrac{1}{M}\sum_{m=1}^M N_{s,m}(a,a')-\dfrac{1}{M}\sum_{m=1}^M N_{s,m}^*(a,a')|\\
        &\leq \dfrac{1}{M}\sum_{m=1}^M |N_{s,m}(a,a')-N_{s,m}^*(a,a')|\\
        &\leq  \sqrt{\dfrac{2\log{(2MSA^2/\delta)}}{\min_m n_0(m,a,a',s)}}.
    \end{align*}
    Thus we conclude the proof.
\end{proof}

Now we present the detailed proof of Theorem \ref{th:von-pes}.
\begin{proof}
    First, from Lemma \ref{le:mc-von}, with probability at least $1-\delta/2$,
    \begin{equation*}
        |N_s(a,a')-N_s^*(a,a')|\leq B_s(a,a')=\sqrt{\dfrac{2\log (4MSA^2/\delta) }{\min_m n_0(m,a,a',s)\lor 1}}, \quad \forall s,a,a'.
    \end{equation*}
    We note that $n_{0}(m,a,a^\prime,s)\sim B(n,\rho(s)d_{s}^{g}(a,a^\prime))$, then from from Proposition \ref{le:bino}, with probability at least $1-\delta/2$,
    \begin{equation}\label{ber_dg}
        \dfrac{\rho(s)d_s^g(a,a')}{n_0(m,a,a',s)\lor 1}\leq \dfrac{8\log(2MSA^2/\delta)}{n},\quad\forall m,s,a,a'.
    \end{equation}
    Thus the above inequalities hold simultaneously with probability at least $1-\delta$.
    Next, recall that $\hat{p}_s,\hat{q}_s=\arg\max_{p}\min_{q}p^{\top}(N_s-B_s)q$ and $p_s^*,q_s^*=\arg\max_{p}\min_{q}p^{\top}N_s^*q$, we have:
    \begin{align*}
        \min_q\hat{p}_s^\top N_s^*q &\geq \min_q\hat{p}_s^\top (N_s-B_s)q & (\text{using }N_s^*\geq N_s-B_s)\\
        & =\hat{p}_s^\top (N_s-B_s)\hat{q}_s\\
        & \geq p_s^{*\top} (N_s-B_s)\hat{q}_s &  (\text{using the definition of }\hat{p}_s)\\
        & =p_s^{*\top} N_s^*\hat{q}_s+p_s^{*\top} (N_s-N_s^*)\hat{q}_s-p_s^{*\top} B_s\hat{q}_s\\
        & \geq \min_q p_s^{*\top} N_s^* q -2p_s^{*\top} B_s\hat{q}_s & (\text{using }N_s^*\geq N_s-B_s)\\
        &=-2p_s^{*\top} B_s\hat{q}_s & (\text{the value of the matrix game }N_s^*=0)\\
        &=-2\mathbb E_{\substack{a\sim p_s^*\\ a'\sim \hat{q}_s}} B_s(a,a')\\
        &=-2 \mathbb E_{\substack{a\sim p_s^*\\ a'\sim \hat{q}_s}}\sqrt{\dfrac{2\log (4MSA^2/\delta) }{\min_m n_0(m,a,a',s)\lor 1}}\\
        &\geq -2\mathbb E_{\substack{a\sim p_s^*\\ a'\sim \hat{q}_s}} 4\sqrt{\dfrac{\log ^2(4MSA^2/\delta) }{n \rho(s) d_s^g(a,a')}} & (\text{using \eqref{ber_dg}}).
    \end{align*}
    To simplify the notation, let $\eta=\log (4MSA^2/\delta), $ and denote $d_s^{\pi}=d_s^{p_s^*,\hat{q}_s}$. We have
    \begin{align*}
        \min_q\hat{p}_s^\top N_s^*q&\geq -8\sum_{a,a'}\eta d_s^{\pi}(a,a')\sqrt{\dfrac{1}{n\rho(s)d_s^g(a,a')}}\\
        &=-8\sum_{a,a'} \sqrt{d_s^{\pi}(a,a')}\eta\sqrt{\dfrac{d_s^{\pi}(a,a')}{n \rho(s) d_s^g(a,a')}}\\
        &\geq -8\sqrt{\sum_{a,a'} d_s^{\pi}(a,a') A^2}\eta\sqrt{\dfrac{d_s^{\pi}(a,a')}{n \rho(s) d_s^g(a,a')}}& (\text{Cauchy-Schwartz inequality})\\
        &\geq -8A\eta\sqrt{\dfrac{d_s^{\pi}(a,a')}{n \rho(s) d_s^g(a,a')}}\\
        &\geq -8A\eta\sqrt{\dfrac{C^*}{n}}:=-\epsilon.
    \end{align*}

    Thus we conclude that $\hat{p}_s$ satisfies $\min_q\hat{p}_s^\top N_s^*q \geq-\epsilon$, which implies the following equivalent conditions:
    \begin{align}\label{eq:act}
        &\sum_{a\in\mathcal{A}}\hat{p}_s(a)N_s^*(a,a')\geq -\epsilon, \quad \forall a',s,\\
        &\sum_{a\in\mathcal{A}}\hat{p}_s(a)=1, \quad  \forall s.\nonumber
    \end{align}
    Our objective is to establish that the probability distribution $\hat{p}$ over policy space $\Pi$ satisfies the following conditions:
    \begin{align*}
        &\sum_{\pi\in \Pi}\hat{p}(\pi) T^*(\pi,\pi')\geq -\epsilon, \quad \forall \pi',\\
        &\sum_{\pi\in \Pi}\hat{p}(\pi) =1.
    \end{align*}
    Since $\hat{p}(\pi)=\prod_{s\in\mathcal{S}}\hat{p}_s(\pi(s))$, it follows straightforwardly that: $\sum_{\pi\in \Pi}\hat{p}(\pi) =1$.
    Besides, notice that
    \begin{align}\label{eq:rhos}
        \sum_{\pi\in \Pi}\hat{p}(\pi) T^*(\pi,\pi')&=\sum_{\pi\in \Pi}\hat{p}(\pi)\sum_{s\in\mathcal{S}}\rho_sN_s^*(\pi(s),\pi'(s))\nonumber\\
        &=\sum_{s\in\mathcal{S}}\rho_s\sum_{\pi\in \Pi}\hat{p}(\pi)N_s^*(\pi(s),\pi'(s)).
    \end{align}
    
    We now separate the terms for different states. For simplicity, let $\mathcal{S}=\{s_i\}_{i=1}^S$ and $\hat{p}_{s_i}=\hat{p}_i$ for $i\in [S]$. Thus, for each state $s_i$, we have
    \begin{align*}
        &\sum_{\pi\in \Pi}\hat{p}(\pi)N_{s_i}^*(\pi(s_i),\pi'(s_i))\\
        =&\sum_{\pi\in \Pi}\prod_{i=1}^S\hat{p}_i(\pi(s_i))N_{s_i}^*(\pi(s_i),\pi'(s_i))\\
        =&\sum_{\pi\in \Pi} [\prod_{j\neq i}\hat{p}_j(\pi(s_j))] \hat{p}_i(\pi(s_i))N_{s_i}^*(\pi(s_i),\pi'(s_i))\\
        =&\left(\sum_{
	\substack{\pi(s_j)\in\mathcal{A}\\ \forall j\neq i}}
        \prod_{j\neq i}\hat{p}_j(\pi(s_j))\right) \left(\sum_{\pi(s_i)\in \mathcal{A}} \hat{p}_i(\pi(s_i))N_{s_i}^*(\pi(s_i),\pi'(s_i))\right)\geq -\epsilon.
    \end{align*}
    
    The last inequality holds because the first term equals $1$, and the second term is greater than or equal to $-\epsilon$ for all $a'=\pi'(s_i)$ using \eqref{eq:act}. Therefore, combining with \eqref{eq:rhos} implies that
    \begin{equation*}
        \sum_{\pi\in \Pi}\hat{p}(\pi) T^*(\pi,\pi')\geq -\sum_s \rho_s\epsilon=-\epsilon.
    \end{equation*}
    Now we conclude the proof.
\end{proof}

\subsection{Proof of Theorem \ref{th:expost}}\label{pf:expost}
This proof follows a result similar to the one presented in \cite{fishburn1984probabilistic}.
\begin{proof}
    Suppose $N_{s,m}^*(a,b)\geq0$ for $\forall m$ and the inequality holds for some specific $m$, then from transitivity we have $N_{s,m}^*(a,c)\geq N_{s,m}^*(b,c)$ for $\forall c$. Therefore, $N_{s}^*(a,b)>0$ and $N_{s}^*(a,c)\geq N_{s}^*(b,c)$. 

    Now, define $\hat{p}_s'$ as the probability distribution $\hat{p}_s$ with the following modifications: 
    $$\hat{p}_s'(a) = \hat{p}_s(a) + \hat{p}_s(b),\quad \hat{p}_s'(b) = 0,\quad \hat{p}_s'(c)=\hat{p}_s(c), c\notin \{a,b\}.$$
    It then follows that
    \begin{align}\label{eq:expostbound}
        \hat{p}_s'{}^{\top} N_{s}^*\hat{p}_s=&N_{s}^*(a,b)\left[(\hat{p}_s(a)+\hat{p}_s(b))\hat{p}_s(b)-0\cdot \hat{p}_s(a)\right]\nonumber\\
        +&\sum_{c\in \mathcal{S}\backslash\{a,b\}}N_{s}^*(a,c)\left[(\hat{p}_s(a)+\hat{p}_s(b))\hat{p}_s(c)-\hat{p}_s(c)\hat{p}_s(a)\right] \nonumber\\
        +& \sum_{c\in \mathcal{S}\backslash\{a,b\}}N_{s}^*(b,c)\left[0\cdot \hat{p}_s(c)-\hat{p}_s(c)\hat{p}_s(b)\right]\nonumber\\
        =&\hat{p}_s(b)\left[N_{s}^*(a,b)(\hat{p}_s(a)+\hat{p}_s(b))+\sum_{c\in \mathcal{S}\backslash\{a,b\}}\hat{p}_s(c)(N_{s}^*(a,c)-N_{s}^*(b,c))\right]\nonumber\\
        \geq& \hat{p}_s(b)^2 N_{s}^*(a,b).
    \end{align}
    Notice that with probability at least $1-\delta$, we have
    \begin{align*}
        &\hat{p}_s'{}^{\top} N_{s}^*\hat{p}_s=-\hat{p}_s^{\top} N_{s}^*\hat{p}_s'\leq\epsilon. & (\text{using } \hat{p}_s^{\top}N_{s}^*\hat{p}_s'\geq\min_{q}\hat{p}_s^{\top}N_{s}^*q\geq-\epsilon)
    \end{align*}
    where $\epsilon$ is defined in Theorem \ref{th:von-pes}. Therefore, \eqref{eq:expostbound} implies that $\epsilon\geq\hat{p}_s(b)^2 N_{s}^*(a,b)$, which leads to the following inequality for $\hat{p}_s(b)$:
    \begin{align*}
        \hat{p}_s(b)\leq \sqrt{\epsilon/N_{s}^*(a,b)}&=\left(8A\log (2MSA^2/\delta)\sqrt{\dfrac{C^*}{n}}/N_{s}^*(a,b)\right)^{\frac{1}{2}}\\
        &=K\left(A\log(2MSA^2/\delta)\sqrt{\frac{C^*}{n}}\right)^{1/2},
    \end{align*}
    where $K$ is a constant that depends on $N_s^*(a,b)$.
\end{proof}

We now proceed to demonstrate the \textbf{necessity of transitivity}. Consider a simple example with three alternatives: Suppose that $N_{s,m}^*(a,b)=1,N_{s,m}^*(b,c)=1,N_{s,m}^*(c,a)=1$ for $\forall m$. This means that for each individual, given comparison of $(a,b)$, $a\succ b$ with probability $1$. The same holds for the other pairs. Therefore, all individuals have deterministic preferences with respect to the three alternatives, and the preference matrix takes the following form:
\begin{equation*}
    N_s=N_s^*=
    \begin{bmatrix}
      0 & 1 & -1 \\
      -1 & 0 & 1 \\
      1 & -1 & 0 \\
    \end{bmatrix}.
\end{equation*}

Since everyone has absolute preferences, there is no need to introduce a confidence bound for the rankings. As a result, the von Neumann winner directly yields the outcome $(1/3,1/3,1/3)$. However, as item $b$ is Pareto dominated by item $a$, the result stated in the previous theorem does not hold anymore, where transitivity is assumed. Without transitivity, it may not have consistent preferences. The transitivity is indeed a necessary condition for the result to hold.

\section{Experimental Details}\label{app:exp}

This appendix provides additional details for the experiments in
Section~\ref{sec:exp}.

\subsection{Experimental Details for Section~\ref{sec:exp1}}
\label{app:exp1}

\subsubsection{Dataset Construction.}
We construct a finite contextual-bandit instance designed to illustrate two
features of the proposed method. First, the parties have heterogeneous
preferences but share a low-dimensional reward representation. Second, the
comparison features span a substantially larger ambient space, so the learner
must identify the reward-relevant subspace from the observations. Specifically,
we consider \(3\) states, \(9\) actions, and \(M=24\) parties, with ambient
dimension \(d=24\) and latent dimension \(r=3\). The state distribution
\(\rho\) is uniform over
\(\mathcal S=\{s_1,s_2,s_3\}\), and
\(\mathcal A=\{a_1,\ldots,a_9\}\). We write
\([q]=\{1,\ldots,q\}\).

The true parameter matrix satisfies
\[
\Theta^*
=
(\theta_1^*,\ldots,\theta_M^*)
=
U^*\alpha^*,
\qquad
\alpha^*
=
(\alpha_1^*,\ldots,\alpha_M^*)
\in\mathbb R^{3\times24},
\]
where
\[
U^*
=
\begin{pmatrix}
I_3\\
0_{21\times3}
\end{pmatrix}
\in\mathcal O_{24\times3}.
\]
Thus, the true party parameters lie in the three-dimensional subspace spanned
by the first three ambient coordinates. The representation \(U^*\) is used
only to generate the data and is unavailable to the learner.

We divide the parties into three equally sized clusters. For \(j\in[3]\), let
\[
G_j=\{8(j-1)+1,\ldots,8j\}.
\]Each cluster is associated with one of the three latent reward directions. For every \(j\in[3]\) and \(k\in[4]\), we independently draw
\[
\zeta_{j,k}\sim\operatorname{Unif}(0,0.05).
\]We assign to the eight parties in \(G_j\) the antithetic perturbations
\[
\{\epsilon_m:m\in G_j\}
=
\{\zeta_{j,1},\ldots,\zeta_{j,4},
-\zeta_{j,1},\ldots,-\zeta_{j,4}\},
\]and define
\[
\alpha_m^*=(1+\epsilon_m)e_j,
\qquad
\theta_m^*=U^*\alpha_m^*,
\qquad m\in G_j.
\]The continuous perturbations make the parties within each cluster distinct almost surely, while their antithetic structure ensures
\[
\sum_{m\in G_j}\epsilon_m=0,
\qquad
\frac{1}{|G_j|}\sum_{m\in G_j}\alpha_m^*=e_j.
\]Thus, the construction introduces party-level heterogeneity without shifting the three cluster centers. Because these centers span the three latent directions, the resulting party-parameter matrix has rank \(3\). The parameters are generated once and held fixed across all repetitions, so only the sampled comparison labels vary between repetitions.

We next construct the feature vectors. Write $\phi(s,a)=\bigl(\phi_{1:3}(s,a),\phi_{4:24}(s,a)\bigr),$ where the first three coordinates determine the true rewards. They are given
by
\[
\begin{array}{c|c|c}
\text{action}
&
\phi_{1:3}(s_1,a)
&
\phi_{1:3}(s_2,a)=\phi_{1:3}(s_3,a)
\\ \hline
a_1
&(7/2,-1/4,-3/4)^\top
&(13/4,-1/2,-1/2)^\top
\\
a_2
&(-1/2,3,-1/2)^\top
&(-1/2,3,-1/2)^\top
\\
a_3
&(-1/2,-1/2,3)^\top
&(-1/2,-1/2,3)^\top
\\
a_4
&(3/4,3/4,1/2)^\top
&(3/4,3/4,1/2)^\top
\\
a_5,\ldots,a_8
&(0,0,0)^\top
&(0,0,0)^\top
\\
a_9
&(-1,-1,-1)^\top
&(-1,-1,-1)^\top.
\end{array}
\]
This construction gives the actions distinct roles. Actions \(a_1,a_2\), and
\(a_3\) are specialist actions favored by clusters \(G_1,G_2\), and \(G_3\),
respectively, whereas \(a_4\) provides a compromise that assigns positive
reward to all three clusters. Actions \(a_5,\ldots,a_8\) carry no true reward
signal, and \(a_9\) serves as the common comparison reference. The construction
therefore creates a transparent collective-choice problem in which policies
must trade off the interests of the three party clusters.

We use the remaining \(21\) coordinates to provide ambient feature coverage.
Let \(e_t^{(24)}\) denote the \(t\)-th standard basis vector in
\(\mathbb R^{24}\), and write
\[
\bar\phi_{\ell,h}=\phi_{1:3}(s_\ell,a_h).
\]
Define
\[
\mathcal N
=
\{(1,4)\}
\cup
\bigl(\{2,3\}\times[4]\bigr)
\cup
\bigl([3]\times\{5,6,7,8\}\bigr),
\]
and, for \((\ell,h)\in\mathcal N\), let
\[
c(\ell,h)=
\begin{cases}
4,
&(\ell,h)=(1,4),\\
4+4(\ell-2)+h,
&\ell\in\{2,3\},\ h\in[4],\\
12+4(\ell-1)+(h-4),
&\ell\in[3],\ h\in\{5,6,7,8\}.
\end{cases}
\]
Set
\[
\psi_{\ell,h}
=
\begin{cases}
3e_{c(\ell,h)}^{(24)},
&(\ell,h)\in\mathcal N,\\
0_{24},
&\text{otherwise},
\end{cases}
\]
and define the complete feature vector as
\[
\phi(s_\ell,a_h)
=
\begin{pmatrix}
\bar\phi_{\ell,h}\\
0_{21}
\end{pmatrix}
+\psi_{\ell,h}.
\]
For example, \((s_1,a_4)\) receives nuisance coordinate \(4\);
\((s_2,a_1),\ldots,(s_2,a_4)\) receive coordinates \(5,\ldots,8\);
\((s_3,a_1),\ldots,(s_3,a_4)\) receive coordinates \(9,\ldots,12\); and
the state-action pairs involving \(a_5,\ldots,a_8\) receive coordinates
\(13,\ldots,24\). All other state-action pairs have zero nuisance coordinates.

Because every \(\theta_m^*\) is supported on the first three coordinates,
these nuisance features do not affect any party's true reward. Their role is
statistical: they provide \(21\) mutually orthogonal comparison directions and
make the ambient comparison covariance full rank. An unrestricted estimator
must therefore operate in all \(24\) dimensions, whereas the joint estimator
can reduce estimation error by learning that the party rewards share a
three-dimensional representation.

This construction satisfies the assumptions used in our analysis. In particular, the party-parameter matrix has rank \(3\), the comparison covariance is positive definite, and the three party clusters provide nondegenerate coverage of all latent directions. Moreover, we take the latent preference parameter norm bound $B = 1.06$ and normalization constant $C_{0} = 8.5$, the parameter and feature bounds satisfy
\[
C_0-B\max_{s,a}\|\phi(s,a)\|_2>0,
\]
so all values entering the Nash welfare objective and its pessimistic counterpart remain positive. The concentratability coefficient is also finite because the comparison distribution covers the feature directions used by the optimal policy.

Building on the above construction, we now describe data collection process. For every state, each action \(a_h\), \(h\in[8]\), is compared against the
reference action \(a_9\). This produces the \(24\) comparison cells
\[
\{(s_\ell,a_h,a_9):\ell\in[3],\ h\in[8]\},
\]
and each party's comparisons are allocated as evenly as possible across these
cells. Let $ X_{\ell,h}=\phi(s_\ell,a_h)-\phi(s_\ell,a_9).$
The preference label is generated according to the BTL model in
Section~\ref{sec:rwd}, with winning probability $\Phi\!\left(
\langle\theta_m^*,X_{\ell,h}\rangle
\right).$

The horizontal-axis quantity \(n\) in Figure~\ref{fig:nash_simulation} denotes the minimum per-party sample size.
Within each cluster \(G_j\), random half of the parties 
 provide \(n\) comparisons each, while other parties provide \(2n\) comparisons each. The total number of comparisons across all cells is $36n$.
We use $n=150,300,\ldots,3000$ and conduct \(100\) independent repetitions at every sample size.

\subsubsection{Learning methods and baselines.}
Our method applies the joint low-rank MLE followed by the pessimistic Nash
welfare optimization in Algorithm~\ref{alg1}. It jointly estimates the shared
representation \(\widehat U\) and the party-specific coefficients
\(\{\widehat\alpha_m\}_{m=1}^M\), subject to
\[
U^\top U=I_r,
\qquad
\|U\alpha_m\|_2\leq B,
\qquad B=1.06.
\]
The numerical procedure alternates between updating the shared representation and updating the party-specific coefficients, using three initializations.

We compare our method with three plug-in baselines. The independent baseline $\hat{\pi}_{\text{ind}}$ estimates one unrestricted \(d\)-dimensional parameter for each party and therefore isolates the benefit of learning a shared low-dimensional
representation. The remaining two baselines discard party identities, pool the observations, and fit
a single reward parameter, but differ in how they weight the observations.
Let \(\ell_m(\theta)\) denote the unnormalized BTL negative log-likelihood for
party \(m\). The \textnormal{pooling-obs} baseline $\hat{\pi}_{\text{pool-obs}}$ solves
\[
\min_{\|\theta\|_2\leq B}
\sum_{m=1}^M\ell_m(\theta),
\]
so each comparison observation receives equal weight and parties with more
comparisons contribute more strongly. In contrast, the
\textnormal{pooling-party} baseline $\hat{\pi}_{\text{pool-party}}$ solves
\[
\min_{\|\theta\|_2\leq B}
\frac{1}{M}\sum_{m=1}^M\frac{\ell_m(\theta)}{n_m},
\]
so every party receives equal total weight regardless of its sample size.
Comparing these two pooling baselines separates the effect of unequal
observation counts from the more fundamental loss of information caused by
pooling heterogeneous party preferences.

All policies are deterministic. We enumerate the full policy class
$\Pi=\mathcal A^{\mathcal S}$ with $|\Pi|=9^3=729,$
and define $\mu_\pi=\mathbb E_{s\sim\rho}[\phi(s,\pi(s))].$
For party \(m\), let
\[
\Sigma_m
=
\frac{1}{n_m}
\sum_{i=1}^{n_m}X_{mi}X_{mi}^\top.
\]
The confidence set used by our method is
\[
\mathbf\Theta_B(\widehat\theta_m)
=
\left\{
\theta:
\|\theta\|_2\leq B,\ 
\|\theta-\widehat\theta_m\|_{\Sigma_m}\leq\Gamma
\right\},
\]
where
\[\Gamma=K\sqrt{\frac{r^2}{n}+\frac{dr^2\log d+r\log(M/\delta)}{Mn}},\]
and we set $K=2.6, \delta=0.05.$ We shall validate empirically in \ref{app:exp1cov} that this choice of $K$ provides consistently high marginal and simultaneous empirical coverage across the reported sample-size range. 
For every candidate policy, the implementation computes the pessimistic value
of each party by solving
\[
\min_{\theta\in\mathbf\Theta_B(\widehat\theta_m)}
C_0+\mu_\pi^\top\theta.
\]
It then selects the policy that maximizes the geometric mean of these
party-specific worst-case values. The optimization is performed over the exact intersection of the confidence ellipsoid and the parameter ball. The
true parameters are used only to evaluate the resulting policies and are
never provided to the learning methods.

\subsubsection{Evaluation Metric.}
Let \(\widehat{\pi}\), \(\widehat{\pi}_{\mathrm{ind}}\), \(\widehat{\pi}_{\mathrm{pool\text{-}obs}}\), and \(\widehat{\pi}_{\mathrm{pool\text{-}party}}\) denote the policies produced by our method, the independent baseline, and the two pooling baselines, respectively.
For a deterministic policy \(\pi\), define party \(m\)’s true value as
\[
V_m(\pi)
=
\mathbb E_{s\sim\rho}
[R_{m,\theta_m^*}(s,\pi(s))].
\]We evaluate the Nash, Utilitarian, and Egalitarian welfare functions
\[
J_{\mathrm N}(\pi)
=
\left(\prod_{m=1}^M V_m(\pi)\right)^{1/M},
\qquad
J_{\mathrm U}(\pi)
=
\frac{1}{M}\sum_{m=1}^M V_m(\pi),
\qquad
J_{\mathrm E}(\pi)
=
\min_{m\in[M]}V_m(\pi).
\]For \(F\in\{\mathrm N,\mathrm U,\mathrm E\}\), define the corresponding welfare suboptimality by
\[
\operatorname{SubOpt}_F(\widehat{\pi})
=
\max_{\pi\in\Pi}J_F(\pi)-J_F(\widehat{\pi}).
\]The left panel of Figure~\ref{fig:nash_simulation} reports the mean Nash welfare suboptimality of our method and the three baselines, together with \(95\%\) Monte Carlo confidence intervals over the \(100\) repetitions. The right panel evaluates the policies produced by our method and pooling-obs under the Utilitarian and Egalitarian objectives. These policies remain Nash-targeted and are not re-optimized for the alternative welfare objectives. For each \(F\in\{\mathrm U,\mathrm E\}\), the reported quantity is
\[
\frac{
\mathbb E_{\mathrm{rep}}
[\operatorname{SubOpt}_F(\widehat{\pi})]
}{
\mathbb E_{\mathrm{rep}}
[
\operatorname{SubOpt}_F(
\widehat{\pi}_{\mathrm{pool\text{-}obs}})
]
}.
\]This is the ratio of the mean suboptimalities across repetitions.

We additionally measure the dispersion of party-level regret. For any learned policy \(\widetilde{\pi}\), define party \(m\)’s average classical regret as
\[
\overline{\operatorname{Regret}}_m(\widetilde{\pi})
=
\mathbb E_{\mathrm{rep}}
\left[
\max_{\pi\in\Pi}V_m(\pi)-V_m(\widetilde{\pi})
\right].
\]The mean pairwise difference in average party regret is
\[
D(\widetilde{\pi})
=
\frac{1}{M(M-1)}
\sum_{m\neq m'}
\left|
\overline{\operatorname{Regret}}_m(\widetilde{\pi})
-
\overline{\operatorname{Regret}}_{m'}(\widetilde{\pi})
\right|.
\]The Difference curve in the right panel reports $\frac{D(\widehat{\pi})}{D(\widehat{\pi}_{\mathrm{pool\text{-}obs}})},$ where a ratio below one means that our method produces a smaller mean pairwise disparity in parties’ average regrets than pooling-obs. This metric describes how evenly regret is distributed across parties in this experiment, serving as a partial evidence of fairness.

\subsubsection{Confidence-set coverage.}\label{app:exp1cov}
\begin{figure}[H]
    \centering
    \includegraphics[width=0.98\linewidth]{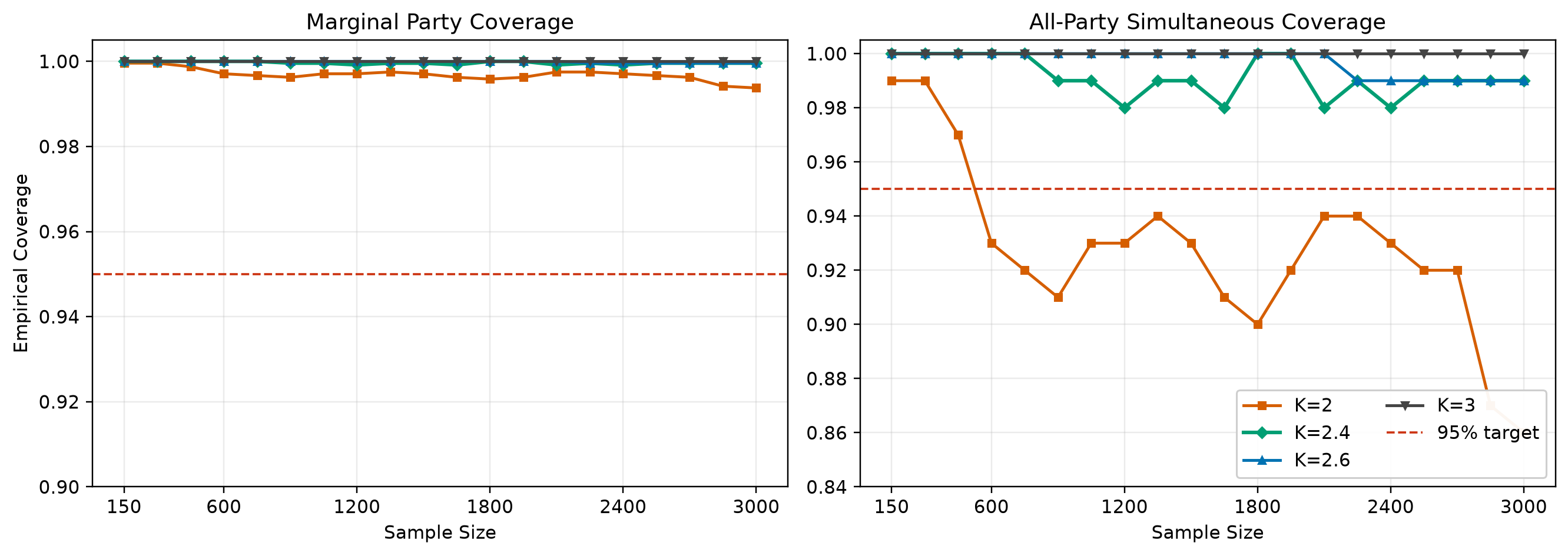}
    \caption{Empirical confidence-set coverage. The left panel reports
    marginal coverage over parties and repetitions, while the right panel
    reports simultaneous coverage over all \(24\) parties.}
    \label{fig:low-rank-coverage}
\end{figure}

For party \(m\), confidence-set coverage is the empirical probability of the event $\|\widehat\theta_m-\theta_m^*\|_{\Sigma_m}\leq\Gamma.$
Marginal coverage averages this event over parties and repetitions, whereas
simultaneous coverage requires the event to hold for all \(24\) parties in the
same repetition. The latter is the more demanding criterion and corresponds
more closely to the joint event required when optimizing a collective policy.
Figure~\ref{fig:low-rank-coverage} reports both measures for
\(K\in\{2,2.4,2.6,3\}\).

For the main choice \(K=2.4\), the aggregate marginal coverage is \(99.96\%\), while simultaneous coverage averaged over the sample-size grid is \(99.15\%\). The minimum simultaneous coverage at an individual sample size is \(98\%\).
Thus, \(K=2.4\) provides consistently high marginal and simultaneous empirical coverage across the reported sample-size range under the current experimental setup.

\subsubsection{Estimation accuracy and efficiency.}
To separate the benefit of shared-representation learning from the effect of
pessimistic policy selection, we compare the joint and independent estimators
using
\[
\mathcal E_{\Sigma_*}(\widehat\Theta)
=
\left\{
\frac{1}{M}
\sum_{m=1}^M
\|\widehat\theta_m-\theta_m^*\|_{\Sigma_*}^2
\right\}^{1/2}.
\]
The fixed population covariance \(\Sigma_*\) is used only for this diagnostic;
the pessimistic policy optimization continues to use the party-specific
empirical covariance matrices \(\Sigma_m\).

\begin{figure}[H]
    \centering
    \includegraphics[width=0.64\linewidth]
    {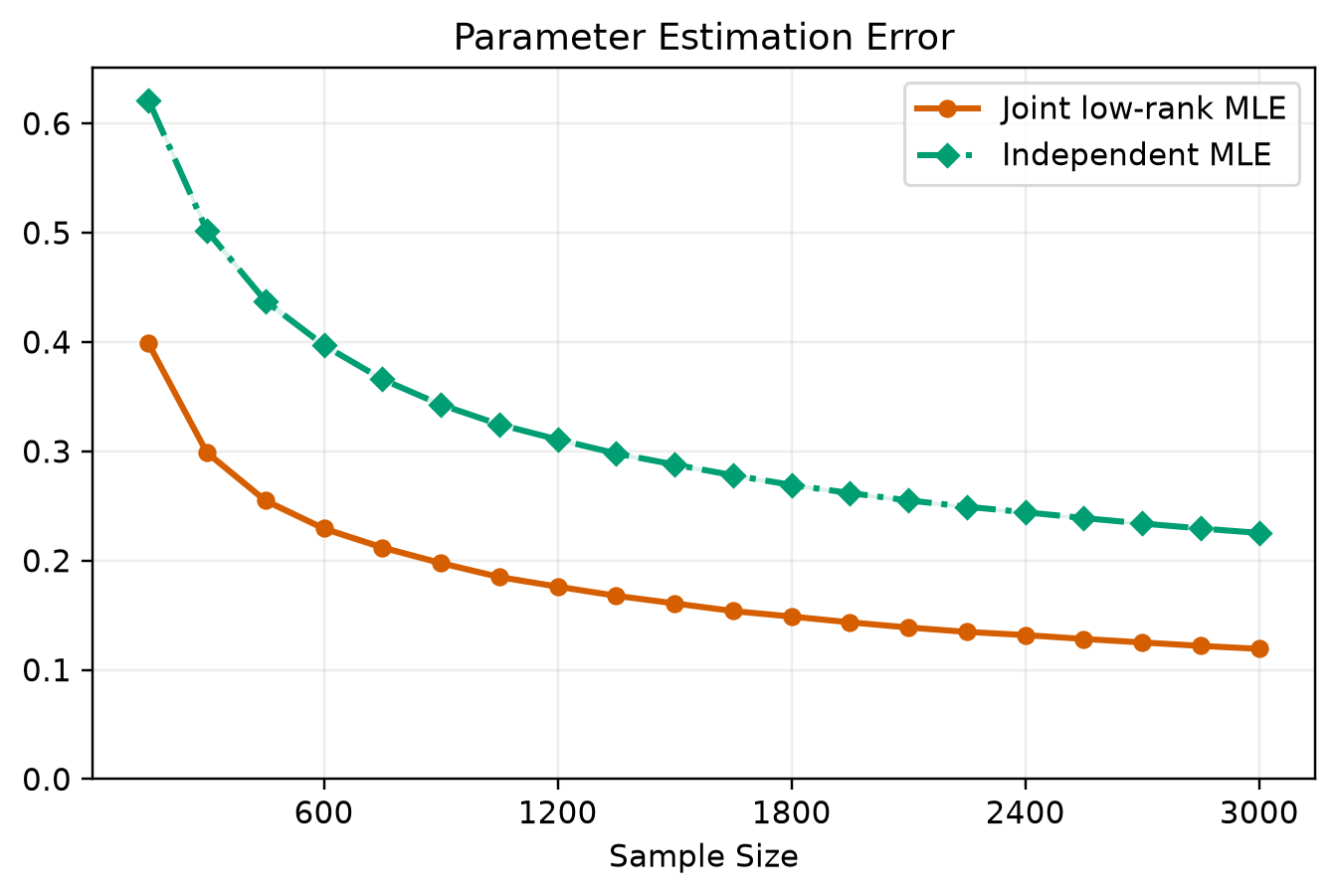}
    \caption{Population-covariance-weighted estimation error of the joint
    low-rank and independent MLEs.}
    \label{fig:low-rank-estimation}
\end{figure}

The joint estimator has lower estimation error than the independent estimator
throughout the reported sample-size range. At \(n=150\), its mean error is
\(0.399\), compared with \(0.622\) for the independent estimator. At
\(n=3000\), the corresponding errors are \(0.119\) and \(0.225\). Both
estimators improve as the sample size increases, but the persistent gap shows
the statistical benefit of estimating the shared three-dimensional
representation rather than learning \(24\) coordinates
separately for every party.

\subsection{Further Details for the General Pairwise-Preference Experiment (Section~\ref{sec:exp2})}
\label{app:general-preference-experiment}

This section provides implementation and diagnostic details for the general
pairwise-preference experiment reported in Figure~\ref{fig:von_simulation}.
We use a separate one-state setting. Although we use a BTL model to generate a controlled population preference matrix, the proposed method and the pooling baseline observe only pairwise comparison outcomes. Denote the state by $s$, let $\mathcal A=\{a_1,a_2,a_3\}$, and set $M=2$.  The feature
vectors and ground-truth parameters are
\[
    \phi(a_1)=(1,0)^\top,\quad
    \phi(a_2)=(0,1)^\top,\quad
    \phi(a_3)=(0,0)^\top,
    \qquad
    \theta_1^*=(1,0.1),\quad \theta_2^*=(0.2,1).
\]
Thus, $R_{\theta_m^*}(s,a)=\langle\theta_m^*,\phi(a)\rangle$.  We use the
BTL model only to generate a controlled population preference matrix:
\[
    \mathbb P_m(y=1\mid a,a',s)
    =\Phi\!\left(R_{\theta_m^*}(s,a)-R_{\theta_m^*}(s,a')\right),
    \qquad
    \Phi(x)=\frac{1}{1+\exp(-x)}.
\]
Neither the proposed method nor the pooling baseline observes the rewards,
features, or parameters.  Following Section~\ref{sec:ext}, the party-specific
population preference matrix is
\[
    N_{s,m}^*(a,a')
    :=2\mathbb P_m(y=1\mid a,a',s)-1,
    \qquad N_{s,m}^*(a,a)=0,
\]
and the equally weighted population target is
$N_s^*:=\frac{1}{M}\sum_{m=1}^M N_{s,m}^*=(N_{s,1}^*+N_{s,2}^*)/2$.

\paragraph{Sampling and estimation.}
The minimal party sample size is
$n\in\{400,800,\ldots,4000\}$.  We allocate
$|\mathcal D_1|=n$ comparisons to Party 1 and
$|\mathcal D_2|=2n$ to Party 2, and distribute each party's
comparisons evenly among the three unordered action pairs.  Each setting is repeated 500 times.
Using the notation in Section~\ref{sec:pes-von}, let
\[
    n_0(m,a,a',s)
    :=\#_m(a\succ a';s)+\#_m(a'\succ a;s)
\]
and define
\[
    N_{s,m}(a,a')
    :=\frac{\#_m(a\succ a';s)-\#_m(a'\succ a;s)}
    {n_0(m,a,a',s)}.
\]
We then form the party-balanced matrix and its confidence bonus as
\[
    N_s(a,a')=\frac{1}{M}\sum_{m=1}^M N_{s,m}(a,a'),
    \qquad
    B_s(a,a')=
    \sqrt{\frac{2\log(4MSA^2/\delta)}
    {\min_m n_0(m,a,a',s)\lor 1}},
\]
for $a\neq a'$, with zero diagonal and $\delta=0.05$.  Here $S=1$ and
$A=3$.  We implement Algorithm~\ref{alg2} exactly by solving
\[
    \hat p_s,\hat q_s
    =\arg\max_{p\in\Delta(\mathcal A)}
      \min_{q\in\Delta(\mathcal A)}p^\top(N_s-B_s)q.
\]
The observation-pooling baseline first pools the two parties' comparison
counts to construct $N_{s,\mathrm{pool}}$, and then solves the corresponding
matrix game without a confidence bonus.  Both methods are evaluated on the
same simulated data in each repetition.

\paragraph{Evaluation and the pooling limit.}
Because $N_s^*$ is skew-symmetric, its game value is zero.  We therefore
report the nonnegative game-value suboptimality
\[
    -\min_{q\in\Delta(\mathcal A)}\hat p_s^\top N_s^*q.
\]
Figure~\ref{fig:von_simulation} shows its mean and 95\% Monte Carlo confidence
interval over the 500 repetitions.  For this instance,
\[
N_s^*\approx
\begin{pmatrix}
0 & 0.02098 & 0.28089\\
-0.02098 & 0 & 0.25604\\
-0.28089 & -0.25604 & 0
\end{pmatrix}.
\]
Hence $a_1$ is the Condorcet winner of $N_s^*$: it has positive population
preference margin against every other action.  In contrast, because Party 2
provides twice as many comparisons, observation pooling converges to
$(N_{s,1}^*+2N_{s,2}^*)/3$, whose Condorcet winner is $a_2$.  Selecting $a_2$
has suboptimality $0.02098$ under the equally weighted target $N_s^*$, which
explains the non-vanishing pooling curve.  Algorithm~\ref{alg2} preserves the
intended equal party weights by normalizing within each party before
aggregation.

\paragraph{Lower-bound coverage.}
We additionally examine whether \(N_s-cB_s\) forms a valid simultaneous lower confidence bound for the population preference matrix. For each \(c\in\{0.2,0.3,\ldots,1.0\}\), we test whether
\[N_s(a,a')-cB_s(a,a')\leq N_s^*(a,a')\]
holds simultaneously for all \(a\neq a'\). We remark that the multiplier $c$ is not an algorithmic
hyperparameter: Algorithm~\ref{alg2} and the main experiment use $N_s-B_s$, corresponding to $c=1$.  

\begin{figure}[t]
    \centering
    \includegraphics[width=0.68\textwidth]
        {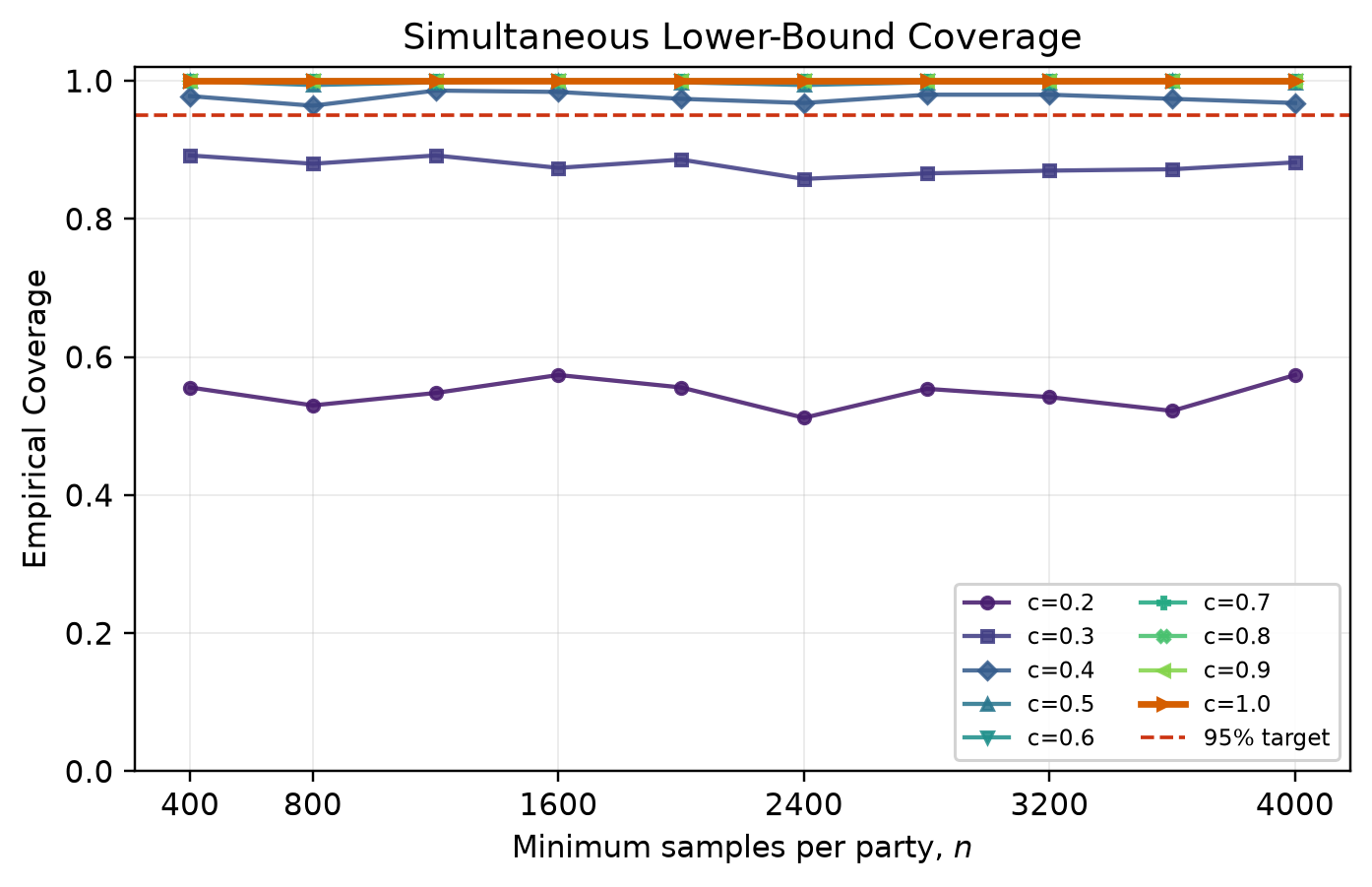}
    \caption{Empirical simultaneous lower-bound coverage for the general
    pairwise-preference experiment.  Each curve corresponds to a diagnostic
    multiplier $c$ on $B_s$, and the dashed line marks the nominal
    $1-\delta=95\%$ level.  Algorithm~\ref{alg2} and the main experiment use
    the full confidence bonus ($c=1$).}
    \label{fig:von-coverage}
\end{figure}

Because $N_s$ and $N_s^*$ are skew-symmetric and $B_s$ is symmetric, this is
equivalent to simultaneous two-sided coverage over the three unordered
pairs.   Figure~\ref{fig:von-coverage} shows that the full
bonus covers in every repetition at every sample size.  Coverage is at least
$99.4\%$ for $c=0.5$ and falls below the nominal 95\% level for $c\leq0.3$,
indicating that the theoretical bonus is conservative on this instance.

\end{document}